\documentclass[12pt]{article}
\usepackage{amssymb,amsmath,amsthm,graphicx,ulem}

\usepackage{graphicx,subfigure}
\usepackage{epsfig}
\usepackage{amsmath}
\usepackage{amsfonts}
\usepackage{amssymb}
\usepackage[usenames]{color}
\usepackage[letterpaper,left=2.cm,right=2.cm,top=2.5cm,bottom=2.5cm]{geometry}
\usepackage{float}
\usepackage[table]{xcolor}
\usepackage{hyperref}
\usepackage{wrapfig}
\usepackage{subfigure}
\usepackage{adjustbox}
\hypersetup{
    colorlinks = true,
    linkcolor = {black},
    urlcolor = {blue},
    citecolor = {red},
}
\usepackage{bm}
\usepackage[utf8]{inputenc}
\usepackage[english]{babel}
\usepackage{url}
\usepackage{authblk}
\usepackage{cite}
\usepackage{hhline}
\usepackage{tabularx}
\usepackage{caption}
\usepackage{comment}
\usepackage{booktabs}

\usepackage{fourier}
\usepackage{makecell}

\usepackage[ruled,vlined]{algorithm2e}

\newcommand{\gavin}[1]{\textcolor{blue}{}}

\DeclareMathOperator*{\argmax}{argmax} %

\begin{document}
 
\allowdisplaybreaks
\normalem
\title{Deep Generative Modeling in Network Science \\with Applications to Public Policy Research}

\author{Gavin~S.~Hartnett%
\thanks{Correspondence may be addressed to Gavin Hartnett (hartnett@rand.org) or Raffaele Vardavas (rvardava@rand.org). 
\newline \newline
This document has not been formally reviewed, edited, or cleared for public release. It should not be cited without the permission of the RAND Corporation. RAND's publications do not necessarily reflect the opinions of its research clients and sponsors. RAND\textregistered\ is a registered trademark.
}}
\author{Raffaele~Vardavas$^*$%
}
\author{Lawrence~Baker%
}
\author{Michael~Chaykowsky%
}
\author{C.~Ben~Gibson%
}
\author{Federico~Girosi%
}
\author{David~Kennedy%
}
\author{Osonde~Osoba%
}

\affil{RAND Corporation, Santa Monica, CA 90401, USA}

\date{}
\renewcommand\Affilfont{\itshape\small}

\date{}
\maketitle 

\begin{abstract}
Network data is increasingly being used in quantitative, data-driven public policy research. These are typically very rich datasets that contain complex correlations and inter-dependencies. This richness both promises to be quite useful for policy research, while at the same time posing a challenge for the useful extraction of information from these datasets - a challenge which calls for new data analysis methods. In this report, we formulate a research agenda of key methodological problems whose solutions would enable new advances across many areas of policy research. We then review recent advances in applying deep learning to network data, and show how these methods may be used to address many of the methodological problems we identified. We particularly emphasize deep generative methods, which can be used to generate realistic synthetic networks useful for microsimulation and agent-based models capable of informing key public policy questions. We extend these recent advances by developing a new generative framework which applies to large social contact networks commonly used in epidemiological modeling. For context, we also compare and contrast these recent neural network-based approaches with the more traditional Exponential Random Graph Models. Lastly, we discuss some open problems where more progress is needed.
\end{abstract}

\newpage
\tableofcontents
\baselineskip16pt
\clearpage
\section{Introduction \label{sec:introduction}}
The convergence between computational simulation modeling, large datasets of unprecedented scale, and recent advancements in machine learning represent an exciting new frontier for the scientific analysis of complex systems. This is particularly the case in the study of large networked systems, such as those found in biology, critical infrastructures, and social science.  For example, in biology this convergence promises to help us better understand how the behavior of individual neurons affects overall brain activity. For large networked infrastructures that are critical to the modern world, such as power grids, communication networks, or transportation networks, large datasets together with machine learning can be used to optimize their design and ensure that they are robust to failures. As yet another example, many social systems may be represented by graphs describing interactions among various complex entities (e.g., people, firms, organizations, etc.). These entities are often adaptive and change their behavior based social interactions and changes in the landscape and environment that they in part help form. The graph structure governs how these entities (also known as actors or agents) interact. Agents are represented by vertices  (also known as nodes) on the graph, and their interaction occurs through the edges (also known as ties or links). Microsimulation and Agent-Based Models (ABMs) are examples of two powerful computer simulation techniques that are increasingly becoming essential tools to predict societal evolution and to compare the impacts of different “what-if” scenarios. Agent-based models in particular can model the complex social systems with adaptive agents that interact over social and physical networks. These models can be used by quantitative public policy researchers to inform policy problems such as those involving public health strategies and the design of economic markets. 

The scope of public policy challenges is broad and evolving, as is the range of recently developed quantitative and data-driven methodologies, especially those methodologies that employ machine learning. With this report, we aim to help bridge the gap between the machine learning research community and the community of quantitative public policy researchers to accelerate the application of new and powerful learning algorithms to complex public policy challenges. We have written the report to be broadly useful for many different policy problem areas, although the underlying motivation for many of the datasets and methodologies we explore is the problem of modeling the spread of contagion over a complex and evolving population. This is because modeling a contagion provides a clear example of a system where the dynamics can strongly depend on network structure. The question of whether a pathogen will percolate through a population and infect a significant fraction of individuals depends both on properties intrinsic to the pathogen (such as the degree of infectivity and the incubation period) as well as properties of the underlying network of connections between individuals. The spread of a pathogen in a densely connected population will be much more rapid than in a sparsely connected one - hence the critical importance of social distancing measures in response to the COVID-19 pandemic~\cite{courtemancheStrongSocialDistancing2020}. The case of COVID-19 has underscored the importance of another network phenomenon on contagion spreading, namely the spreading of rumors, risk perceptions, and protective behaviors. In particular, it should be clear that it is not merely enough to model the spread of an infectious disease through some static population, which can be done rather straightforwardly with deterministic, population-based compartmental models (i.e. variants of the SEIR-model), where the population is divided into compartment classes including the susceptible, exposed (latent), infectious, and removed (e.g., recovered or dead). In reality, the population is adaptive and individual behaviors will be greatly affected by any official policy actions (school closures, mandatory social distancing measures, etc.), as well as by the risk tolerance, beliefs, and values held by individuals. Additionally, there is evidence that in situations such as this an individual's actions are strongly influenced by the actions of their immediate social network~\cite{SocialSamplingExplains}, which underscores the importance of the network structure \cite{bruinedebruinReportsSocialCircles2019, bruinedebruinRoleSocialCircle2020}.

As with most models, the strength of simulating social systems with microsimulation models and agent-based models depends on the way these models are informed and parametrized. In contagion modeling, it is often the case that the entities in these simulation models are individual people. To inform these models, researchers are increasingly reliant on large individual-level datasets to accurately represent the heterogeneities of the population and the many ways the individuals may interact over social and physical networks. These datasets typically include socio, economic, health, and demographic attribute information, as well as information on attitudes, perceptions, preferences, and behaviors. Some of these attributes (also known as features) are typically assumed to be static (e.g., gender and race) while others are dynamic attributes. Of the dynamic attributes, some change due to simple processes that do not depend on the network (e.g. age), while others such as behaviors change with individual and aggregate level outcomes. For example, protective behaviors during an epidemic such as getting vaccinated or wearing a mask change with evolving risk perceptions, which are in turn affected by personal experiences and observations of social network outcomes, as well as by broadcast media reporting on population-level outcomes. Microsimulation and agent-based models need as inputs an initial condition describing in a representative way the population it will model; this includes the individual attributes and the initial network structure.  

Motivated by a long-term goal of improving microsimulation and agent-based models of complex social systems for use in public policy research, in this report we consider the precursor problem of how to best inform these models using network datasets that contain different information and are of different scales. We will review and summarize recent developments in deep learning for networks and show how they can be used to address a broad range of challenges associated with building large-scale and high-fidelity simulation models. For example, the network datasets often have missing values as a result of the difficulty of obtaining a complete survey of a population. Deep neural networks may be used to impute these missing values, allowing the completed dataset to be used in subsequent simulations. Deep neural networks may also be used to generate entirely new, artificial network datasets that are statistically similar to a real dataset. The newly generated dataset may be thought of as an anonymized version of the real dataset, which would then allow compliance with privacy and data protection requirements constraining what types of data can be used in the simulations. We also consider the problem of merging or fusing, two non-overlapping datasets. By addressing this problem, we can take two datasets that capture different parts of a population and/or different individual attributes, and fuse them to form a richer and more complete dataset. 

\begin{figure}[!ht]
\centering
\begin{subfigure}
  \centering
  \includegraphics[width=0.75\textwidth]{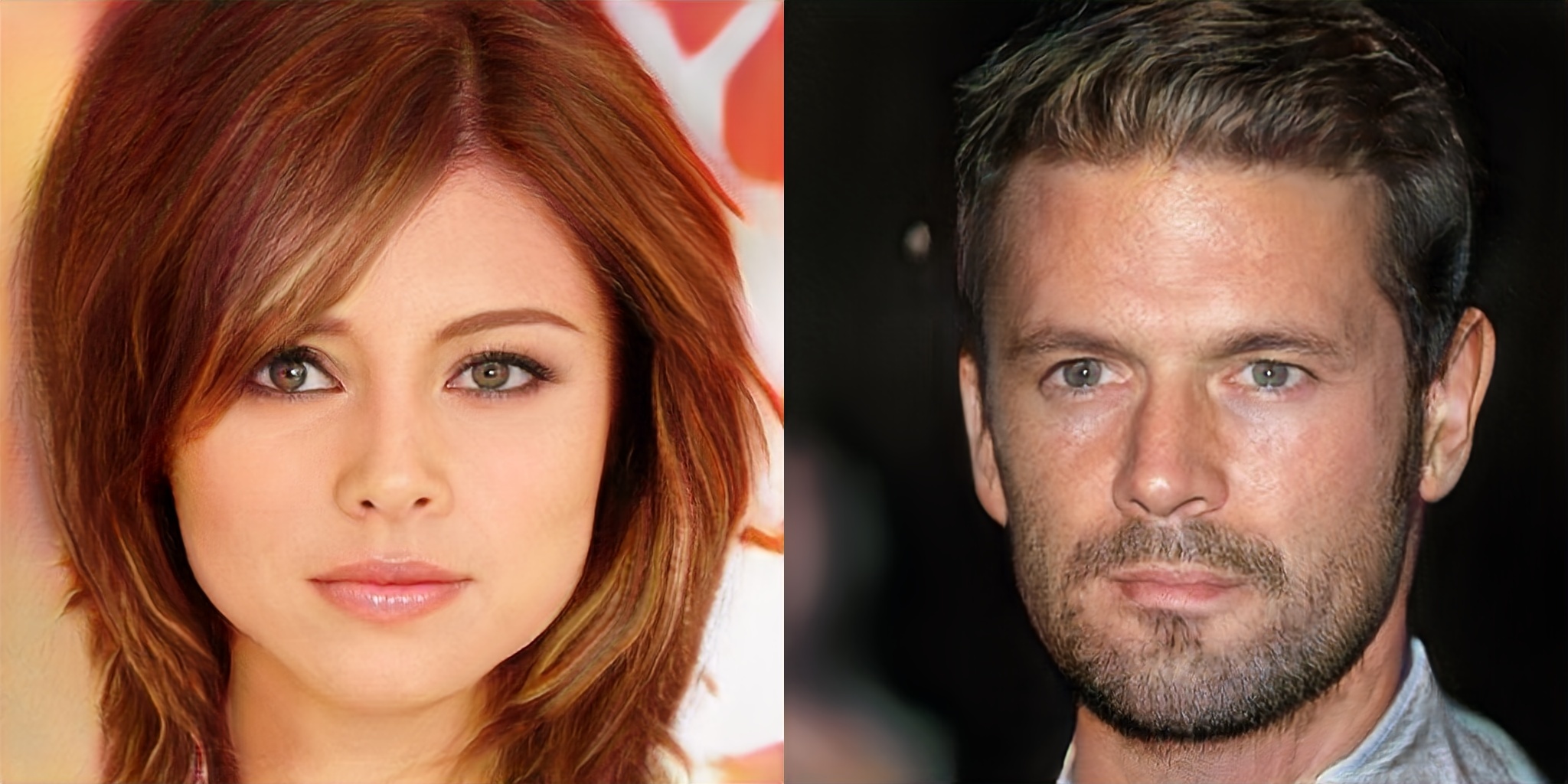}
\put(-400,80){(a)}
\end{subfigure}%
\\
\begin{subfigure}
  \centering
  \includegraphics[width=0.75\textwidth]{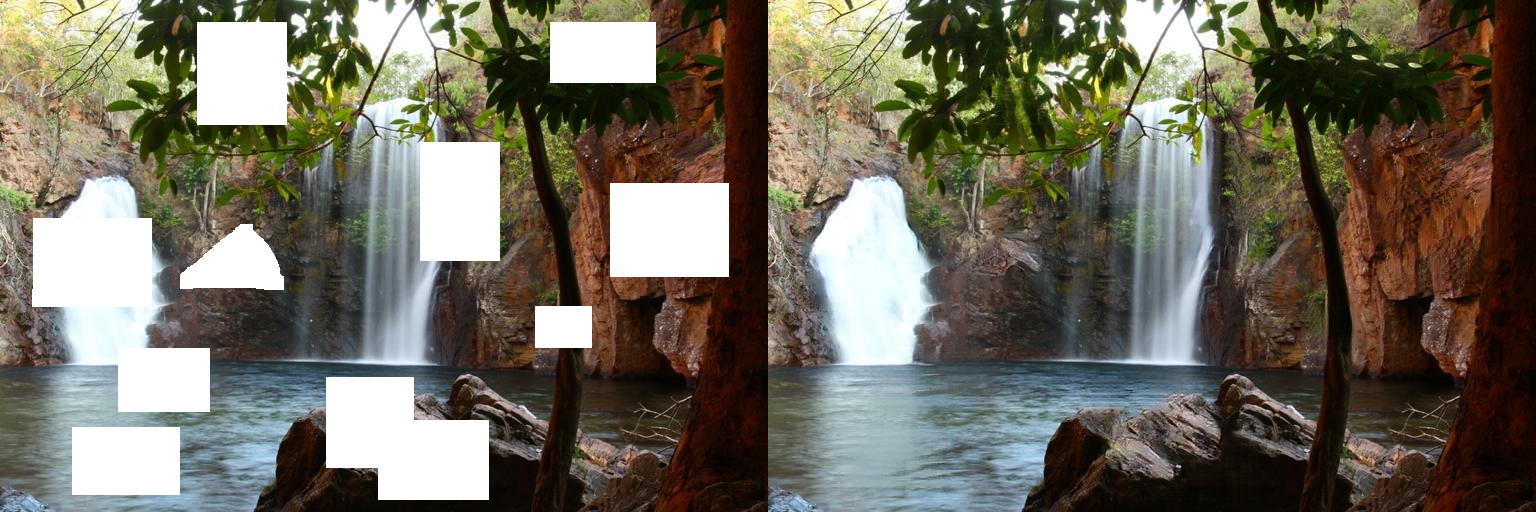}
\put(-400,80){(b)}
\end{subfigure}%
\\
\begin{subfigure}
  \centering
  \includegraphics[width=0.75\textwidth]{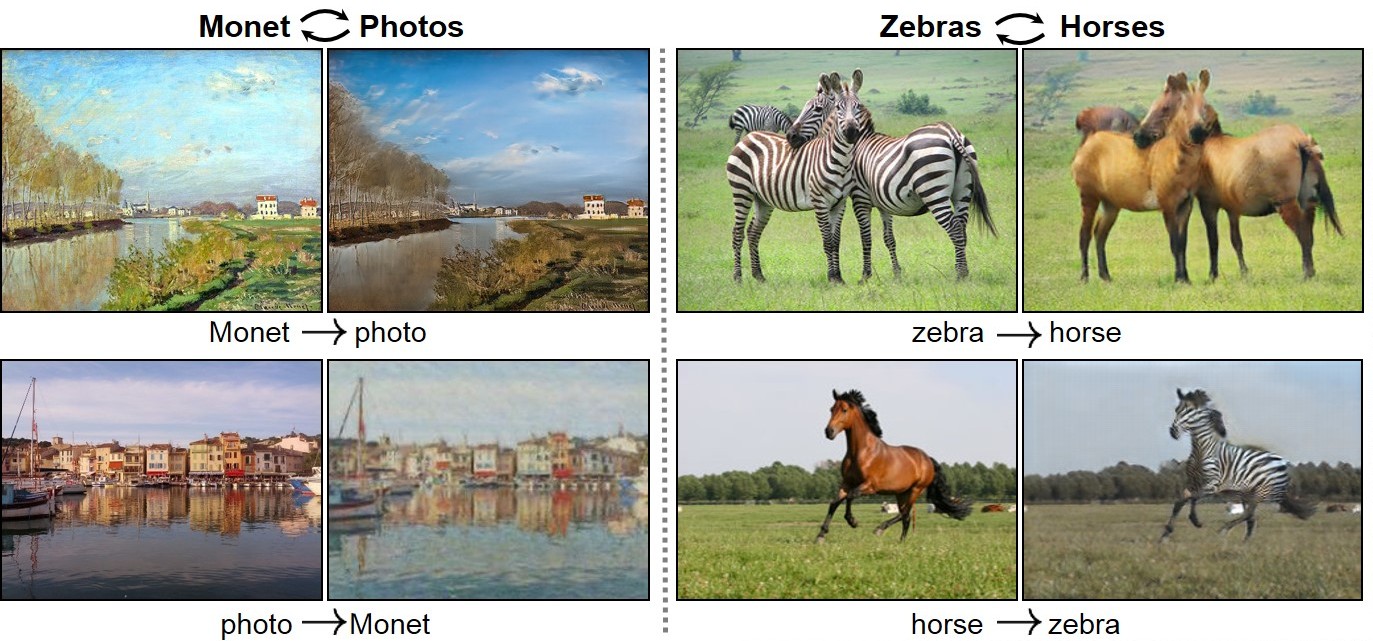}
\put(-400,100){(c)}
\end{subfigure}%
\caption{(a) GANs can be used to generate high-quality, high-resolution artificial faces, image taken from \cite{karras2017progressive}. (b) An example of how GANs may be used to impute missing pixel values of images, image taken from \cite{yu2018generative}. (c) CycleGAN may be used to transfer the style of images, for example by converting a Monet painting into a realistic photo, or vice versa, image taken from \cite{zhu2017unpaired}.
\label{fig:GANs_for_images} }
\end{figure}  

In recent years, deep neural networks have facilitated the development of algorithms and methodologies capable of solving analogous problems in the image domain. For example, Generative Adversarial Networks (GANs) \cite{goodfellow2014generative} are neural network-based generative models that have proven very successful at generating artificial images after being trained on large datasets of real images. GANs can now generate high-resolution artificial faces that are extremely realistic \cite{karras2017progressive}, Figure \ref{fig:GANs_for_images} (a). GANs can also be used to impute (or impaint) images with missing pixel values \cite{yu2018generative}, Figure \ref{fig:GANs_for_images} (b), and they can modify an image to more closely match a second image while preserving the key features in the original \cite{zhu2017unpaired, karras2019style}, Figure \ref{fig:GANs_for_images} (c). These results suggest that deep neural networks could also be used to develop corresponding generative models over network data. This is especially true if one regards images as a special kind of network where the underlying graph is a square lattice, with the node features corresponding to pixels. And indeed, within the deep learning literature there has also been a parallel development of neural architectures tailored to network data that goes far beyond this simple analogy with images by taking into account the mathematical properties of graphs. Much of this recent work has been focused on developing predictive models, whereas here our main focus will be on generative models. We will also be especially interested in network data corresponding to the social contact networks of a large population of individuals, whereas this recent body of work addresses a much wider range of domains, including for example chemical bonds and road networks. The next section outlines the key methodological aims of this report and describes its layout.  

\section{Aims \label{sec:aims}}

The overarching aim of this report is to introduce methods in deep-neural networks and use them to develop methodological advancements in social network analysis with applications to inform simulation models. We focus on four research aims: synthetic network generation, rescaling networks, data imputation, and network fusion. The choice of these aims was the result of our understanding of existing methodological gaps and in-person interviews with RAND policy researchers working on applications of social networks. These interviews are described in Appendix~\ref{app:interviews}. We note that many of these aims are not new, and in some cases, methods that can address these challenges have existed for decades. In particular, the aims of synthetic network generation and network rescaling can be addressed using Exponential Random Graph Models (ERGMs). However,  existing methods fall short in their ability to inform simulation models with large-scale synthetic networks that are representative of those in the real-world. To address these aims and the limitations of existing methods our research investigates and uses deep learning-based methods. This approach is innovative and ambitious but at the same time has the potential to help guide a broader research program extending beyond the current work. In this section, we discuss these four aims in detail and present the layout of the subsequent sections that address these aims.

\subsection{List of Research Aims \label{sec:Reschaims}}

\begin{figure}[!ht]
\begin{center}
\includegraphics[width=0.65\linewidth]{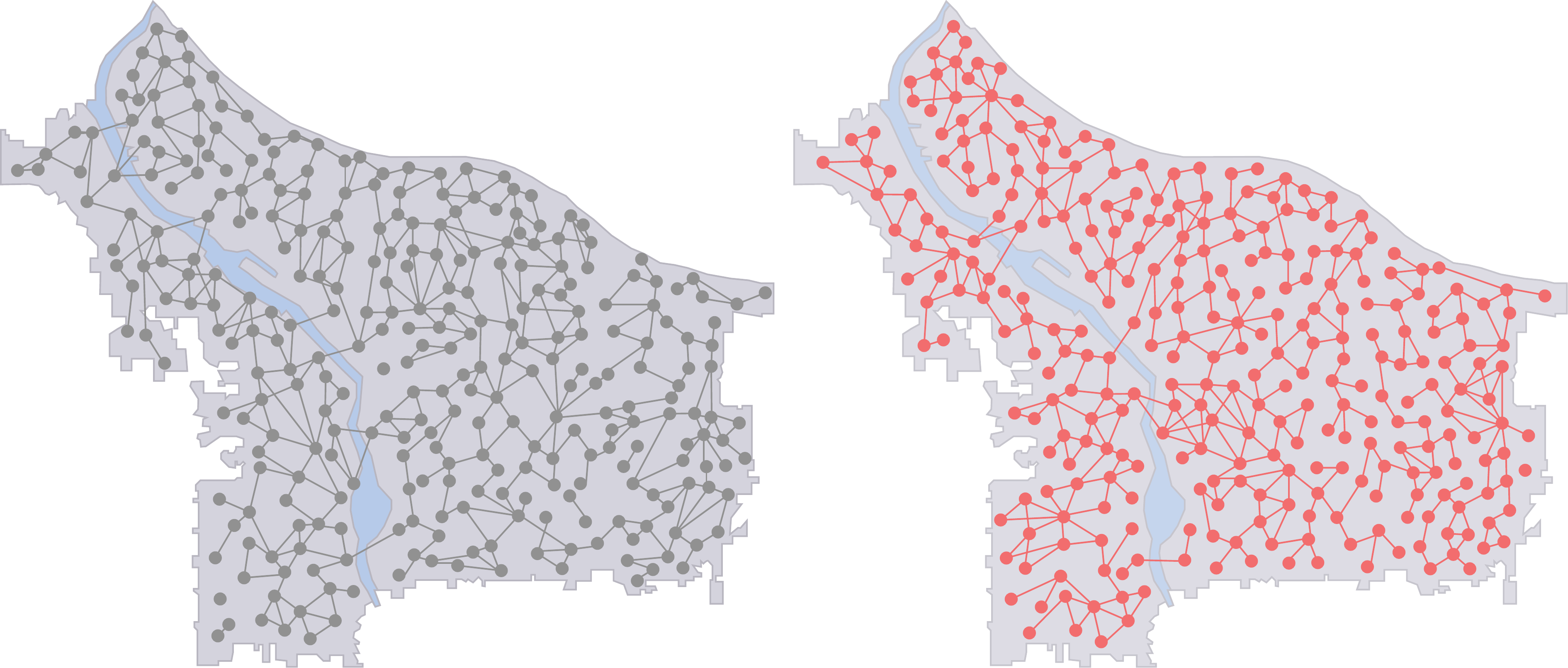}
\end{center}
\caption{Illustration of the aim to generate a new large-scale network for Portland, OR (shown on the right-hand side with red nodes and edges) using the NDSSL network for Portland, OR (shown on the left-hand side with gray nodes and edges). \label{Fig:gen_new_NDSSL}}
\end{figure}

\paragraph{Synthetic Network Generation:}
The first aim is to generate new large-scale synthetic networks that are representative of a city in the United States (U.S.) using an existing dataset. We aim to develop this technique using an existing synthetic network that is representative of the city of Portland, OR. This dataset is described in detail in Section \ref{sec:data}. We refer to this dataset as the NDSSL network  as it was generated by the Network Dynamics and Simulation Science Laboratory group at the Biocomplexity Institute of Virginia Tech. A schematic illustration of this aim is shown in Figure \ref{Fig:gen_new_NDSSL}. The aim is to learn from this large-scale  network and generate a new network with different nodes and edges such that the new population and their network of interactions is statistically equivalent to the original. By statistically equivalent we mean that (i) the macro-level joint distributions of node attributes at the city and zip level are the same as the original and, (ii) the new edge list preserves the way people mix across the different node attributes, which includes preserving network centrality distributions such as the degree distribution, as will be defined in the next section. 

Synthetic Network Generation would allow users to create an ensemble of large-scale networks that are statistically equivalent to the network provided by the dataset. Applications of this method include producing a de-identified network dataset that is statistically equivalent to the original dataset but that cannot be used to reproduce any parts of the original network. The resulting networks can more easily be distributed to research teams for analyses and modeling. Moreover, an ensemble of statistically equivalent networks can be used as initial condition inputs of simulation models allowing these models to explore sensitivities to the status of the initial population and network structure.    

\begin{figure}[!ht]
\begin{center}
\includegraphics[width=0.65\linewidth]{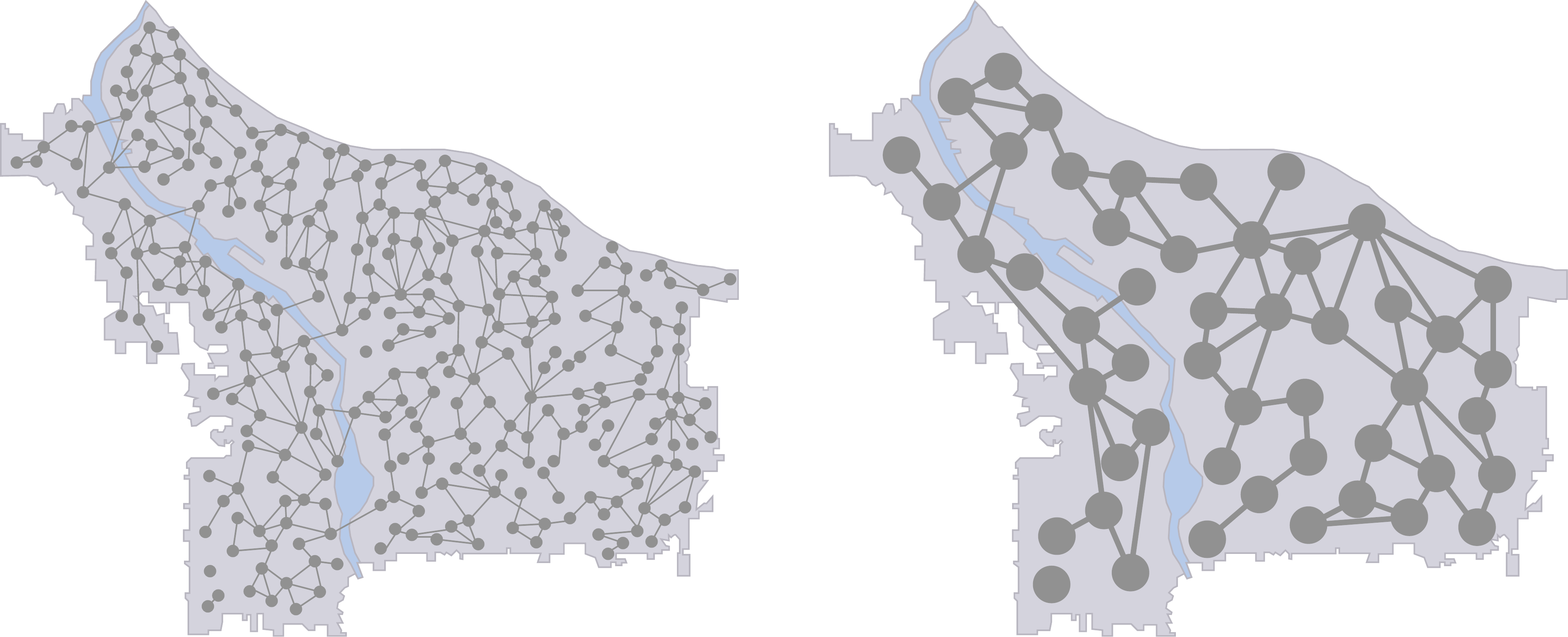}
\end{center}
\caption{Illustration of the aim to generate a new large-scale network for Portland, OR that has been scaled-down (shown on the right-hand side) using the NDSSL network for Portland, OR (shown on the left-hand side). 
\label{Fig:gen_new_down-scaled_NDSSL}}
\end{figure}

\paragraph{Rescaling Networks:}
The second aim is to generate a scaled-down version of an existing dataset of a large-scale network. An illustration of this aim is shown  in Figure \ref{Fig:gen_new_down-scaled_NDSSL}. We will again use the NDSSL network for Portland as our starting network and generate new down-scaled versions of the network. This aim is motivated by agent-based models and microsimulation models. These simulations operate at the level of individuals (as opposed to populations), and can be quite computationally expensive as a result. Developing these simulation models requires many trial runs for model verification, validation and calibration. This means that these simulation model need to run a very large number of times to explore different model assumptions and different combinations of model parameter values as defined within their plausible range. Therefore, it is desirable to verify, validate and calibrate these models using a scaled-down population and network before running policy experiments on the full network size. Rescaling a network is a known problem in network science and many methods have been developed for this purpose. Many of these methods use network cluster identification and generate a new network by replacing a whole cluster of nodes from the original network with a single node in the new network. For example, if individuals belong to households as they do in the NDSSL network, then an easy way to down-scale is to cluster individuals into their household units and have the households unit represent the new nodes in the down-scaled network. The problem with this approach is that node attributes now represent household attributes and individual-level heterogeneities that may be important to the simulation model are lost. We thus need a scaled-down network of individuals not of clusters of individuals. 

One major challenge with rescaling a network is that there is no single and unambiguous way to consider one network to be a rescaled version of another. For example, if the rescaled graph is to have twice as many nodes as the original, how should the other graph statistics be expected to scale? Should the number of edges also increase by a factor of 2, which would preserve the edge/node ratio? Or perhaps for a given application it would make more sense to increase the number of edges by a factor of $2^2 = 4$, to preserve the edge density of the new network. How should the number of triangles scale, and what should the new degree distribution look like? It is important to recognize that any approach towards rescaling will, either explicitly or implicitly, involve answering these questions, and ideally the answers would be informed by the application at hand.

\begin{figure}[!ht]
\centering
\includegraphics[width=1.0\textwidth]{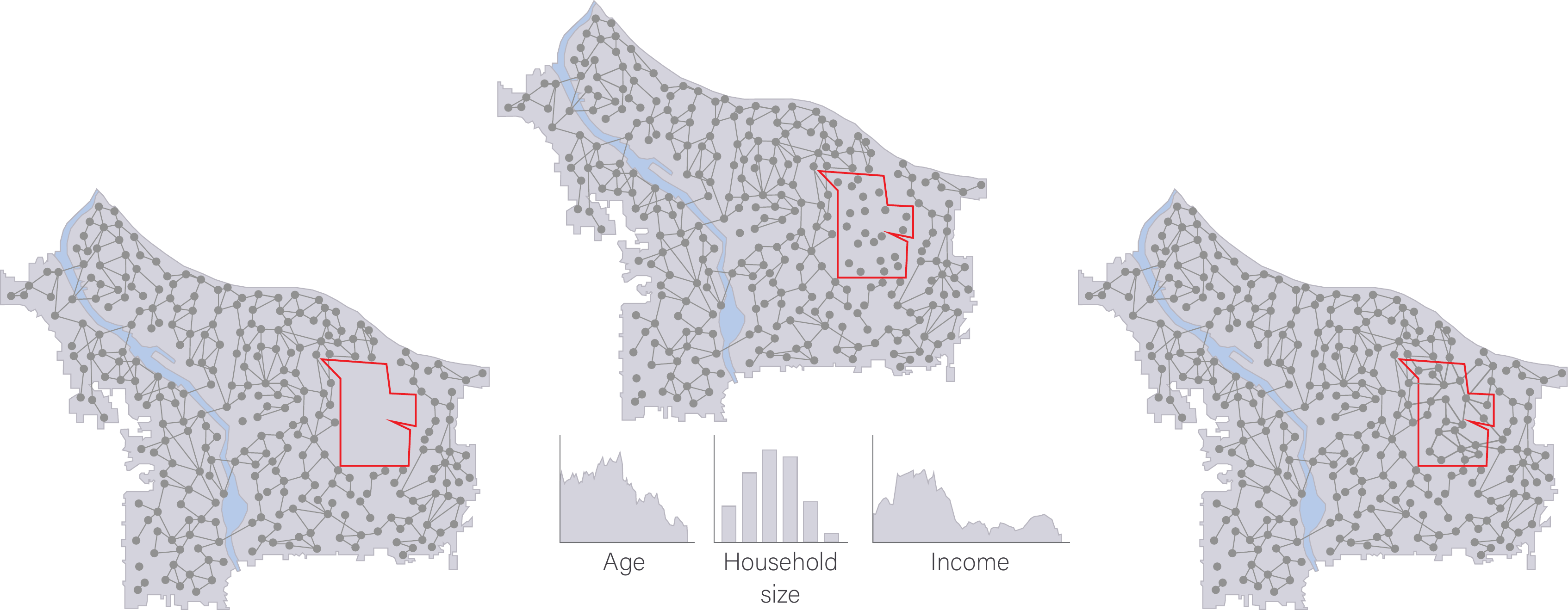}
\caption{Illustration of the aim to impute missing nodes and edges of a zip code in Portland. 
\label{Fig:gen_NDSSL_imputed_Network}}
\end{figure}

\paragraph{Data Imputation:}
The third aim is network imputation, which requires generating the nodes and edges of either an existing geographical region whose data is missing or of an entirely new geographical region. Figure \ref{Fig:gen_NDSSL_imputed_Network} illustrates our aim. 
\begin{figure}[!ht]
\begin{center}
\includegraphics[width=0.65\textwidth]{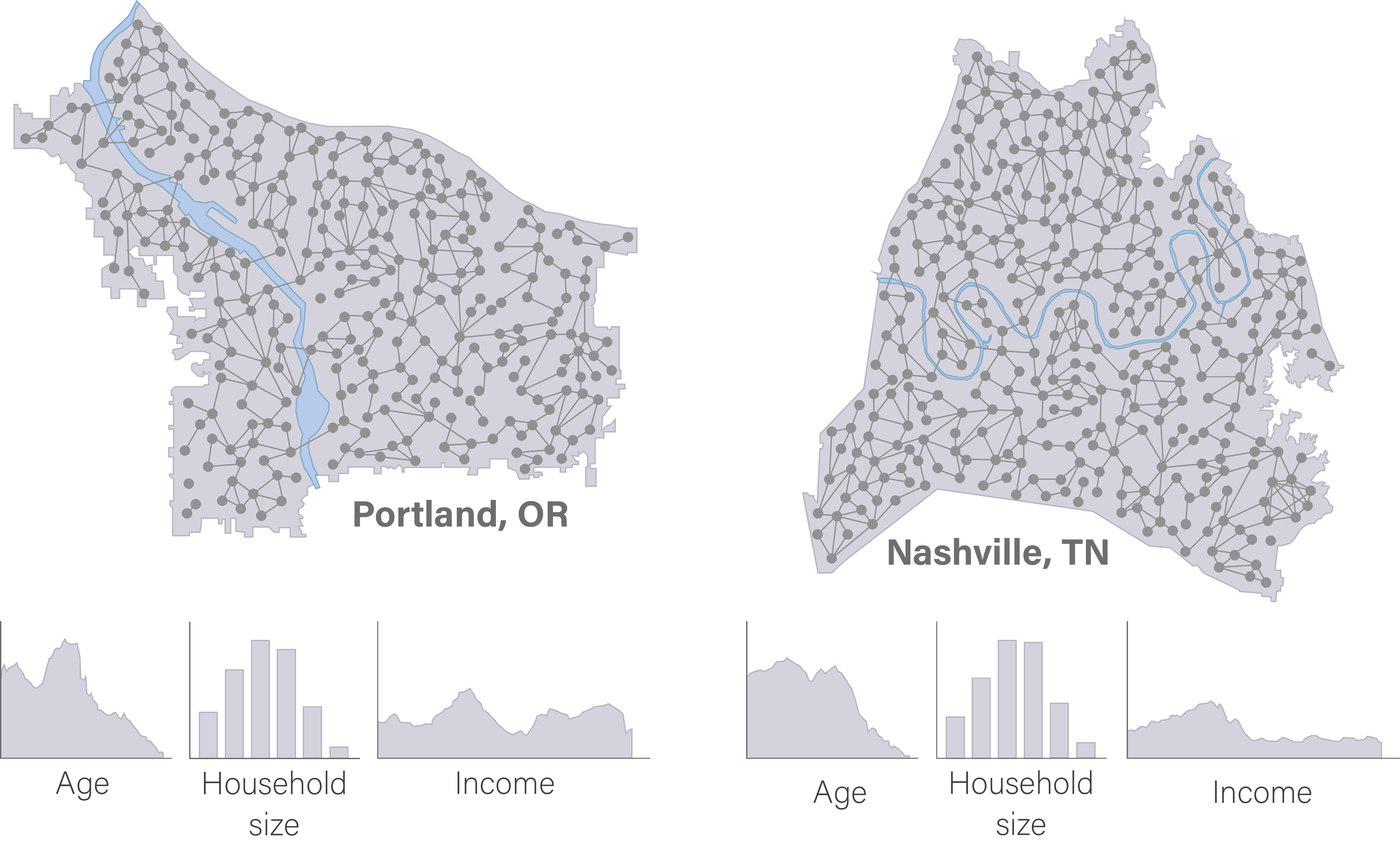}
\end{center}
  \caption{Applying the imputation to generate an entire different city.
   \label{fig:NDSSL-Nashville}}
\end{figure}
Using the NDSSL network for Portland, an example of the first case is when the data of individuals belonging to households of a given zip code are entirely missing. The goal is to generate the population of individuals and household in the missing zip code using (i) the learned network structure of the entire network that includes individuals belonging to households in the other zip codes, and (ii) the known joint or marginal demographic distributions of individual attributes of the blanked zip code (e.g., age group, household income, etc.). A natural extension of this aim is to generate the population and network of an entirely new city. Figure \ref{fig:NDSSL-Nashville} illustrates this sub-aim. Using the known joint or marginal demographic distributions of individual attributes for the new city (e.g., Nashville, TN) and for each zip code in the new city, we can use the imputation method to generate a synthetic network dataset from our NDSSL network for Portland, OR data.

Data imputation on networks has many applications. It allows users to impute missing data in less complete networks or sub-networks and thus can be used as part of a data collection strategy in which easily observed network or relationship data is used to infer difficult to observe relationship data. For example, it could be used to impute unreported associations in a network of financial relationships and to infer the reach of information on social media based on observed interactions. Network data imputation has application in informing simulation models as it would allow transforming a network dataset that is representative of a given city or geographical area to be representative of a different city or geographical area while taking into account the differences in population size, density, demographics, and geographical features.  

\paragraph{Network Fusion:}
Our fourth aim is network fusion, by which we mean the generation of a new, large-scale network that captures aspects of two network datasets. In dealing with real-world networks, it is often difficult or impossible to completely map out a large-scale network for a population. Policy researchers are frequently faced with multiple, partially overlapping datasets describing the population of interest. Just as data fusion is a broad term representing the process of integrating multiple data sources, we will use network fusion to broadly represent the challenge of fusing data with a network structure.

\begin{figure}[!ht]
\begin{center}
\includegraphics[width=0.55\linewidth]{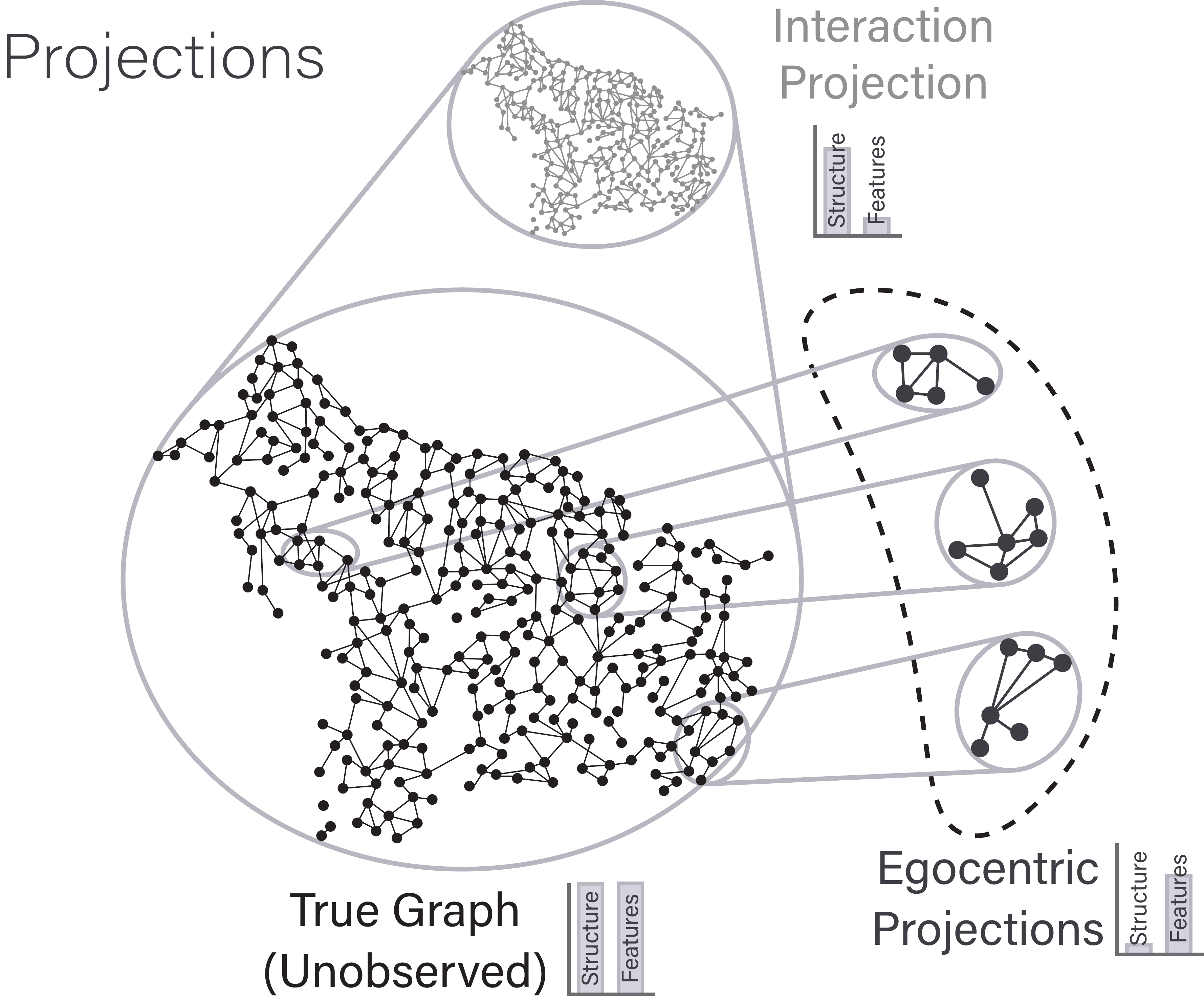}
\end{center}
  \caption{Illustration of network data fusion. We postulate that there exist a true unobserved graph that is consistent with our NDSSL sociocentric network data (i.e., the interaction projection) and the FluPaths ALP egocentric data (i.e., egocentric projections).   
   \label{fig:NDSSL_Projection}}
\end{figure}

As a concrete example, we will consider the fusion of the two datasets reviewed in Section \ref{sec:data}, the NDSSL sociocentric network data and the FluPaths egocentric data. As an example of a data fusion problem in this context, Figure \ref{fig:NDSSL_Projection} illustrates a problem where we are given both the NDSSL and FluPaths data, and the goal is to find a third graph that ``contains'' both of these datasets. We imagine that there is a true, unobserved graph from which either the NDSSL or FluPaths data could be obtained. For example, if an egocentric sampling procedure were hypothetically applied to this true graph, then we would generate a new dataset that would statistically agree with the FluPaths data. Similarly, if the entire network were sampled, but only a subset of the node and edge attributes were recorded, then we would recover the NDSSL network. Therefore, the postulated ``true graph'' is implicitly defined as being consistent with both the NDSSL and FluPaths data. Of course, this is just one possible interpretation of the general concept of network fusion, and much work remains to convert this general description into a well-defined mathematical problem. We will postpone any discussion of this until Section~\ref{sec:fusion}, where fusion is addressed in more detail.

\subsection{Report Layout \label{sec:layout}}
The layout of this report is as follows. In Section \ref{sec:graphNN} we review Graph Neural Networks (GNNs), a broad family of neural network architectures capable of processing network data. This will introduce the methodological foundations that we use and build from in the later sections that address our aims. In Section \ref{sec:data} we introduce and review two policy-relevant datasets. These datasets will be used throughout the report to illustrate how the methods we introduce can be used in practice to address our aims. In Section \ref{sec:predictive} we show how GNNs can be used to solve predictive problems and the aim of {\bf Data Imputation}. Then in Section \ref{sec:generative} we show how these methods can be used for generative problems. We will introduce a new approach to generative modeling called Graph Generation By Iterated Link Prediction, which we use to directly addresses the aims of 
{\bf Synthetic Network Generation} and {\bf Rescaling Networks}. In Section \ref{sec:ergm} we compare and contrast these deep learning-based generative approaches to Exponential Random Graph Models (ERGMs). Following that, in Section \ref{sec:fusion} we discuss how deep learning-based generative models can be used to address the aim of Network Fusion.  The report concludes with a discussion and outlook to future work in Section \ref{sec:outlook}.  

We have attempted to keep the report as self-contained and clear as possible to an audience with some familiarity with social network research and machine learning methods. We have included appendices containing a brief review of useful mathematical concepts (Appendix \ref{app:glossary}) as well as additional details about the datasets we use (Appendices~\ref{app:ndssl}, \ref{app:EgoFluPaths}). However, we note that in our attempt to keep the report clear and intuitive to a broader and less technical audience, we have had to sacrifice some mathematical rigor in the statements of the problems and the methods used.  Finally, we have made the code used in the many examples we consider available here: \url{https://github.com/RANDCorporation/dgmnet}.

\section{Graph Neural Networks \label{sec:graphNN}}
Graph Neural Networks (GNNs) are a powerful class of models that allow deep neural networks to be properly applied to data with a network structure.\footnote{To head off any potential confusion, we pause to note that the `network' in `neural network' bears no relation to the data `network'.} GNNs were initially introduced in 2008 \cite{scarselli2008graph}, and in recent years there has been significant interest in these models. Many variants and extensions of the original GNN model have been developed, and the term GNN has now come to refer to this large class of models, rather than the initial model \cite{scarselli2008graph}. We will not attempt to cover every GNN variant here; for comprehensive reviews on these models and their applications to network data, see \cite{bronstein2017geometric, battaglia2018relational, zhang2018deep, zhang2018network, zhou2018graph, wu2019comprehensive}. 
Our goal in this section is to review two popular network architectures, Graph Convolutional Networks \cite{kipf2016semi} and GraphSAGE \cite{hamilton2017inductive}, which we then use in Section \ref{sec:predictive} and Section \ref{sec:generative} to solve predictive and generative problems for network data, respectively. First however, we review the Multi-Layer Perceptron, a feed-forward neural network that acts on regular (i.e. non-network data), since GNNs may be thought of as extending MLPs to network-structured data. We have attempted to keep the presentation as self-contained as possible, and have included a brief review of the graph theoretical concepts utilized here in Appendix~\ref{app:glossary}. Only a basic understanding of machine learning will be required, but interested readers may wish to consult  \cite{goodfellow2016deep, nielsen2015neural} for additional information.

\subsection{Multi-Layer Perceptrons}
The Multi-Layer Perceptron (MLP) is the simplest neural network and serves as a natural starting point for the extension to graph neural networks. MLPs are composed of multiple simpler functions, often referred to as layers. Each layer consists of a linear (affine) transformation followed by a pointwise non-linearity. By composing many such layers together, these networks become ``deep'' and they become capable of approximating a large class of functions, provided that they are trained successfully. Letting $\bm{x}_i$ denote the model input, 
an MLP with $L$ hidden layers may be expressed as:
\begin{align}
    \label{eq:MLP}
    \bm{z}^{(1)}_i &= g^{(1)}\left( \bm{W}^{(0)} \bm{x}_i + \bm{b}^{(0)} \right) \,, \\
    \bm{z}^{(2)}_i &= g^{(2)}\left( \bm{W}^{(1)} \bm{z}^{(1)}_i + \bm{b}^{(1)} \right) \,, \nonumber \\
    ... \nonumber \\
    \bm{z}^{(L+1)} &= g^{(L)}\left( \bm{W}^{(L)} \bm{z}^{(L)}_i + \bm{b}^{(L)} \right) \,, \nonumber 
\end{align}
where $g^{(\ell)}$ is a non-linear activation function, and $\bm{W}^{(\ell)}, \bm{b}^{(\ell)}$ are the weight matrix and bias vector at layer $\ell$. Some common choices of activation function are the logistic sigmoid function
$\sigma(x) = (1 + e^{-x})^{-1}$, the hyperbolic tangent $\tanh(x)$, or the Rectified Linear Unit (ReLU) $\max(x,0)$. The dimension of the output at layer $\ell$ is $F_{\ell}$, with $F_0$ the dimension of the input, and $F_{L+1}$ the dimension of the final output. The intermediate dimensions, $F_{\ell}$ for $\ell = 1, ..., L$, are also known as the number of hidden units. Finally, the dimensions of the weight and bias matrices are $\bm{W}^{(\ell)} \in \mathbb{R}^{F_{\ell+1} \times F_{\ell}}$, and $\bm{b}^{(\ell)} \in \mathbb{R}^{F_{\ell+1}}$. A more succinct way of representing Eq.~\ref{eq:MLP} is
\begin{align}
    \bm{z}^{(\ell+1)}_i &= g^{(\ell)}\left( \bm{W}^{(\ell)} \bm{z}^{(\ell)}_i + \bm{b}^{(\ell)} \right) \,, \qquad \ell = 0, 1, ..., L \,,
\end{align}
with $\bm{z}_i^{(0)} = \bm{x}_i$. 

To help put MLPs on more familiar footing, when the number of hidden layers is zero, $L=0$, the dimension of the output is one, $F_{1} = 1$, and when the activation function is taken to be the logistic sigmoid, $g^{(1)}(x) = \sigma(x)$, the MLP reduces to a simple logistic regression model. In this case, Eq.~\ref{eq:MLP} becomes:
\begin{equation}
    z^{(1)}_i = \sigma\left( \bm{W}^{(0)} \bm{x}_i + \bm{b}^{(0)} \right) \,.
\end{equation}
If the MLP output is then converted to a probability via $p(\bm{x}_i) = \sigma^{-1}(z^{(1)}_i)$, then this is just the standard logistic regression relation.\footnote{In terms of generalized linear models used in statistical analysis, $\sigma^{-1}$ is known as the  link function which in this case is the logistic function.} Therefore, for classification problems, MLPs may be thought of as extending logistic regression to more complex functions with many additional trainable parameters. By adding more layers or increasing the number of hidden units within a layer, the number of parameters of the MLP is increased, and the class of functions that can be accurately modeled grows larger. In fact, MLPs are well-known to be \textit{universal approximators}, \cite{cybenko1989approximation, hornik1989multilayer}, which means that they can approximate any function arbitrarily well, given enough hidden units and/or layers.

\subsection{Graph Convolutional Networks \label{sec:gcn}}
Graph Convolutional Networks (GCNs) \cite{kipf2016semi} are a particular example of GNNs. GCNs are inspired by the success of standard convolutional neural networks (CNNs) in computer vision applications. As with most neural network models, GCNs are a general class of functions consisting of multiple layers composed together. The layers in GCNs differ from more traditional convolutional layers in that they are designed to operate on data with network structure, rather than on 2d images. The GCN layer relation is\footnote{Our presentations differs from others in that we are explicitly indicating the bias term, whereas many other papers including \cite{kipf2016semi} suppress this.}
\begin{equation}
\label{eq:GCN}
    \bm{Z}^{(\ell+1)} = g^{(\ell)} \left( \hat{\bm{A}} \bm{Z}^{(\ell)} \bm{W}^{(\ell)} + \bm{b}^{(\ell)} \right) \,.
\end{equation}
This relation requires some unpacking and notational comments. As before, $\ell = 0, ..., L$ refers to the layer. The matrix $\bm{Z}^{(\ell)} \in \mathbb{R}^{N \times F_{\ell}}$ is formed by stacking the output vectors $\bm{z}_i^{(\ell)}$ for each node in order to form a matrix. The initial activation is just taken to be the node feature matrix, $\bm{Z}^{(0)} = \bm{V}$, and the final activation $\bm{Z}^{(L+1)} = \bm{Z}$ defines the output of the model, which is often referred to as the embedding. Each layer features a point-wise non-linearity $g^{(\ell)}$, same as in the MLP. The network structure enters into Eq.~\ref{eq:GCN} via the modified adjacency matrix $\hat{\bm{A}}$, which is defined to be $\hat{\bm{A}} = \tilde{\bm{D}}^{-1/2} \tilde{\bm{A}} \tilde{\bm{D}}^{-1/2}$, where $\tilde{\bm{A}}$ is the adjacency matrix with self-loops added, i.e. $\tilde{\bm{A}} = \bm{A} + \bm{I}_N$ (where $\bm{I}_N$ is the $N\times N$ identity matrix), and lastly $\tilde{D}_{ii} = \sum_j \tilde{A}_{ij}$ is the associated degree matrix.\footnote{The theoretical motivation for using the modified adjacency matrix in Eq.~\ref{eq:GCN} (as opposed to the un-modified adjacency matrix) comes from a linear approximation to spectral convolutions defined over graphs. The benefit of the linear approximation is that the GCN layer becomes efficient to compute even as $N$ grows large, but of course this comes with a cost of reduced expressivity in terms of representing general functions defined over graphs. However, because the layers in a GCN may be stacked to form networks of arbitrary depth, this limitation may be somewhat mitigated by increasing $L$.} Finally, the $\bm{W}^{(\ell)}$ are weight matrices and the $\bm{b}^{(\ell)}$ bias vectors. The dimensions of all quantities appearing in Eq.~\ref{eq:GCN} are: $\hat{\bm{A}}: N \times N$, $\bm{Z}^{(\ell)}: N \times F_{\ell}$, $\bm{W}^{(\ell)}: F_{\ell} \times F_{\ell+1}$ and $\bm{b}^{(\ell)}: F_{\ell+1}$, with $F_0 = F$ the dimension of the node features, $F_{L+1}$ the final output dimension, and all other $F_{\ell}$ arbitrary. 

\begin{figure}[!ht]
	\centering
	\includegraphics[width=0.85\textwidth]{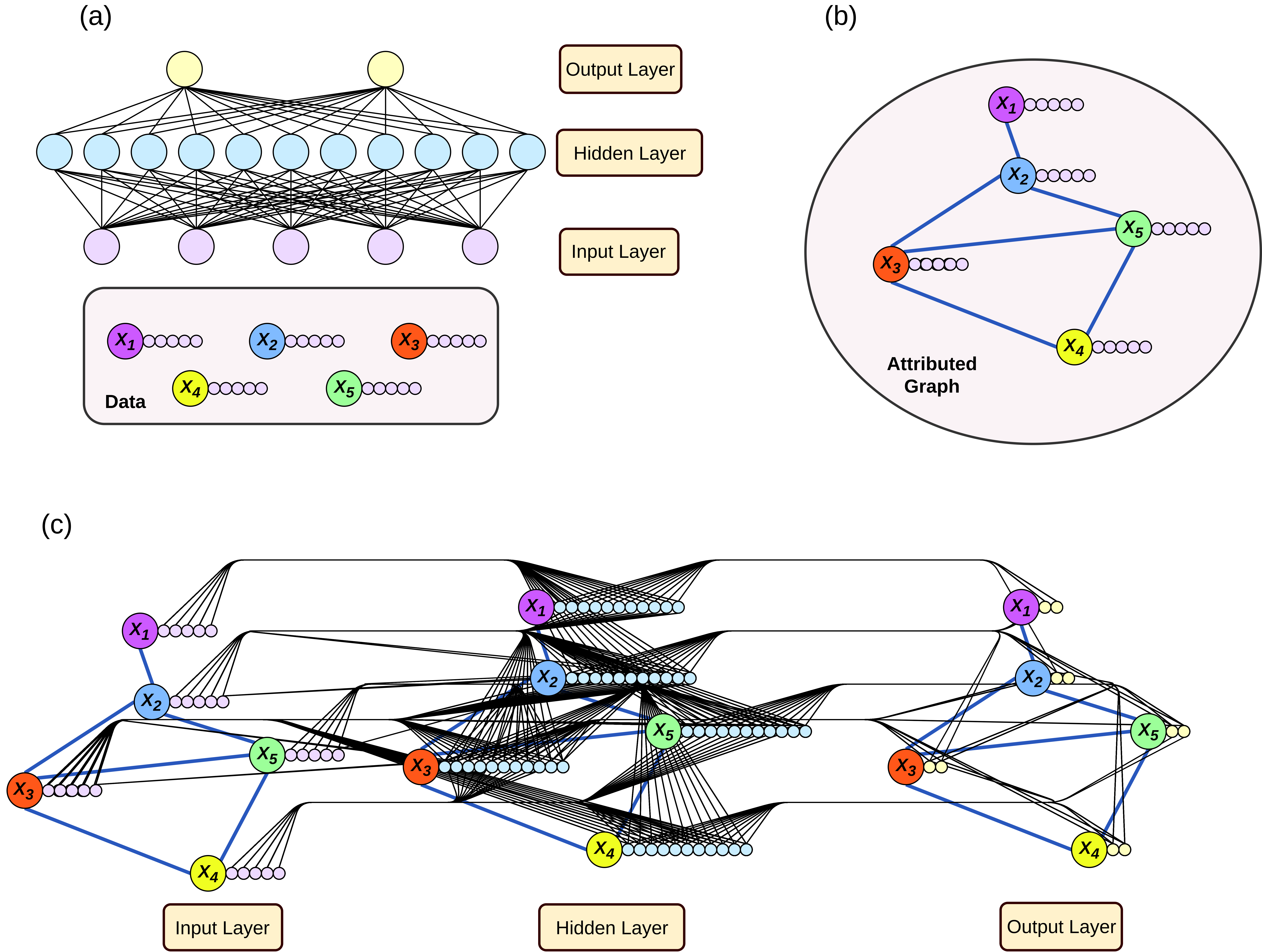}
	\caption{Comparison between MLP and GNN. (a) A standard MLP which maps 5-dimensional input data (lavender dots) to a 2-dimensional output (yellow dots) using a single hidden layer with 11 hidden units (blue dots). There are 5 distinct data entries, $\bm{x}_{i}$ with $i=1,2,3,4,5$, depicted as small dots of various colors. Each data entry should be regarded as a row in a data table with 5 columns representing different properties of each entry. (b) An attributed graph wherein the previously unrelated data points are interconnected. (c) A graph neural network such as a GCN or GraphSAGE model. The 5 data entries are no longer fed into the neural network separately; instead, in order to process node $i$ the network must aggregate the data due to node $i$'s neighborhood, $\mathcal{N}(i)$.}
	\label{fig:mlp_vs_gnn.png}
\end{figure}

To compare the GCN with the simpler MLP network, it is useful to rewrite Eq.~\ref{eq:GCN} in terms of the action on a single node, node $i$:
\begin{align}
    \label{eq:gcn2}
    \bm{z}^{(\ell+1)}_i &= g^{(\ell)}\left( \bm{W}^{(\ell)} \hat{A}_{ii} \bm{z}^{(\ell)}_{i} + \bm{W}^{(\ell)} \sum_{j \in \mathcal{N}(i)} \hat{A}_{ij} \bm{z}^{(\ell)}_{j} + \bm{b}^{(\ell)} \right) \,.
\end{align}
Here, $\bm{z}_i^{(\ell)}$ is the $i$-th row-vector in the matrix $\bm{Z}^{(\ell)}$, i.e. $\bm{z}_i^{(\ell)} = \bm{Z}^{(\ell)}_{i,:}$. The first term inside the activation function, $\bm{W}^{(\ell)} \hat{A}_{ii} \bm{z}_i^{(\ell)}$, is quite similar to the MLP expression, with the only difference being that each data point is weighted by the corresponding diagonal entry in the modified adjacency matrix. The second term, $\bm{W}^{(\ell)} \sum_{j \in \mathcal{N}(i)} \hat{A}_{ij} \bm{z}^{(\ell)}_{j}$, is the key difference between the MLP and the GCN, and it is here that the graph structure is incorporated into the neural network architecture. The notation $\mathcal{N}(i)$ denotes the neighborhood of $i$, and so this term is a weighted sum of neighboring node features. Consequently, at each layer the non-linearity acts on a linear combination of the features at node $i$ and the node features in the neighborhood of $i$, and this is done for each node in the graph. This is schematically shown in Figure \ref{fig:mlp_vs_gnn.png}.

The original application of GCNs is node classification, in which case the final layer $\bm{Z}^{(L)}$ output is used to model the class probabilities $\bm{P}$. Applied to problems with an underlying network structure, GCNs typically far outperformed MLPs because, unlike MLPs, GCNs can leverage the information of the network structure. For example, in many networks similar nodes are more likely to be connected than dissimilar nodes. This information can be useful for node classification.

\subsection{GraphSAGE \label{sec:GraphSAGE}}
GraphSAGE, introduced in \cite{hamilton2017inductive}, is a graph neural network architecture that is more expressive than the GCN architecture. The layer relation in GraphSAGE may be expressed as:
\begin{equation}
    \bm{z}^{(\ell + 1)}_i = g^{(\ell)}\left( \bm{W}_1^{(\ell)} \bm{z}^{(\ell)}_i + w_{\mathcal{N}(i)} \bm{W}_2^{(\ell)}  \sum_{j \in \mathcal{N}(i)}  \bm{z}^{(\ell)}_j + \bm{b}^{(\ell)} \right)
\end{equation}
The pre-activation consists of three terms. The first term is a linear combination of the features in the embedding vector of the same node $i$, with $\bm{W}_1^{(\ell)} \in \mathbb{R}^{F_{\ell+1} \times F_{\ell}}$ the weight matrix. The second term is computed by first summing the feature vectors of the nodes in the neighborhood of node $i$. This sum is given the weight $w_{\mathcal{N}(i)}$. For example $w_{\mathcal{N}(i)} = 1$ would correspond to a raw sum of the neighboring vectors, and $w_{\mathcal{N}(i)} = |\mathcal{N}(i)|$ would correspond to the average. A linear combination of the aggregated neighboring vectors is then taken with the weight matrix $\bm{W}_2^{(\ell)} \in \mathbb{R}^{F_{\ell+1} \times F_{\ell}}$. Lastly, the third term is just the usual bias vector $\bm{b}^{(\ell)} \in \mathbb{R}^{F_{\ell+1}}$.

Expressed in this way, it is clear that the GraphSAGE model is more expressive than GCN because it has a separate weight matrix for node $i$ and the aggregation node $i$'s neighborhood, whereas the GCN architecture ties these two terms together. This is an important property, as it allows GraphSAGE to reduce to a standard Multi-Layered Perceptron (MLP) if $\bm{W}_2^{(\ell)} = 0$. Thus, if GraphSAGE is applied to network data for which the network structure is completely irrelevant, it should be able to learn to ignore the aggregation of neighboring nodes, whereas the GCN model will be constrained to use a fixed combination of $\bm{z}_i$ and $\bm{z}_{j \in \mathcal{N}(i)}$. For convenience, the layer relations for all three neural network architectures are summarized in Table~\ref{table:GNN_summary}.
\begin{table}[!htbp]
\caption{\label{table:GNN_summary}
Neural Network Architectures}
\centering
\begin{tabular}{lll}
    \toprule
    Model & GNN & Layer Relation \\ \midrule
    \multicolumn{1}{l}{MLP} & No & $\bm{z}^{(\ell+1)}_i = g^{(\ell)} \left( \bm{W}^{(\ell)} \bm{z}^{(\ell)}_i + \bm{b}^{(\ell)} \right)$ \\
    \multicolumn{1}{l}{GCN} & Yes & $\bm{z}^{(\ell+1)}_i = g^{(\ell)}\left( \bm{W}^{(\ell)} \hat{A}_{ii} \bm{z}^{(\ell)}_{i} + \bm{W}^{(\ell)} \sum_{j \in \mathcal{N}(i)} \hat{A}_{ij} \bm{z}^{(\ell)}_{j} + \bm{b}^{(\ell)} \right)$ \\  
    \multicolumn{1}{l}{GraphSAGE} & Yes & $\bm{z}^{(\ell + 1)}_i = g^{(\ell)}\left( \bm{W}_1^{(\ell)} \bm{z}^{(\ell)}_i + w_{\mathcal{N}(i)} \bm{W}_2^{(\ell)}  \sum_{j \in \mathcal{N}(i)}  \bm{z}^{(\ell)}_j + \bm{b}^{(\ell)} \right)$ \\
    \midrule
\bottomrule
\end{tabular}
\end{table}

\subsection{Equivariance Under Graph Isomorphisms}
Unlike the MLP architecture, both the GCN and GraphSAGE neural network architectures incorporate the network structure. Moreover, these and other GNN architectures incorporate the network structure in a way that respects a very important mathematical property known as equivariance. Although equivariance is often implicit in many of the leading approaches for learning data with a network structure, we wish to explicitly (though briefly) review it here to underscore its theoretical importance.

Graph equivariance captures the simple but important idea that the ordering of the nodes in a graph is arbitrary, and any machine learning model that operates on network data should be agnostic with respect to this ordering. Essentially, this means that the ordering of the indices that label the nodes in both the node feature vector and adjacency matrix can be changed without affecting the network structure. In particular, applying a function on the network where the node indices have been reordered is the same as reordering the indices after applying the function. Stated mathematically, the transformation represented by graph neural networks should transform equivariantly under graph isomorphisms (which are discussed in Appendix \ref{app:glossary}). Specifically, under a permutation $\pi: \{1, ..., N\} \rightarrow \{1, ..., N\}$ (or node re-labeling), the node feature and adjacency matrices transform as
\begin{equation}
    V_{if} \rightarrow V_{\pi(i) f} \,, \qquad A_{ij} \rightarrow A_{\pi(i) \pi(j)} \,,
\end{equation}
A GNN layer is said to be equivariant if this transformation law applies to both the input and output of that layer, i.e. 
\begin{equation}
\bm{z}_i^{(\ell)} \rightarrow \bm{z}_{\pi(i)}^{(\ell)} \,, \qquad \bm{z}_i^{(\ell+1)} \rightarrow \bm{z}_{\pi(i)}^{(\ell+1)} \,.
\end{equation}
A GNN consisting of equivariant layers is said to be equivariant because the final output will transform the same way as the node attribute matrix under an isomorphism. Both the GCN and GraphSAGE architectures are equivariant, and as a result the learned functions will not depend on the arbitrary ordering of the nodes in the network. 

\section{Policy-Relevant Network Data \label{sec:data}}
As alluded to in the Introduction, there are many different kinds of networks that are relevant for public policy research. Guided by a desire to improve agent-based modeling efforts, in this report we exclusively consider networks where the nodes correspond to individuals and the edges correspond to some kind of tie or interaction between those individuals. In particular, we focus on two concrete network datasets, a sociocentric network and an egocentric network. A sociocentric network captures the entire population of interest. Each node is also a research subject; all ties between subjects are represented in the network. Sociocentric network data is typically broad and shallow - broad in the sense that they capture the ties of the entire population and shallow in the sense that they often contain just a few key attributes describing individuals and their relationships. Node attributes often are limited to demographic variables, and they often do not contain edge attributes besides a weight describing the strength of the relationship. In contrast, egocentric  networks (also known as personal networks) are a collection of local networks each centered around an individual of interest known as the ego. These local networks only capture the immediate neighbors of the ego (known as alters), and as such they only describe the personal communities of the egos rather than the entire network of the full population. Hence, in contrast to sociocentric networks, egocentric may be described as narrow and deep - narrow because they only capture the immediate neighborhood of the ego nodes and deep because they often contain a very rich set of node attributes. This is because egocentric data is created by surveying real people, and it is of course a simpler matter to collected detailed survey responses from a small number of people than it is to reach every single person in a population. Egocentric data is also often collected as part of a longitudinal study and can thus contain information on how the node and edge attributes change over time. However, since the research subjects in these networks are the egos themselves, the node and edge attributes of the alters are those reported by the ego. Hence, for example, attributes representing preference and behaviors of the alters are interpretations or views as given by the ego respondents and not by the alters.  

In the following, we introduce a concrete example of each type of network dataset, both of which we frequently refer to throughout the rest of the report. These datasets are being used to inform an agent-based model currently under development at RAND that is aimed at improving our understanding of the dynamic interplay between social and epidemiological drivers of influenza transmission and the effectiveness of controlling influenza outbreaks through vaccination and vaccination policies. The model builds upon existing literature and considers the heterogeneity in how people interact over complex social contact networks affecting disease transmission chains \cite{Volz2007,Meyers2005,Salathe2010,Barrat2008,Barthelemy2004,Pastor-Satorras2001}. In this agent-based model, an agent's personal and social-network experiences from past influenza seasons affect current decisions to get vaccinated and thus influence the course of an epidemic. In particular, the model considers how influenza vaccination behaviors and hesitancy can spread over social networks, further affecting disease transmission dynamics~\cite{Fu2011,Cornforth2011,Bhattacharyya2013,Xia2014,vardavasModelingInfluenzaVaccination2013}.

\subsection{The NDSSL Sociocentric Dataset}
Large-scale social contact networks represent the interactions within a community or population. By `large-scale' we generally mean networks consisting of 100,000 or more individuals, although we will avoid using a precise cut-off for this characterization. A large-scale freely available sociocentric network dataset describing a synthetic population representative of Portland, Oregon, was developed as part of the transportation analysis and simulation system (TRANSIMS) project by the Network Dynamics and Simulation Science Laboratory (NDSSL) group at the Biocomplexity Institute of Virginia Tech \cite{maratheSyntheticDataProducts2014}. The data provides a snapshot of a time-varying social contact network for a typical day between all individuals in the city, with detailed geographic information regarding the location, duration, and purpose of their interactions derived from transportation data. It should be noted that the data was artificially generated through an agent-based model simulation, and therefore does not describe real individuals. Often real data is preferable to synthetic data, although one benefit of artificial data is that there are no restrictions due to privacy concerns. This dataset also encodes any interactions between individuals as they go about their daily activities. Synthetic network data produced by the NDSSL group, including that for Portland has been used to inform transmission dynamics models of an infectious disease~\cite{bissetIndemicsInteractiveHighPerformance2014, venkatramananUsingDatadrivenAgentbased2018}. Together with other datasets, this data has also recently been used by RAND researchers to inform an SEIR model and web-based tool of COVID-19 that compares the effects of different nonpharmaceutical public health interventions (NPIs) on health and economic outcomes of each U.S. state~\cite{vardavasHealthEconomicImpacts2020}.\footnote{The web-based tool is available here: \url{https://www.rand.org/pubs/tools/TLA173-1/tool.html}.} This data is also being used to inform an agent-based model of influenza vaccination behaviors.

\begin{figure}[!ht]
\begin{center}
\includegraphics[width=0.4\textwidth]{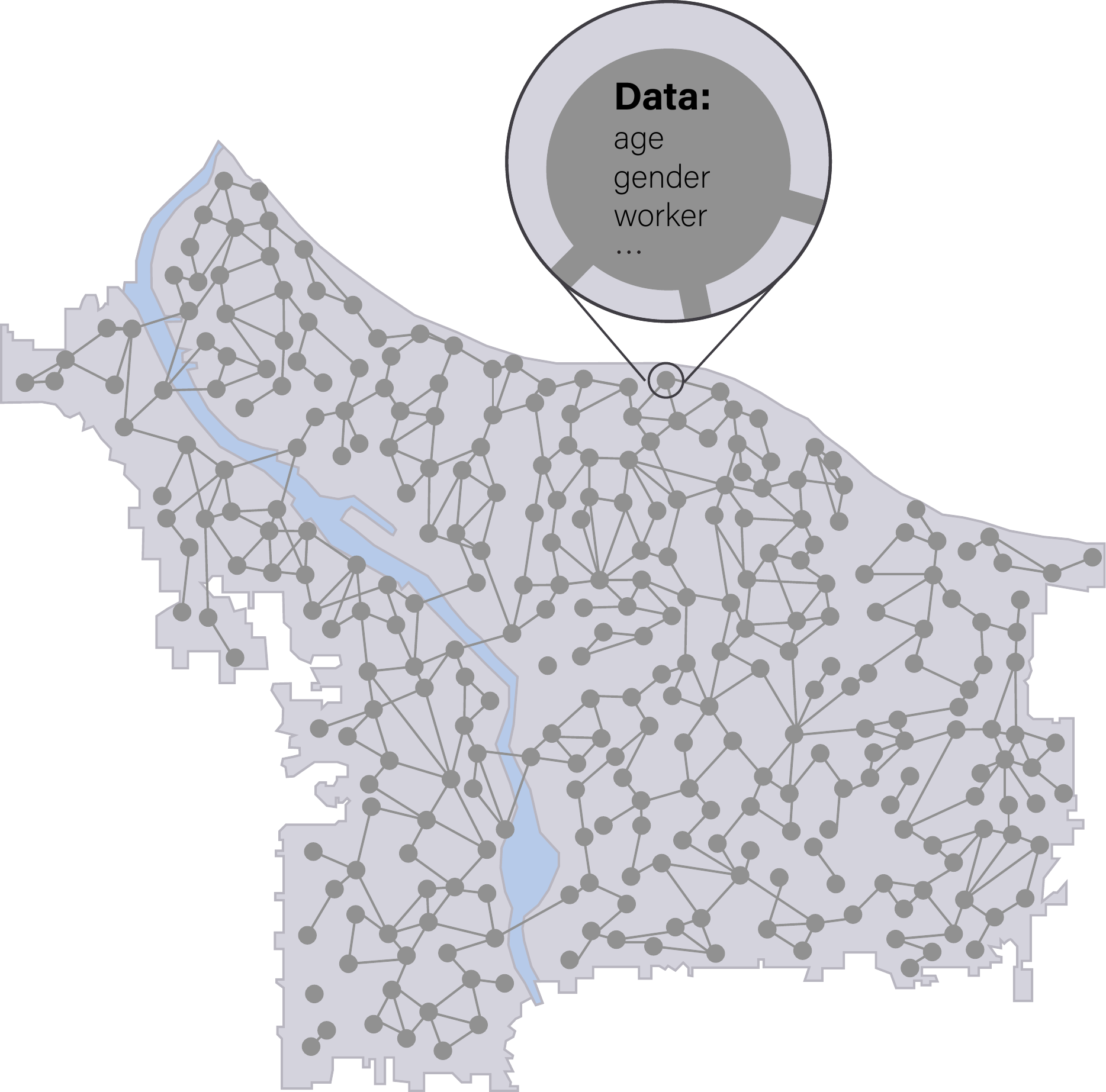}
\end{center}
\caption{Illustration of the NDSSL network for Portland, OR. Nodes represent individuals and each belonging to a household. Individuals move and interact with other individuals during a day, and this creates an edge list of interactions. Each individual has various demographic attributes, including age, gender, household income, and household location. 
\label{Fig:ill-NDSSL-Portland}}
\end{figure}

\begin{figure}[!ht]
	\centering
	\includegraphics[width=0.85\linewidth]{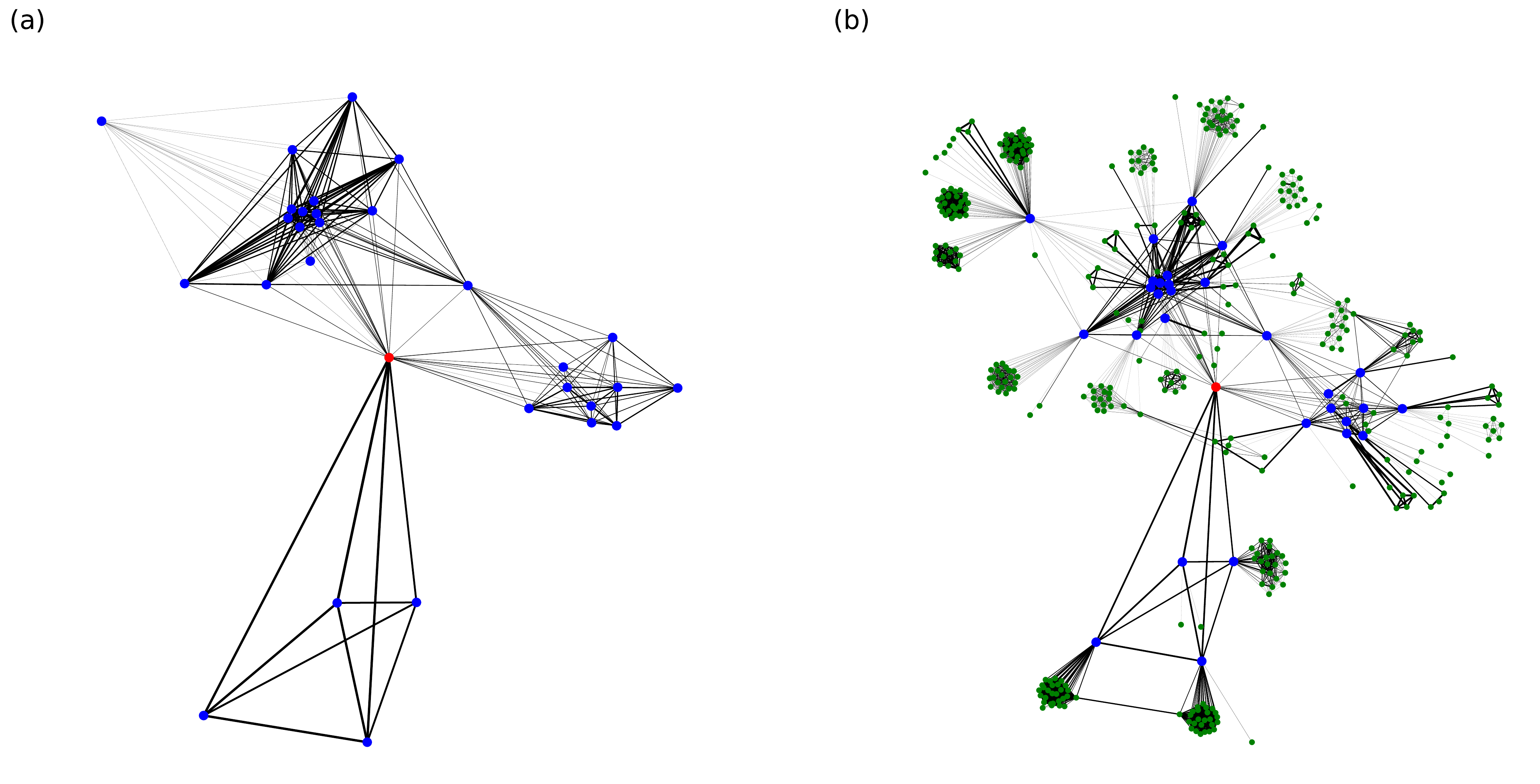}
	\caption{Neighborhood of a randomly chosen node in the NDSSL dataset. (a): The 1-neighborhood induced subgraph around a randomly chosen root node (shown in red). The thickness of the edges corresponds to the duration of the contact. (b): The 2-neighborhood induced subgraph around the same root node, which includes the distance 1 nodes (shown in blue) as well as the distance 2 nodes (shown in green).}
	\label{fig:NDSSL_egograph}
\end{figure}

The data from the TRANSIM simulation may be used to construct a large temporal bipartite network connecting people to locations. For our purposes, we are mainly interested in the person-to-person connectivity represented by the data. Therefore, we considered the sociocentric network formed by connecting two individuals if they had some in-person interaction throughout the day. Each interaction has an associated duration and set of activities that each individual was engaged in when the interaction occurred (for example, an interaction could have occurred when one person was shopping and another was at work). Additionally, for each individual multiple demographic variables are available, including variables at the individual level and at the household level. We refer to this person-to-person sociocentric data as the NDSSL data. Overall, this network contains about $N = 1.7 \times 10^6$ individuals, organized into $6.3 \times 10^5$ households, as well as $M = 2.0 \times 10^7$ distinct interactions. We would like to gain some intuition for the network structure by visualizing the graph. However, with $10^6$ nodes this graph is far too large to display directly. Figure \ref{Fig:ill-NDSSL-Portland} provides a schematic illustration of the distribution of households throughout the city of Portland. To give a sense of the neighborhood around a typical node, in Figure \ref{fig:NDSSL_egograph} we show the neighborhood graph for a randomly chosen root node, both for immediate neighbors (a) and neighbors that are either 1- or 2-edges away from the root node (b). The nodes in this graph are organized into many groups or clusters. For example, the lower cluster in the left panel of Figure \ref{fig:NDSSL_egograph} corresponds to the household of the root node, and the other clusters represent workplace interactions.

As a network dataset, the NDSSL data contains information about individuals as well as about the pairwise interactions between individuals. In order to apply machine learning algorithms to this data, it will be important to put the data into a standardized format (described in more detail in Appendix \ref{app:glossary}). For this purpose, it is useful to organize the data fields into two groups: node-level data and edge-level data. These are shown in Table \ref{table:ndssl_features}. 
\begin{table}[!h]
    \caption{\label{table:ndssl_features} The node and edge features for the NDSSL data. }
\rowcolors{1}{blue!30}{blue!10}
    \centering
    \begin{tabular}{|l|l|}
        \hline
        \textbf{node features} & \textbf{edge features} \\
        \hline 
        person id & person1 id \\
        household id & person2 id \\
        age & person activity1 \\         
        gender & person activity2 \\         
        worker &  weight\\         
        relationship &  \\         
        household income &  \\ 
        household size &  \\         
        zipcode &  \\         
        household vehicles &  \\     
        household workers &  \\      
        \hline         
    \end{tabular}
    \caption*{}
\end{table}
Each person is assigned a unique id (person id), and each person belongs to a unique household (household id). The node features then consist of a combination of individual-level and household-level data. Here, the term `worker' refers to whether the individual is employed or not, and `household workers' counts the number of employed household members. The `relationship' field refers to the relationship of the individual to the household.  The edge features describe an interaction between two individuals, each of whom was engaged in some activity prior to the interaction. The list of activities is: home, work, shop, visit, social/recreation, other, pick up or drop off a passenger, school, college.  The node-level data can be assembled into a $N \times F_V$ matrix $\bm{V}$ we call the node features, and the edge-level data can be assembled into a $M \times F_E$ edge feature matrix $\bm{E}$. Here, $F_V$ and $F_E$ are the number of features (or data fields) for each group. More details about the dataset and its statistical properties are provided in Appendix~\ref{app:ndssl}, and the code used for this analysis can be found \href{https://code.rand.org/hartnett/dgmnet/-/blob/master/NDSSL\%20Data\%20Processing\%20Notebook.ipynb}{here}.

\subsection{The FluPaths Egocentric Dataset \label{sec:EgoFluPaths}}
In addition to large-scale sociocentric contact networks, we also consider so-called egocentric datasets. These can be considered as a set of subgraphs sampled from some unobserved larger sociocentric network. Each of the sampled subgraphs is focused around a particular individual referred to as the ego. The sampling procedure returns the neighborhood of each ego, as well as any connections between those neighbors. Larger egocentric networks may be constructed by additionally considering the 2nd-degree neighborhood, the 3rd-degree neighborhood, and so on. We refer to these as $k$-degree egocentric networks and they are simply the induced subgraphs formed from the $k$-neighborhood of a given ego node. Just as with the social contact networks described above, these networks will contain attribute information about the nodes and their connections, in addition to the graph of connectivities. 

\begin{figure}[!ht]
	\centering
	\includegraphics[width=0.58\linewidth]{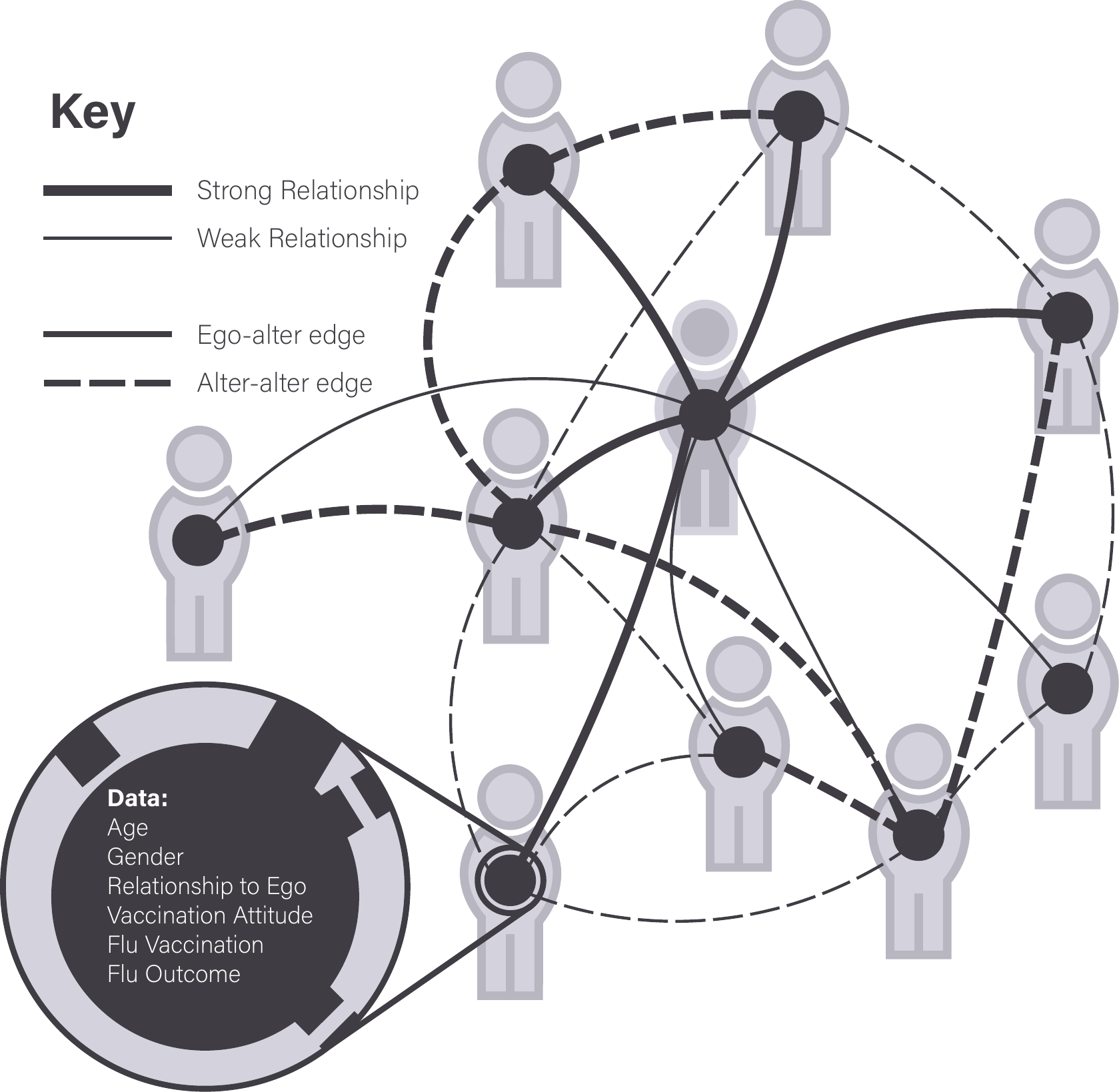}
	\caption{An illustration of the egocentric network of one of the respondents in the wave 1 FluPaths data}
	\label{Fig:FluPathEgoIllustration}
\end{figure}

As our example egocentric dataset, we consider the FluPaths dataset that was collected from a four-year longitudinal survey on influenza fielded from fall 2016 to spring 2020 carried out through the RAND American Life Panel (ALP) \cite{AmericanLifePanel}. FluPaths was collected for the purpose of informing the behavioral structures and parameters of an agent-based simulation model describing influenza vaccination behavior. Respondents were asked eight waves of questions throughout the duration of the study. Core pre-flu-season questions collected the respondent’s intentions to vaccinate, risk perceptions of catching the flu, and whether they received a recommendation to vaccinate by a health care professional. Core post-flu-season questions collected whether respondents were vaccinated, whether this resulted from a recommendation by a health care professional, whether they thought they caught the flu, were tested for it, and were prescribed antiviral medication. FluPaths surveys collected detailed information on each respondent’s social network structure (including alter-alter ties) and assessed network influence on the respondent’s risk perceptions and attitudes regarding influenza and vaccination over time. Wave 1 network data collection was carried out using EgoWeb, a public-source social network data collection tool~\cite{Green2011, Kennedy2015}. Each respondent in the FluPaths study was asked to identify up to 15 individuals with whom respondents have regular contact and elicit characteristics of each alter.  The first set of characteristics elicited pertain to demographics and the strength of relationships. This was followed by a set of elicitations describing perceptions of flu and vaccination experiences of their alters.  Subsequent pre- and post-season surveys included an abbreviated version that allowed respondents to update the list of alters and elicited flu and vaccination experiences. The data collection approach focused on the most active and salient alters in an individual’s social network~\cite{Roberts2008, Stiller2007}. Figure~\ref{Fig:FluPathEgoIllustration} provides an illustration of the egocentric networks in the wave 1 FluPaths data of a respondent's egocentric network.
Tables~\ref{table:FluPaths_Node_features} and~\ref{table:FluPaths_Edge_features} respectively provide the node features or attributes and the edge list contained in the network data structure. Data associated with each respondent (i.e., the ego) goes beyond that contained in the network data structure. The characteristics and features showed in Table~\ref{table:FluPaths_Node_features} for the respondents include data fields that are common to the features elicited for the alters. The data shown for the alters is provided by the respondent and may not be accurate or complete.  
\begin{table}[!htbp]
\rowcolors{1}{blue!30}{blue!10}
\footnotesize
\centering
\caption{\label{table:FluPaths_Node_features} Example entries of the node feature table of the wave 1 FluPaths egocentric data.  }
\begin{tabular}{llllllllr}
  \hline
\thead{Node \\id}\hspace{4pt} & \thead{Age Group} &\thead{Gender} & \thead{Relationship \\to Ego}\hspace{14pt} & \thead{Medical\\ Worker}\hspace{8pt} & \thead{Vaccination\\ Attitude} & \thead{Vaccinated\\ for the Flu} & \thead{Contracted\\ Flu} & \thead{Network\\ id} \\ 
  \hline
100 & 26-35 & Female & Ego & No & Neutral & Yes & Unknown & 1 \\ 
  108 & 26-35 & Female & Friend & No & Positive & No & No & 1 \\ 
  103 & 56-65 & Male & Family member & No & Positive & No & No & 1 \\ 
  104 & 26-35 & Female & Family member & No & Positive & Yes & Yes & 1 \\ 
  107 & Unknown & Female & Friend & No & Positive & No & No & 1 \\ 
  102 & 66+ & Female & Family member & No & Positive & No & No & 1 \\ 
  101 & 26-35 & Male & Spouse & No & Positive & No & No & 1 \\ 
  105 & 26-35 & Male & Family member & No & Positive & No & No & 1 \\ 
  109 & 26-35 & Female & Friend & No & Positive & No & No & 1 \\ 
  110 & Unknown & Male & Friend & No & Positive & No & No & 1 \\ 
  106 & 46-55 & Female & Friend & No & Positive & Unknown & Unknown & 1 \\ 
  200 & 46-55 & Female & Ego & No & Neutral & Unknown & Unknown & 2 \\ 
  211 & 56-65 & Female & Family member & Unknown & Neutral & Unknown & Unknown & 2 \\ 
  214 & 46-55 & Female & Family member & Unknown & Neutral & Unknown & Unknown & 2 \\ 
  208 & 18-25 & Male & Family member & Unknown & Neutral & Unknown & Unknown & 2 \\ 
   \hline
\end{tabular}
\end{table}
Respondents or egos in the datasets can be identified by the Node id variable. The last two digits of the Node id of the egos are always zeros.  The relationship to ego field can be used to identify entries that represent the egos (i.e., the respondents). The relationship of the alter to the ego also includes spouse, family member, friend, service provider, and others. The primary characteristics elicited were demographics and include the gender and age of the respondents and the alters. In addition, respondents were asked whether their occupation and those of their alters is in the medical field, such as being a physician, nurse, health care worker, and more generally anyone working in a field associated with health. Behavioral features include (i) the respondent's attitude and what they perceive is the attitude of their alters towards vaccines and getting vaccinated, (ii) whether they and their alters were vaccinated for the flu, and (iii)  and whether they and their alters caught the flu or think had the flu in the previous season.
\begin{table}[!ht]
\rowcolors{1}{blue!30}{blue!10}
\footnotesize
\centering
\caption{\label{table:FluPaths_Edge_features} Example entries of the edge list of the wave 1 FluPaths egocentric data.}
\begin{tabular}{lllll}
  \hline
\thead{From\\  Node id} & \thead{To\\  Node id} & \thead{Edge Type  \& Strength}\hspace{0pt}& \thead{Face-to-Face\\ Interaction\\ Frequency} & \thead{non-Face-to-Face\\ Interaction\\ Frequency} \\ 
  \hline
100 & 108 & Ego-Alter & Weekly & Weekly \\ 
  100 & 103 & Ego-Alter & Daily & Daily \\ 
  100 & 104 & Ego-Alter & Weekly & Weekly \\ 
  108 & 107 & Alter-Alter Weak&  &  \\ 
  104 & 102 & Alter-Alter Strong&  &  \\ 
  104 & 101 & Alter-Alter Weak&  &  \\ 
  101 & 105 & Alter-Alter Weak&  &  \\ 
  108 & 101 & Alter-Alter Weak&  &  \\ 
  100 & 109 & Ego-Alter & Weekly & Monthly \\ 
  100 & 110 & Ego-Alter & Weekly & Monthly \\ 
  108 & 106 & Alter-Alter Strong&  &  \\ 
  102 & 101 & Alter-Alter Strong&  &  \\ 
  109 & 110 & Alter-Alter Strong&  &  \\ 
  101 & 109 & Alter-Alter Strong&  &  \\ 
  101 & 106 & Alter-Alter Strong&  &  \\ 
   \hline
\end{tabular}
\end{table}

The edge list includes the type of connection that describes whether the relationship is an ego-alter tie or is an alter-alter tie. For the latter, respondents were asked to identify if the tie between the alters could be considered a strong or a weak tie. Respondents were also asked about the frequency of their face-to-face and non-face-to-face interactions with each alter. 

Our FluPath Wave 1 EgoWeb data includes 1,898 respondents (egos) and 19,250 alters,  and contains 82,300 edges of which 62,050 are alter-alter ties. Of the alter-alter ties, 26,496 were considered strong ties by the respondent.  Table~\ref{table:FluPaths_degree_summary} provides the summary statistics of the degree distribution of our egocentric networks. The summary statistic shown in each column respectively including only (i) ego-alter ties, (ii) frequent ego-alter ties  (iii) alter-alter ties, (iv) strong alter-alter ties, and (v) weak alter-alter ties.

\begin{table}[!htbp]
\rowcolors{1}{blue!30}{blue!10}
\centering
\caption{\label{table:FluPaths_degree_summary} Summary statistics of the degree distribution of the wave 1 FluPaths egocentric data.}
\footnotesize
\begin{tabular}{rrrrrr}
  \hline
 & Ego-Alter & Frequent-Ego-Alter & Alter-Alter & Strong & Weak \\ 
  \hline
Min. & 2.00 & 1.00 & 1.00 & 1.00 & 1.00 \\ 
  1st Qu. & 6.00 & 3.00 & 4.00 & 2.00 & 2.00 \\ 
  Median & 10.00 & 5.00 & 7.00 & 3.00 & 4.00 \\ 
  Mean & 10.14 & 6.44 & 7.45 & 3.83 & 4.55 \\ 
  3rd Qu. & 15.00 & 10.00 & 10.00 & 5.00 & 6.00 \\ 
  Max. & 15.00 & 15.00 & 15.00 & 14.00 & 14.00 \\ 
   \hline
\end{tabular}
\end{table}

\begin{figure}[!ht]
\centering
\begin{subfigure}
  \centering
  \includegraphics[width=0.5\linewidth]{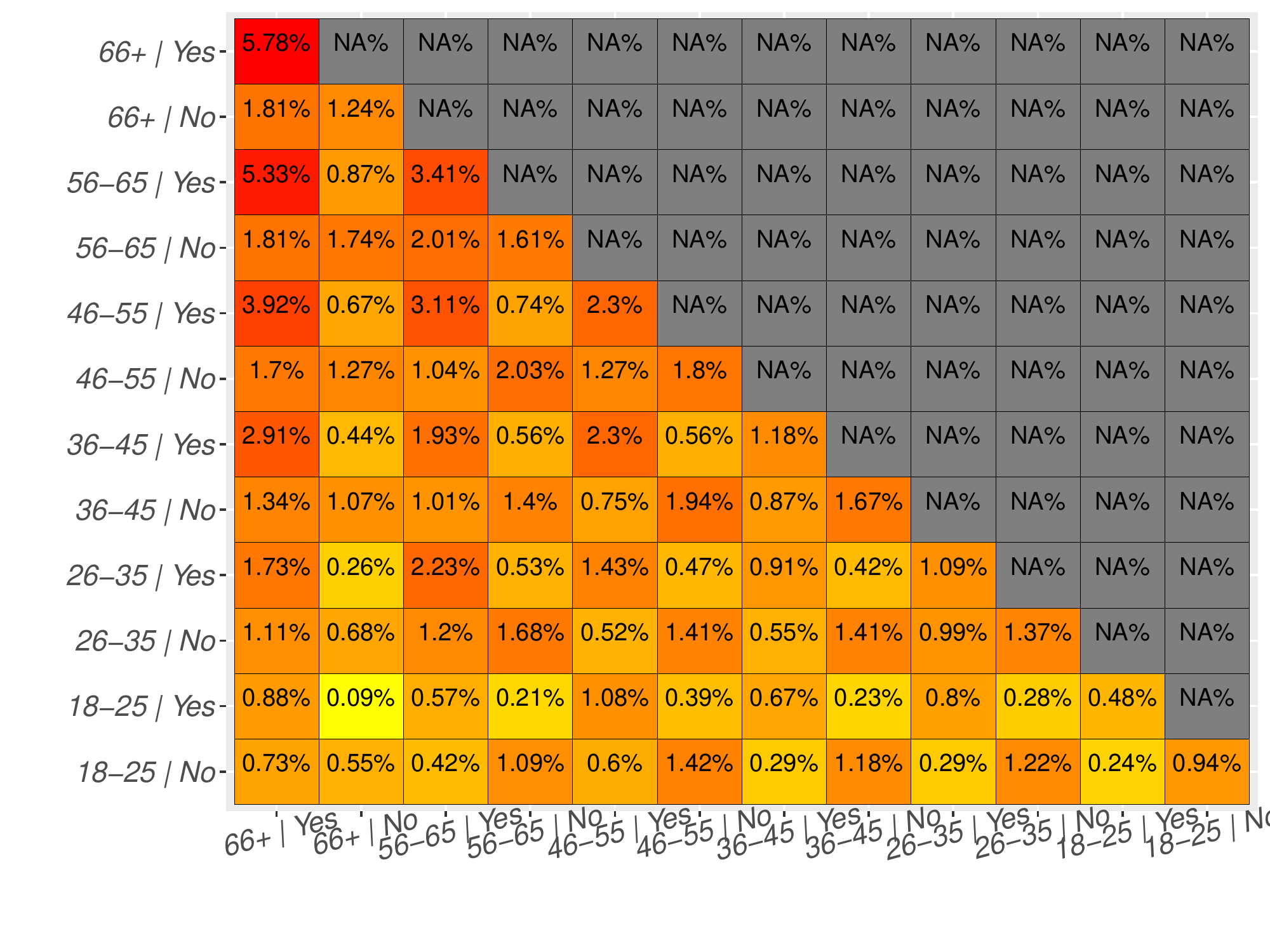}
\put(-220,190){(a)}
\end{subfigure}%
\hfill
\begin{subfigure}
  \centering
  \includegraphics[width=0.5\linewidth]{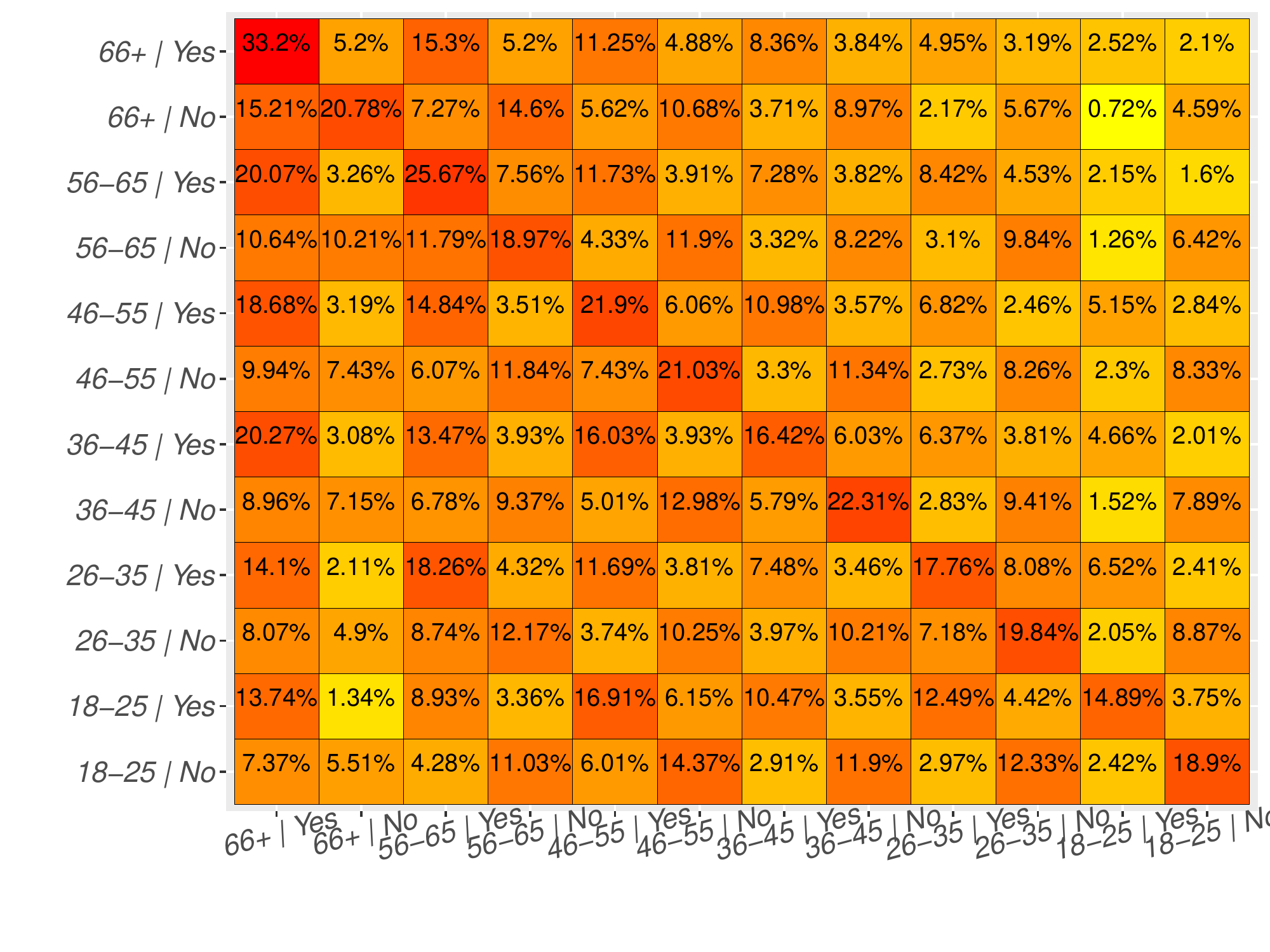}
\put(-220,190){(b)}
\end{subfigure}%
\caption{Example mixing matrices generated using the wave 1 FluPath egocentric network data and  for two node features: age group and influenza vaccination status by the end of fall 2016. The first matrix (a) shows the percentage of edges connecting nodes with different node features. As explained in the text, each row in the second matrix (b) shows the  distribution of the alters of nodes with different node features. The color scale ranges from yellow to red to indicate the frequency of the interaction type.     
\label{Fig:EgoMixMatAgeVacc}}
\end{figure}

An other important statistic generated by this data that is relevant for this project is the mixing matrix. The mixing matrix gives a tabulation of the frequency of edges in the network based on the node features connected by the network edges.  Figure~\ref{Fig:EgoMixMatAgeVacc} provides examples of two types of mixing matrices that can be generated by analyzing the FluPath egocentric data. The first matrix provides the frequency of edges connecting nodes (both ego and alters) based on the age group feature and the influenza vaccination decision feature of each node. The elements of this matrix sum to 100 percent. Since the network is non-directional, one way of representing this is using a symmetric matrix. However, as shown in Figure~\ref{Fig:EgoMixMatAgeVacc}a, a better way of representing the matrix is by using a triangular matrix. This is because edge frequencies can be compared against each other more efficiently.  
A problem of this matrix representation is that it suffers from biases in how representative the FluPath survey is of the population of interest. The ALP strives to provide a nationally representative sample and provide sample weights of the respondents to reweigh statistics to help researchers generate statistics that are nationally representative. Therefore, we can use the respondents sampling weights to reweigh the edges and generate a representative mixing matrix. An alternative approach is to provide a conditional mixing matrix shown in Figure~\ref{Fig:EgoMixMatAgeVacc}b. Here each row sums to 100 percent. This mixing matrix provides the distribution of edges connecting nodes with a fixed set of feature values with nodes with a different set of feature values.  Generating this matrix does not require the sampling weights and can be considered to be representative of the population we describe. However, what is lost is the information on the distribution of people belonging to groups with different node features (i.e., population strata).  Note that for ease of presentation, the mixing matrices shown in Figure~\ref{Fig:EgoMixMatAgeVacc} consider the subset of the data where alters of unknown vaccination status have been removed. More details about the dataset and its statistical properties are provided in Appendix~\ref{app:EgoFluPaths}.

\section{Predictive Modeling Over Networks \label{sec:predictive}}
As mentioned in the Introduction, recent years have seen significant progress in applying deep learning to network or graphical data. Much of these recent advances have been motivated by improving performance on prediction problems. Although our main focus in this report is on generative problems, in this section we review some common graph prediction problems and show how the GNN models introduced in the previous section can be used to address these. Graph prediction problems can be classified by the nature of the dependent variable that is being modeled. This variable could exist at the node-level, edge-level, or the level of an entire graph. As we now show, graph neural networks may be used to address each type of problem, although how they are employed depends on the problem.

\subsection{Node-Level Prediction \label{sec:nodeprediction}}
The goal in node prediction problems is to predict a dependent variable that exists at the node level. In general, the node features $\bm{V}$ are split into a set of independent variables and a set of dependent variables. For concreteness, we restrict our attention to node classification where the dependent variable is a single categorical variable for each node, which we denote as $\bm{y}$. Also, use $\bm{V}'$ to denote the modified node feature matrix with $\bm{y}$ removed, (hence $\bm{V}'$ is the vector of independent node feature variables) and furthermore, take the overall number of classes to be $C$ and use one-hot encoding, such that $\bm{y}$ is an $N \times C$ matrix with
\begin{equation}
    y_{i c} = 
    \begin{cases}
    1 \,, \quad \text{node $i$ is class $c$} \\
    0 \,, \quad \text{otherwise.}
    \end{cases}
\end{equation}
The goal is to then to use neural networks to model $P_{ic}$, the probability that node $i$ will belong to class $c$. 

This problem can be broken into two steps. First, the probability is parametrized using a neural network, corresponding to a choice of model architecture. For example, the probability could be modeled with an $L$-layer GNN with the number of input features $F_0$ given by the number of features in $\bm{V}'$, and the number of output features $F_{L+1} = C$, the number of classes. In order to transform the raw neural network output $\bm{z}_i^{(L+1)}$ into a probability distribution over $C$ classes, the softmax operation is used:
\begin{equation}
    P_{ic} = \text{softmax}(\bm{z}_{i}^{(L+1)})_c = \frac{}{} \frac{\exp\left( z_{ic}^{(L+1)} \right)}{\sum_{c' = 1}^C \exp\left( z_{ic'}^{(L+1)} \right)} \,.
\end{equation}
The prediction is then the class with the largest predicted probability, i.e. ${\hat{c} = \argmax_c P_{ic}}$. 

The second step is to train or fit the weight parameters by minimizing some objective function on a training subset of the data. For classification problems, the natural choice is to use the cross-entropy loss function on a subset of the data:
\begin{equation}
    \mathcal{L}(\bm{\theta}) = - \sum_{i \in V_{\text{train}}} \sum_{c = 1}^C y_{i c} \ln P_{i c} \,,
\end{equation}
where $V_{\text{train}}$ is the set of node indices in the training set, with $|V_{\text{train}}| = N_{\text{train}}$. The optimization may be conveniently carried out using deep learning software libraries (such as PyTorch or TensorFlow) to implement first order (i.e. gradient-based) optimization algorithms such as stochastic gradient descent.

To illustrate this node classification problem on an example using the NDSSL dataset, suppose that the data were incomplete, and that there were some individuals whose employment status (the ``worker'' variable) is missing. Graph neural networks can be used to learn a predictive model that will allow us to accurately predict the missing values. This problem is therefore an instance of the \textbf{Data Imputation} aim discussed in Section \ref{sec:aims}. The code used for this example is available \href{https://code.rand.org/hartnett/dgmnet/-/blob/master/Node\%20Classification.ipynb}{here}.

As a first step, we split the NDSSL data into a training and testing set. Unfortunately, there are a number of subtleties and technical complications associated with this step. We performed an 80/20 split of the 1.6 million nodes in the NDSSL dataset. One immediate complication is that this split affects the nodes and edges of the graph differently. There are now three kinds of edges: train-train, train-test, and test-test. For the NDSSL dataset, the 80/20 split amongst the nodes resulted in the distribution of edge types shown in Table~\ref{table:edge_types}. The first important point to note is that the fraction of edges between nodes in the train set does not match the fraction of nodes in the train set (64.7 percent versus 80 percent), and similarly for the test set (4.7 percent versus 20 percent). In other words, the act of splitting up the data this way has caused a distributional shift between the distribution of the full graph, the training distribution, and the test distribution. In particular, each set will feature a different edge density. The second observation is that a sizable fraction of edges (30.6 percent) are connections between train/test nodes. Our approach discarded these edges, and they were not used during the training or the testing. Presumably they could be incorporated in the testing stage, but we did not pursue this.
\begin{table}[!htbp]
\caption{\label{table:edge_types}Train/Test Edge Distribution}
\centering
\begin{tabular}{ccc}
\toprule
    train-train & train-test & test-test \\ \midrule
    64.7\% & 30.6\% & 4.7\% \\
\bottomrule
\end{tabular}
\end{table}

In addition to the train/test issue, a second complication concerns the breaking up of the dataset into mini-batches. This also becomes a non-trivial task once the graph structure is taken into account. To evaluate the forward pass of a GNN model on a given batch, we require both the features for nodes in the batch as well as the features for the $L$-degree neighborhood of that graph. This is because each additional layer of a GNN aggregates the features of the neighborhood of the nodes visited by the previous layer. Thus, a 3-layer GNN operating on a batch of 1000 nodes requires as input the node features of all nodes a distance 3 or less from batch nodes. In many networks the overall number of required nodes will scale exponentially, at least for moderate $L$. Indeed, this is the case in the NDSSL network; see for example Figure \ref{fig:snowball} in Appendix \ref{app:ndssl}. For many of the networks commonly used to benchmark GNN models, such as Citeseer, Cora, or Pubmed, this growth does not present too much of a problem. For example, the largest of these, Pubmed, has about 20,000 nodes and 44,000 edges, compared to the 1.6 million nodes and 19 million edges of NDSSL. Thus, there are three orders of magnitude more edges in the NDSSL network, and consequently the exponential growth of snowball samples necessitates new approaches. We chose to address this problem using the GraphSAINT sampling method \cite{zeng2019graphsaint}. GraphSAINT samples the neighborhood around a batch of nodes by performing a weighted random walk. This results in manageable batch sizes, but at the cost of introducing a sampling error at each step.

With the network data processed in this way, we then proceeded to train two GNN models, a GCN and a GraphSAGE model. For comparison, we also trained a regular MLP. In all cases we used three hidden layers with 256 units in each layer, and we optimized the weights of the neural networks using the Adam optimizer with a learning rate of $10^{-3}$. We used roughly 50 epochs, so that each node was visited about 50 times throughout training.\footnote{The GraphSAINT sampler used for the GNN models returns a variable batch size due to the random walk, so the number of nodes in each epoch is not fixed in the GNN case. As a result, one ``epoch" of GNN training does not necessarily correspond to the usual notion of epoch as being one complete pass through the data.} The results are shown below in Figure \ref{fig:node_prediction_results} and in Table.~\ref{table:employment_results}.

\begin{figure}[!ht]
	\centering
	\includegraphics[width=1.0\textwidth]{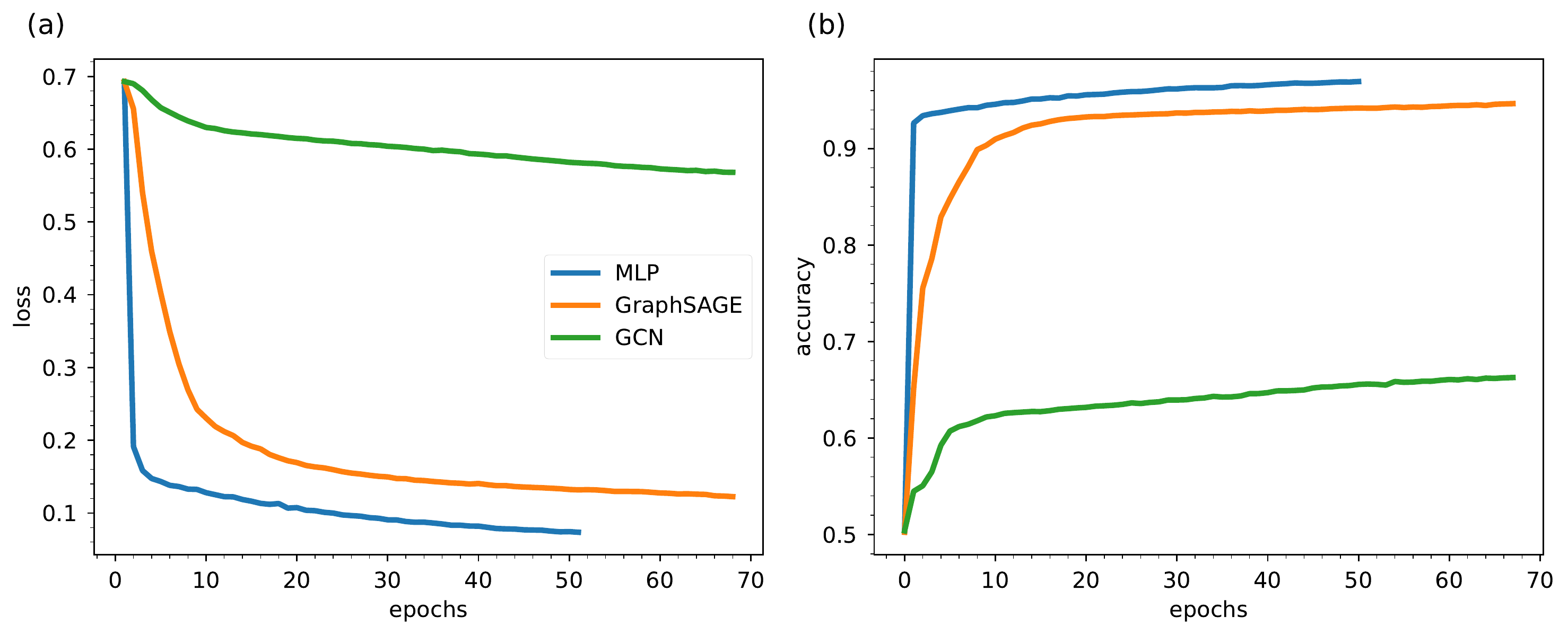}
	\caption{The node prediction results. (a) Training set loss throughout training. The GNN models were trained for 67 epochs and the MLP for 50. Due to the GraphSAINT sampler, the number of data points in an epoch differs between the GNN and MLP models, and the 67 epochs of GNN training was chosen so that the overall number of passes through the data is roughly 50. (b) The accuracy on the held-out test set after training is completed.}
	\label{fig:node_prediction_results}
\end{figure}

\begin{table}[!htbp]
\caption{\label{table:employment_results}NDSSL Node Classification: Employment Status}
\centering
\begin{tabular}{lllllll}
    \toprule
    Model & Layers & Hidden Units & Parameters & Test Loss & Test Accuracy & Test AUC \\ \midrule
    \multicolumn{1}{l}{MLP} & 3 & 256 & 168,450 & 0.0794 & 96.6 \% & 0.966  \\
    \multicolumn{1}{l}{GraphSAGE} & 3 & 256 & 335,618  & 0.1479 & 93.6 \% & 0.936 \\    
    \multicolumn{1}{l}{GCN} & 3 & 256 & 168,450 & 0.5742 & 65.9 \% & 0.659  \\
    \midrule
\bottomrule
\end{tabular}
\end{table}

As could have been expected, the GraphSAGE model performs better than GCN. Recall that the key difference between the two models is that at each layer GraphSAGE learns how to aggregate the features due to a node and its neighborhood, whereas the GCN layer aggregates all these together in a fixed manner. Both models learn how to combine the different feature dimensions, but only the GraphSAGE model learns how to combine the data from a node with its neighborhood. However, the surprising result is that both GNN models are outperformed by the standard MLP model that does not take the network structure into account. Moreover, the GraphSAGE model includes the MLP as a special case. To explain this result, recall that the NDSSL data was artificially generated by a large-scale agent-based model simulation. Although we do not have access to the precise details of this simulation, a natural interpretation of this result is that the employment status variable was generated independently of the node features for neighboring nodes. In other words, in terms of predicting this variable for a given node there is nothing to be gained from knowing the features (such as the age, gender, household income, etc.) of that node's neighbors. While this may explain the fact that GraphSAGE does not perform better than the MLP model, it does not address the fact that we may have expected GraphSAGE to roughly match the MLP. This can probably be attributed to the bias introduced as a result of the train/test split and the GraphSAINT sampler discussed above. 

This is admittedly a disappointing result for our first example of GNNs applied to a large-scale social contact network. We tried many variants of this problem, for example predicting other variables like age, household income, etc. All attempts led to the same basic conclusion that the MLP outperformed the GNN models. Therefore, we believe that this example provides a useful lesson to the limitations of GNNs - they certainly cannot uncover correlations that do not exist within the data. Moreover, their application to large-scale networks can potentially introduce biases and sampling error that can significantly affect performance. 

\subsection{Link-Level Prediction \label{sec:linkprediction}}
As the name suggests, in link prediction problems the dependent variable that we aim to predict is associated with the relationship between pairs of nodes. In particular, the dependent variable could be the adjacency  matrix (i.e., whether or not a link exists at all) or a component of the edge feature matrix (i.e., a property of the link). To give a public policy example, surveys are often used to obtain incomplete samples of social networks by asking respondents to list some of their acquaintances. Due to small sample sizes and restrictions on the amount of data that can be collected from any one individual, this results in a social network where many links are unobserved. More precisely, by this we mean that the existence or non-existence of a relationship between some pairs of nodes is unobserved. Link prediction can be used to ``fill-out'' or impute the missing entries in these network datasets. As another example, imagine that the network is fully mapped out, but the edges are imbued with a categorical variable denoting the nature of the interaction, and the goal is to predict this variable.
\begin{figure}[!ht]
	\centering
	\includegraphics[width=1.0\textwidth]{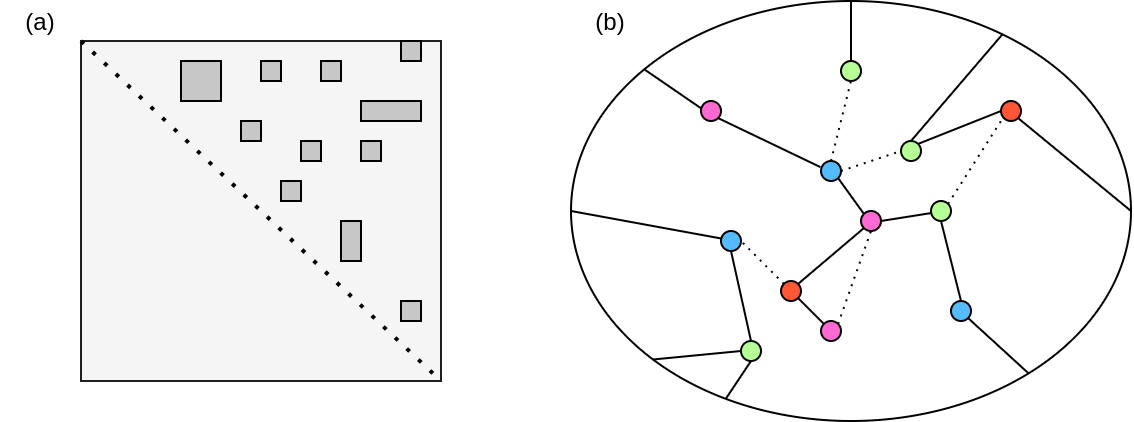}
	\caption{The link prediction problem. (a) The adjacency matrix is only partially observed. In the figure, the dark gray regions correspond to unobserved node pairs that may or may not be connected by an edge. For undirected graphs, the adjacency matrix is symmetric, and so the lower diagonal is completely determined by the upper diagonal, and the unobserved entries in the lower diagonal are not shown. (b) A subset of the graph showing observed edges (straight lines) and unobserved edges (dotted lines). In addition to the possible edges between nodes shown in this subset, there are also `dangling bonds' which are edges to the nodes in the rest of the graph.}
	\label{fig:link_prediction_diagram}
\end{figure}

We focus on the first type of problem where the goal is to predict whether or not a link exists; see Figure \ref{fig:link_prediction_diagram} for a schematic depiction. This may be thought of as a binary classification problem where the dependent variable corresponds to certain entries of the adjacency matrix. Many link prediction techniques exist, and here we focus on a deep learning-based method known as the Graph Auto-Encoder (GAE) \cite{kipf2016variational}. First, a GNN model $f$ (such as a GCN or GraphSAGE) is used to define a $d$-dimensional latent representation, $\bm{Z} \in \mathbb{R}^{N \times d}$:
\begin{equation}
    \bm{Z} = f(\bm{A}, \bm{V}; \bm{\theta}) \,.
\end{equation}
Here $\bm{\theta}$ denotes all the weights and biases of the model. The probability of observing an edge between two nodes $i, j$ is then parametrized in terms of this latent variable according to
\begin{equation}
    \label{eq:link_pred_prob}
    P(A_{ij} = 1|\bm{z}_i, \bm{z}_j) = \sigma( \bm{z}_i^T \bm{z}_j) \,,
\end{equation}
where $\sigma$ is the logistic sigmoid function $\sigma(x) = (1+e^{-x})^{-1}$, and as before $\bm{z}_i$ denotes the $i$-th row of the matrix $\bm{Z}$. The significance of the dot product $\bm{z}_i^T \bm{z}_j = |\bm{z}_i| |\bm{z}_j| \cos(\phi_{ij})$ is that the nodes $i, j$ will likely be connected by an edge if the two latent vectors are near parallel and large in magnitude, and where $\phi_{ij}$ denotes the angle between the two vectors.\footnote{Actually, in our implementations we were able to achieve better performance by modifying Eq.~\ref{eq:link_pred_prob} to be ${P(A_{ij} = 1|\bm{z}_i, \bm{z}_j) = \sigma( \alpha \, \bm{z}_i^T \bm{z}_j + \beta)}$, with $\alpha, \beta$ trainable scalar parameters.} Each potential edge $(i,j)$ is treated as a binary variable, which naturally corresponds to the entries of the adjacency matrix $A_{ij}$. Only some of matrix elements of $\bm{A}$ are observed, and the goal is to make use of these as well as the node attributes $\bm{V}$ in order to predict the values of the missing entries; see Figure \ref{fig:link_prediction_diagram}. This therefore corresponds to a binary classification problem, and once again the cross entropy loss function is appropriate:
\begin{align}
    \label{eq:link_pred_loss}
    \mathcal{L}(\bm{\theta}) &= - \frac{1}{|E_{\text{train}}|} \sum_{(i,j) \in E_{\text{train}}} \left( A_{ij} \ln \sigma(\bm{z}_i^T \bm{z}_j) + (1 - A_{ij}) \ln\left(1 -  \sigma(\bm{z}_i^T \bm{z}_j) \right) \right) \,,
\end{align}
where $E_{\text{train}}$ is the training subset of the observed edge list. Throughout the course of training, minimization of this loss function will encourage the learned representation vectors to align/anti-align for nodes that are connected/not connected by an edge, respectively.

As an example application of this method for link prediction, we consider the task of predicting edges in the NDSSL network data. Recall that the NDSSL data was synthetically generated by an agent-based model, and as a result all the ``true" edges in the network are represented in the data; there are no missing entries. However, there are many reasons edges might be unobserved in real-world data, and this example will demonstrate how graph neural networks can be used to address the \textbf{Data Imputation} aim of Section \ref{sec:aims}, this time with respect to edge-valued data. This example will also be useful in Section \ref{sec:generative} where we extend it to build a generative model over social contact networks. The code used for this example is available \href{https://code.rand.org/hartnett/dgmnet/-/blob/master/Link\%20Prediction.ipynb}{here}.

In this example we parametrized the node representation function $f$ using a 3-layer GraphSAGE network, with 256 hidden units in each layer, with an output dimension of $d = 64$. We performed an equal 50/50 train/test split on the \textit{edges}, rather than on the nodes. All nodes are available at both training and testing time, but the edge set is variable. We chose a 50/50 train/test split to avoid the distributional shift issue discussed above in Section \ref{sec:nodeprediction}. While more training data would likely result in a better model, performing an equal split helps to simplify the evaluation of the model. Also, 50 percent of the roughly 20 million edges in the NDSSL graph is still a very large number of edges from which to train. As before, we used the GraphSAINT sampling framework to allow this model to be used with such a large dataset. During each training pass, a mini-batch of nodes is chosen by the GraphSAINT random walker. The training set edges between these nodes is then used as positive training examples. For each such edge, a non-existent ``negative edge'' is sampled, so that the link prediction model is trained in a balanced way on equal proportions of positive and negative edges.

\begin{figure}[!ht]
	\centering
	\includegraphics[width=1\textwidth]{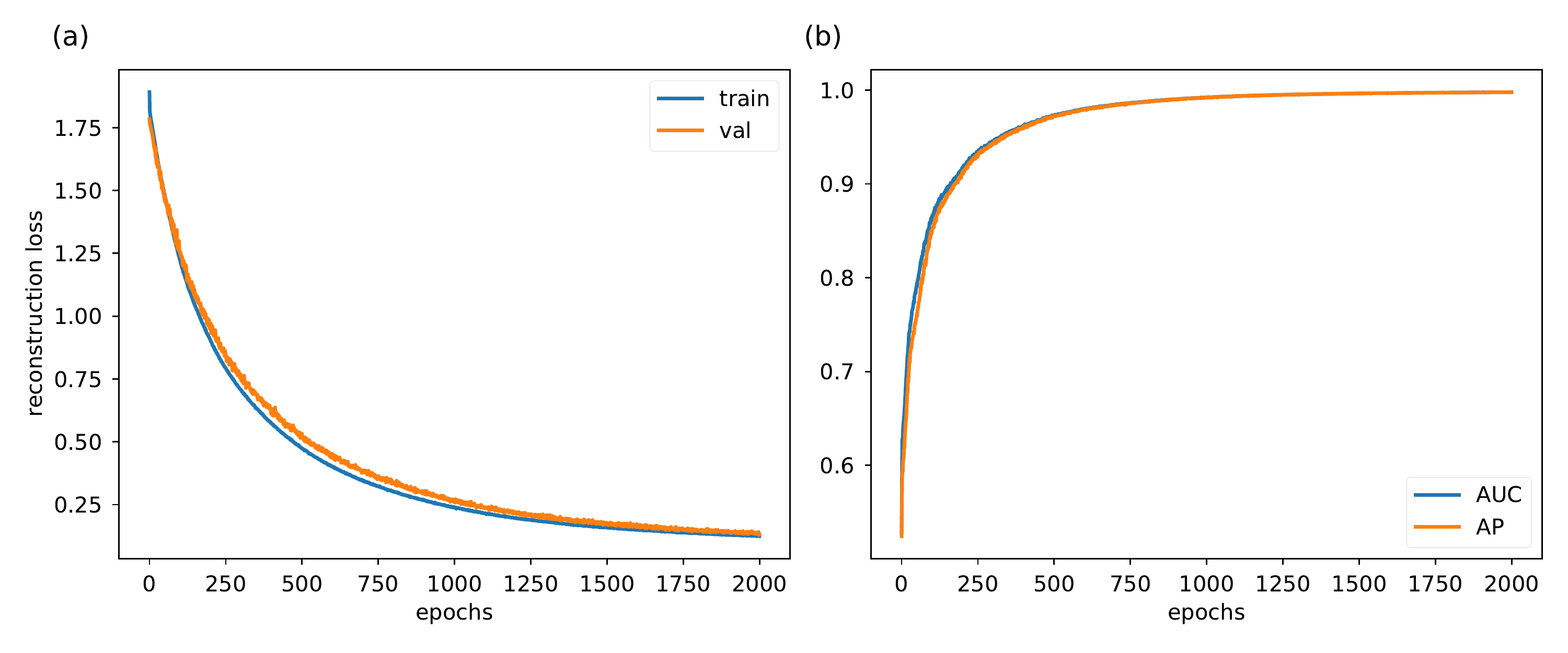}
	\caption{Link prediction performance on the training set. (a) The reconstruction loss. (b) The Area Under the Curve (AUC) and the Average Precision (AP).}
	\label{fig:link_prediction_trainresults}
\end{figure}

\begin{figure}[!ht]
	\centering
	\includegraphics[width=1\textwidth]{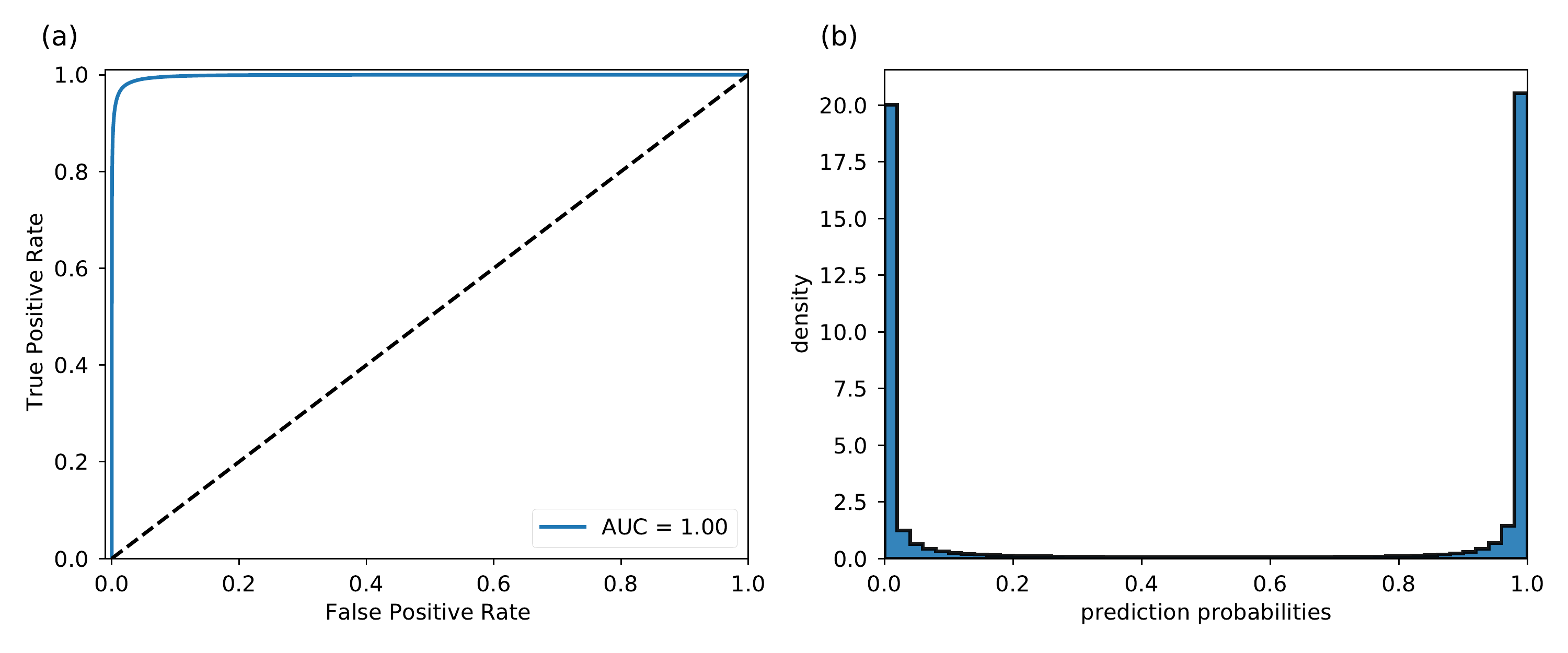}
	\caption{Link prediction results on the held-out test set. (a) The Receiver Operating Characteristic (ROC) curve. (b) A histogram of the probabilities $P(A_{ij} = 1|\bm{z}_i, \bm{z}_j)$ on the test set.}
	\label{fig:link_prediction_testresults}
\end{figure}

The cross entropy loss function Eq.~\ref{eq:link_pred_loss} was minimized over 500 epochs using the Adam optimizer with a learning rate of 1e-3. The results on the training set are shown in Figure \ref{fig:link_prediction_trainresults}. The model achieves near perfect performance. Figure \ref{fig:link_prediction_testresults} shows the results of evaluating the trained model on the test set. The performance on the held-out test set is comparable to the performance on the training set, and the histogram of probabilities shown in Figure \ref{fig:link_prediction_testresults} (b) indicates that the model is quite confident in its determination of an edge as either present or absent. This level of performance is likely due in part to the model learning how to successfully identify the negatively sampled edges, and it is difficult to predict how it would perform under a different scheme of sampling negative edges. This is an important question that we leave for future work. Also, it seems quite likely that the performance will depend on how the missing edges are distributed - for example are the missing edges randomly distributed throughout the graph, or are they associated with a tightly bound cluster of nodes? These issues require further exploration to fully understand the ability of this link prediction algorithm to perform edge imputation.

\subsection{Graph-Level Prediction}
We have considered prediction problems at the node and link levels. The obvious third possibility is to consider prediction at the level of entire graphs. For example, there has been a great amount of work on using GNNs to predict chemical properties of molecules, which may be represented as attributed graphs. In particular, Ref.~\cite{gilmer2017neural} introduced Message Passing Neural Networks (MPNNs) as a way to unify a wide range of graph neural networks, and used these to predict, for example, the atomization energy, vibrational energy, and the dipole moment of various chemical compounds.  

Graph-level prediction is perhaps less naturally useful in public policy contexts, where there may be just one or perhaps a small handful of very large social contact graphs, although one could imagine framing many classic problems in this way. For example, in an epidemiological setting, one could view the network percolation threshold for a disease spreading amongst a population as a function of the attributed social network~\cite{newman1999scaling, newman2002spread, meyers2007contact, piraveenan2013percolation}. Given a dataset of graphs with observed percolation thresholds, graph neural networks could be used to predict the percolation threshold of a new, unseen graph. This sets up a framework somewhat analogous to the use of MPNNs to predict chemical properties of atoms. Just as MPNNs have allowed chemical properties to be directly predicted from data, rather than estimated by complicated models developed by chemists, one could learn how the percolation threshold depends on the attributed graph directly from the data as opposed to relying on the results of epidemiological models and simulations. This just represents one possible example of graph-level prediction in a policy context. We did not pursue this idea in our work; more work is needed to see if approaches like this will be useful for policy research.

\section{Generative Modeling Over Networks \label{sec:generative}}

\subsection{Introduction to Generative Modeling}
The central object of interest in generative modeling over networks is the probability distribution over some ensemble of attributed graphs, $P(\mathcal{G})$. For example, the social contact network of a population may be regarded as a random (attributed) graph, and a network dataset such as NDSSL corresponds to a single realization from the distribution over an ensemble of possible graphs. Generative modeling seeks to model $P(\mathcal{G})$ by fitting some parametrized family of distributions to the data, which we denote as $P(\mathcal{G}; \bm{\theta})$ and refer to as the model. Often (but not always), $P(\mathcal{G}; \bm{\theta})$ is designed to allow for efficient sampling of new graphs so that new graphs may be easily generated once the model has been trained. Generative methods are particularly suited for network modeling problems relevant for public policy. For example, among the challenges listed in Table \ref{tab:RANDresearch}, synthetic network generation, rescaling networks, data imputation, and network fusion may be addressed via generative modeling approaches.

To make this problem more concrete, and to illustrate why it is so challenging, suppose we wish to model the distribution over unattributed, undirected graphs for a fixed population size of $N$ individuals. There are $N(N-1)/2$ possible edges, each of which may be present or absent. The ensemble then consists of $2^{N(N-1)/2}$ distinct graphs.\footnote{This counting treats each distinct graph over $N$ nodes. If the ordering of the individual nodes is irrelevant, then the probability mass assigned to isomorphic graphs should be the same. In terms of the distribution over adjacency matrices, this corresponds to requiring that $P(A) = P(\bm{P}_{\pi} A \bm{P}_{\pi}^T)$. As there are $N!$ permutations over $N$ variables, the ensemble size is then cut down to $2^{N(N-1)/2}/N!$. Using Stirling's approximation, $\ln N! \approx N \ln N + \mathcal{O}(N)$, we can see that the asymptotic growth of the ensemble is unaffected by this distinction.} Similarly, for directed graphs there are $N(N-1)$ possible edges, and the ensemble size is $2^{N(N-1)}$. Including node and edge attributes enlarges this space even further to include all possible matrices $\bm{V}, \bm{E}$ drawn from the appropriate domain (i.e., if there are $F$ continuous node features, then $\bm{V} \in \mathbb{R}^{N \times F}$). To make the problem even more challenging, in stark contrast to the astronomical size of typical ensembles, in many realistic applications one will only have a single observation, for example in the case of the NDSSL network data. Consequently, any attempt to model the $P(\mathcal{G})$ will need to make several assumptions about the functional form of the distribution to make the problem tractable.

Neural network-based generative modeling approaches for networks have been categorized into two main groups: one-shot and recurrent \cite{belli2019image}. One-shot approaches generate the entire graph at once, whereas recurrent approaches grow the graph incrementally. To give examples from other domains, GANs \cite{goodfellow2014generative} would be considered to be one-shot methods and an example of a recurrent approach would be a recurrent neural network (RNN). It is important to emphasize that both approaches involve successive applications of processing layers - the difference is that one-shot approaches produce all dimensions of the output simultaneously, whereas recurrent approaches produce the output dimensions sequentially, one after another.

Many one-shot approaches for graph generation are based on adapting existing generative methods for images to network data. For example, Ref.~\cite{fan2019deep} introduced a technique to train a GAN to generate adjacency matrices for graphs with nodes labels, and \cite{de2018molgan} introduced an adaptation of GANs for generating graphs with labeled nodes and edges, which they then used for the task of modeling distributions over molecular graphs. Normalizing flows \cite{rezende2015variational} are another class of generative models that, like GANs, are based on the principle of transforming a simple latent distribution into the target distribution. Unlike GANs, however, normalizing flows are invertible and their likelihood may be explicitly computed. \cite{liu2019graph} introduced graph normalizing flows to generate artificial graphs of arbitrary size. Another class of one-shot approaches use Variational Auto-Encoders (VAEs) \cite{kingma2013auto}, which include \cite{kipf2016variational, stoehr2019disentangling, simonovsky2018graphvae}.

In contrast, many recurrent approaches draw inspiration from the field of natural language modeling. For example, \cite{you2018graphrnn, li2018learning} both use recurrent neural networks (in particular, LSTM or GRU networks) to sequentially generate the adjacency matrix of a graph. Unfortunately, these methods do not scale well to the large-scale networks often encountered in policy research contexts. RNN-based approaches to language modeling have been recently eclipsed by so-called attention networks \cite{vaswani2017attention}, and these models have also been adapted for deep graph generative modeling \cite{velivckovic2017graph, belli2019image, liao2019efficient}.

Neither approach, one-shot nor recurrent, seem to have a clear advantage over the other. Scalability is a challenge for both frameworks, and many of the methods cited above simply cannot be applied to a large dataset like NDSSL, with millions of nodes and tens of millions of edges. The focus of the current report is on the application of deep neural networks to network science problems, but it is worth noting that there do exist non-deep methods applicable to graphs of this scale, such as Kronecker Graphs \cite{leskovec2010kronecker} and recent variants of Exponential Random Graph Models, which we treat in Section \ref{sec:ergm} below.

\subsection{Graph Generation By Iterated Link Prediction \label{sec:ouralgorithm}}
As briefly reviewed above, there are many existing generative models that could in principle be applied to policy-relevant network data, but in practice many of these methods do not scale well to large datasets. Additionally, as currently formulated, many of these models can be used to solve problems that are closely related to, but not quite the same as, the policy-motivated problems that are of most interest. For example, while there are many existing techniques for generating synthetic networks, we would like to be able to control for certain statistical properties such as the marginal distributions of demographic data (i.e. age, race, gender, income, etc.). Another property we might wish to control for are the demographic mixing ratios, for example the fraction of edges between people of different ethnic groups. It is also unclear whether and how these existing methods may be extended to address the difficult problem of network fusion. For these reasons, we have developed our generative framework that can be used to address these challenges and that scales well to large network sizes. In this section we present our method and illustrate its application to the NDSSL dataset. Our generative model directly addresses both the \textbf{Synthetic Network Generation} and the \textbf{Rescaling Networks} aims of Section \ref{sec:aims}, as the generated networks can have any number of nodes. %

Our generative model consists of three main steps. First, a link prediction model such as the Graph Auto-Encoder (GAE) model reviewed in Section \ref{sec:linkprediction} is trained over the network. Given the node attribute matrix and a partially observed adjacency matrix, this model can be used to predict the remaining links in the graph. In the next step a generative model over the node attribute matrix $\bm{V}$ is learned. This model allows for an artificial node population to be generated, but it provides no guidance as to how these nodes should be connected. In the third step a network over the generated nodes is randomly initialized. The link prediction model is then used to evaluate the probability of each potential edge connecting pairs of nodes. Edges that exist from the random initialization but are unlikely under the link prediction model are removed, and non-existent edges that are likely under the model are added. 

Having outlined the general idea of our algorithm, we now turn to the technical description. As mentioned above, the link prediction model is taken to be the GAE model reviewed in Section \ref{sec:linkprediction}. To develop a generative model over the node attribute matrix $\bm{V}$, we recall that this is a $N \times F$ matrix, with each row corresponding to a node. We regard this matrix as a table, where each row is independent and identically distributed from a joint distribution over the $F$ node feature variables. This is a key observation, as it allows for $\bm{V}$ to be modeled using the Conditional Tabular GAN (CTGAN) model of \cite{xu2019modeling}. CTGAN extends the idea of regular GANs to tabular data, and it is carefully engineered to treat both discrete (categorical) variables as well as continuous variables. Moreover, it can handle continuous variables drawn from multi-modal distributions. We refer the interested reader to the original paper for more details. Once the generative model over the node attributes has been trained, we sample a new table of $N_{\text{gen}}$ node attributes. This serves as the artificial population.

Next, a random adjacency matrix $\bm{A}$ is initialized (to be discussed more below). Then, the edges are iteratively re-wired using the link prediction model. The objective is to adjust the adjacency matrix so that the following equilibrium condition is approximately satisfied:
\begin{equation}
    \label{eq:rewiring_balance}
    P(A_{ij} = 1 | \bm{V}) = \sigma \left( \bm{z}_i^T \bm{z}_j\right) \,,
\end{equation}
where $\bm{z}_i = \bm{Z}_{i,;} = f_i(\bm{A}, \bm{V}; \bm{\theta})$ is the GAE learned representation for node $i$, and $\sigma$ is the logistic sigmoid function. Here we have explicitly indicated the dependence of $\bm{z}_i$ on the node attribute matrix and the adjacency matrix to emphasize that this is a highly non-linear equation. The adjacency matrix is used to determine the learned representations, which are in turn used to determine the probability of observing a given adjacency matrix. Our strategy is to use ${P(A_{ij} = 1 | \bm{V}) = \sigma \left( \bm{z}_i^T \bm{z}_j\right)}$ to re-wire batches of edges at a time, then recompute the learned representations, and then iterate. This procedure defines a Markov chain that should converge to the ``equilibrium" distribution satisfying Eq.~\ref{eq:rewiring_balance} after an initial burn-in period.

\begin{figure}[!ht]
	\centering
	\includegraphics[width=1.0\textwidth]{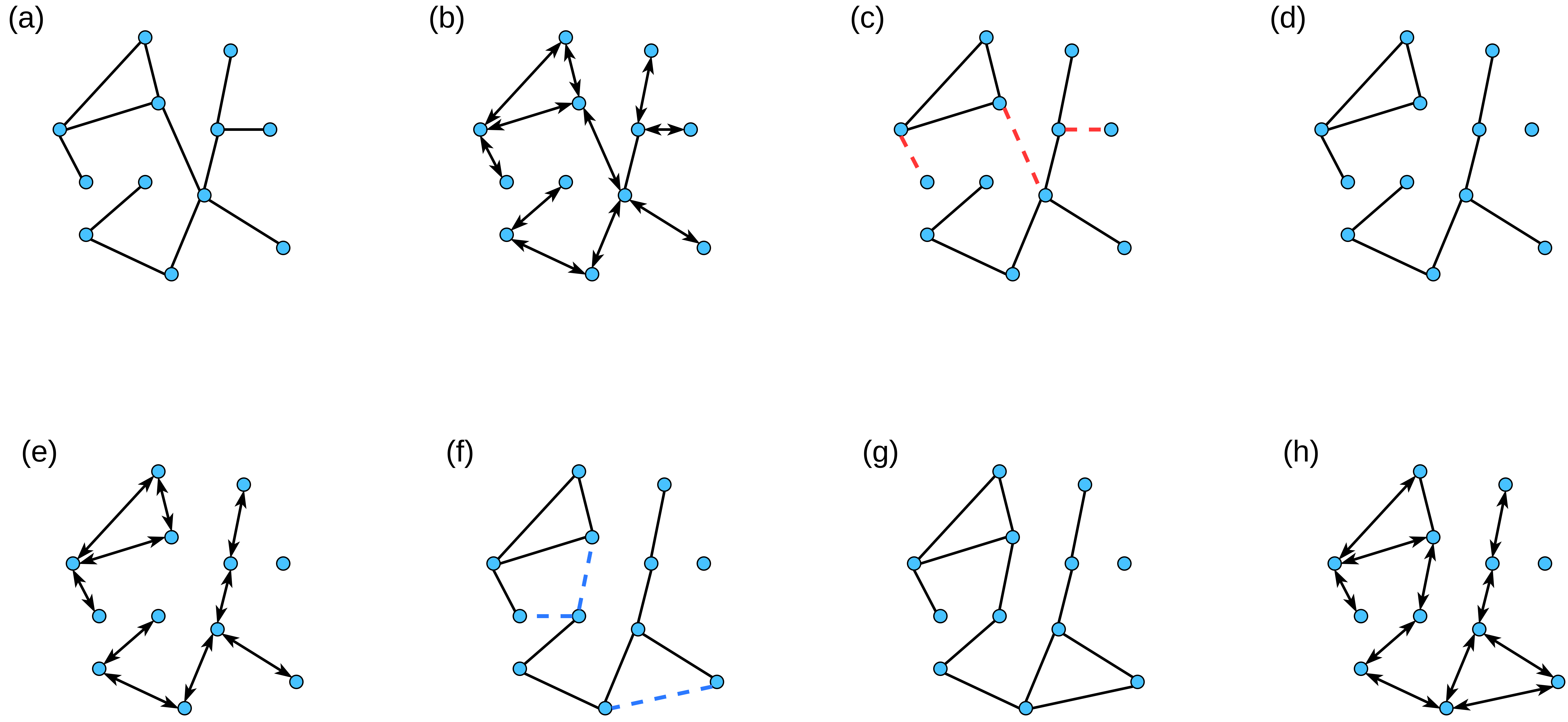}
	\caption{Graph Generation by Iterated Link Prediction. (a) The graph at some time step in the algorithm. (b) The GAE learned representations $\bm{z}_i$ are computed. (c) Edge deletions are proposed. (d) Some deletions are accepted. (e) The learned representations are re-computed. (f) Edge additions are proposed. (g) Some additions are accepted. (h) The learned representations are re-computed, and the algorithm proceeds to either add or delete edges.}
	\label{fig:GILP}
\end{figure}

An important point is that throughout this process we would like to keep the number of edges fixed, in order to stabilize the Markov chain and prevent it from converging to an all-to-all connected graph. Thus, we adaptively alternate between adding and deleting edges so as to keep the number of edges to be roughly constant. This results in a sampling procedure is similar to the improved fixed density (IDF) sampling procedure used for ERGMs \cite{byshkin2016auxiliary}. Our algorithm is summarized in in Algorithm \ref{alg:GILP} and schematically depicted in Figure \ref{fig:GILP}, is iterated many times until the network becomes consistent with the probabilities of the link prediction model.\footnote{This presentation of the algorithm is written in terms of the adjacency matrix but it could just as easily be equivalently formulated in terms of the edge list.} 
\begin{center}
\begin{algorithm}[H]
\caption{Graph Generation by Iterated Link Prediction \label{alg:GILP}}
\SetAlgoNoLine
\SetKwInOut{Input}{input}
\SetKwInOut{Output}{output}
\SetKwInOut{Initialize}{initialize}
\Input{node feature matrix $\bm{V}$ \\
pre-trained GAE link prediction model $f(\bm{A}, \bm{V}; \bm{\theta})$ \\
number of iterations $T$ \\
batch size $n$ \\
target number of edges $M$}
\Output{generated adjacency matrix $\bm{A}$}
\Initialize{Adjacency matrix $\bm{A}$}
\For{$t=1,...,T$}{
    compute representations $\bm{z}_i = \bm{f}_i(\bm{A}, \bm{V}; \bm{\theta})$ \;
    \eIf{$\sum_{i < j} A_{ij} < M$}{
        uniformly sample $n$ positive edges $E^+$\; 
        \For{$(i,j) \in E^+$}{
            set $A_{ij} = 0$ with probability $1 - \sigma(\bm{z}_i^T \bm{z}_j)$
            }
        }{
        uniformly sample $n$ negative edges $E^-$ \;
        \For{$(i,j) \in E^-$}{
            set $A_{ij} = 1$ with probability $\sigma(\bm{z}_i^T \bm{z}_j)$
            }
        }
    }
\end{algorithm}
\end{center}

\subsection{NDSSL Example}
\begin{figure}[!ht]
	\centering
	\includegraphics[width=1\textwidth]{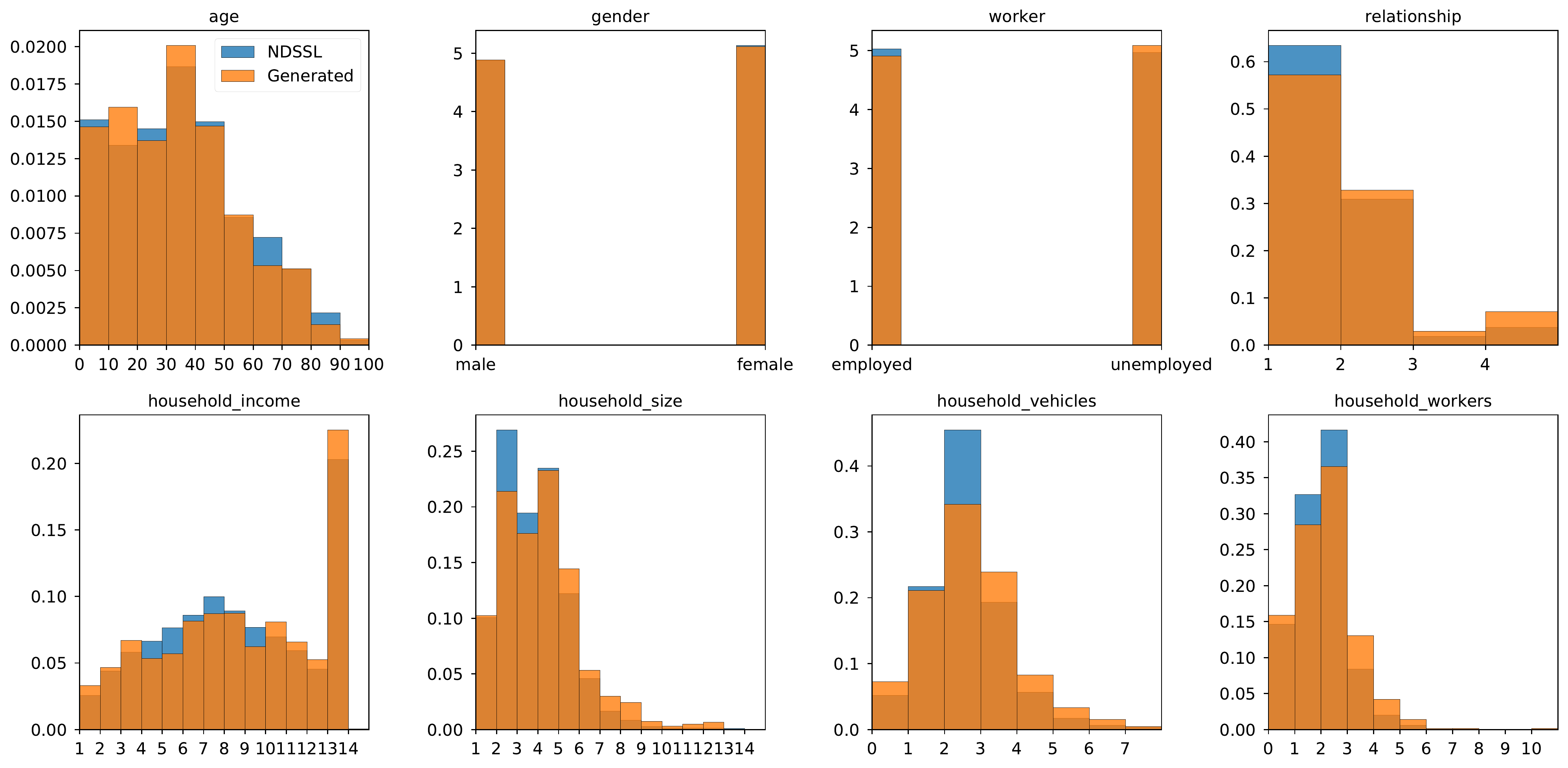}
	\caption{Comparison of the node features artificially generated by CTGAN with the true node features from the NDSSL dataset.}
	\label{fig:ctgan_vs_ndssl_histograms_n_100000_ctgan_epochs_300}
\end{figure}

As an example, we trained our generative framework on the NDSSL data. The Jupyter notebook used for this example is available \href{https://code.rand.org/hartnett/dgmnet/-/blob/master/Graph\%20Generation.ipynb}{here}, and the results can be analyzed using \href{https://code.rand.org/hartnett/dgmnet/-/blob/master/Generated\%20Graph\%20Analysis\%20Notebook.ipynb}{this notebook}. For the link prediction, we used the same GAE model discussed in Section \ref{sec:linkprediction}. We trained a CTGAN model on the NDSSL node attribute matrix and generated new populations with $1000, 10,000$, and $100,000$ nodes. Figure \ref{fig:ctgan_vs_ndssl_histograms_n_100000_ctgan_epochs_300} depicts the histograms of the marginal distribution of many of the node features for the 100,000 nodes artificially generated by CTGAN and compares these against the corresponding plots for the NDSSL data. As the plot shows, the two sets of histograms closely match, although they are not in perfect agreement. This is a strong check that the CTGAN is properly modeling the node features as tabular data. 

In the next step, we initialized an Erdős–Rényi random graph. We decided to set the number of edges $M_{\text{Gen}}$ to satisfy 
\begin{equation}
    \frac{M_{\text{Gen}}}{N_{\text{Gen}}} = \frac{M_{\text{NDSSL}}}{N_{\text{NDSSL}}} = 12.3 \,,
\end{equation}
so that the edge/node ratio in the generated graph matched that of the NDSSL data.\footnote{It is important to note that there is a phase transition in the Erdős–Rényi random graph model. For $M/N < 0.5$, there is no giant component and the graph consists of disconnected trees, and above this threshold the graph does contain a giant component. This ratio is about 12.3 in the NDSSL network, and thus the randomly initialized graph is far from this phase transition and will almost certainly contain a giant component.} Because the number of edges is held roughly constant throughout the algorithm, this initialization represents a significant constraint on the type of graphs that can be explored by the algorithm. In essence, the choice of $M$ helps to define what we mean by our generative model. There is just one NDSSL network, and it is not clear how one might go about defining an NDSSL \textit{distribution} over graphs of different sizes. The Graph Generation by Iterated Link Prediction (Algorithm \ref{alg:GILP}) provides an implicit answer to this question - the distribution should be the one where the probability of edge $(i,j)$ being present agrees with the prediction of the GAE link prediction model trained on the NDSSL data, together with the fact that the node/edge ratio should be the same as in the data. The effect of the initialization and the number of edges is worthy of additional investigation.

\begin{figure}[!ht]
	\centering
	\includegraphics[width=1.0\textwidth]{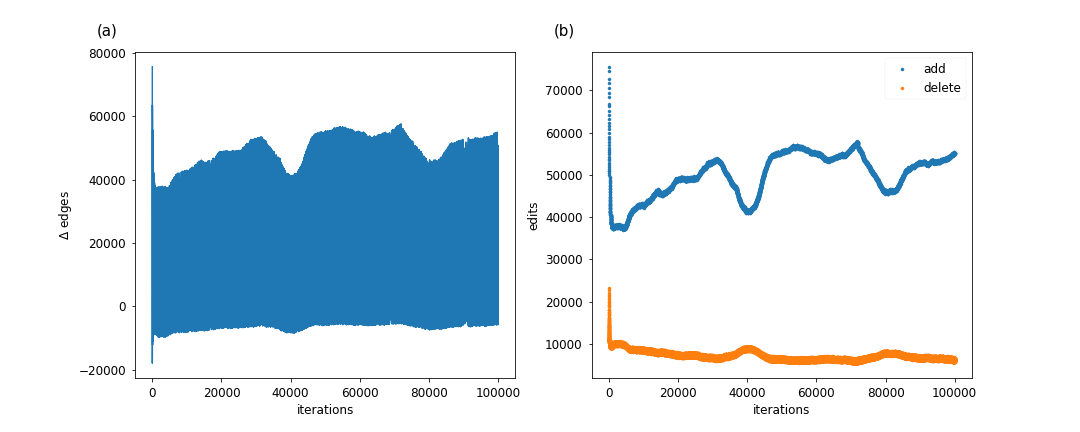}
	\caption{The performance of Algorithm \ref{alg:GILP}. (a) The change in the number of edges at each iteration. (b) The number of additions and deletions during each iteration. Note that in each iteration edges are either added or deleted, but never both.}
	\label{fig:graph_generation_history}
\end{figure}

Figure \ref{fig:graph_generation_history} depicts the behavior of the Markov chain for $N_{\text{Gen}} = 100,000$. The number of edges fluctuates around the target of $M_* = 1,236,242$. Also, the acceptance ratio is different for the addition/deletion steps and varies throughout the algorithm. More additions are accepted in each addition step than deletions in each delete step, and as a result the deletion steps occur more frequently. Without the constraint that the number of edges be roughly constant, the fact that the accept ratios are unequal would push the network to be either empty or complete, neither of which is desirable.

\begin{figure}[!ht]
	\centering
	\includegraphics[width=0.75\textwidth]{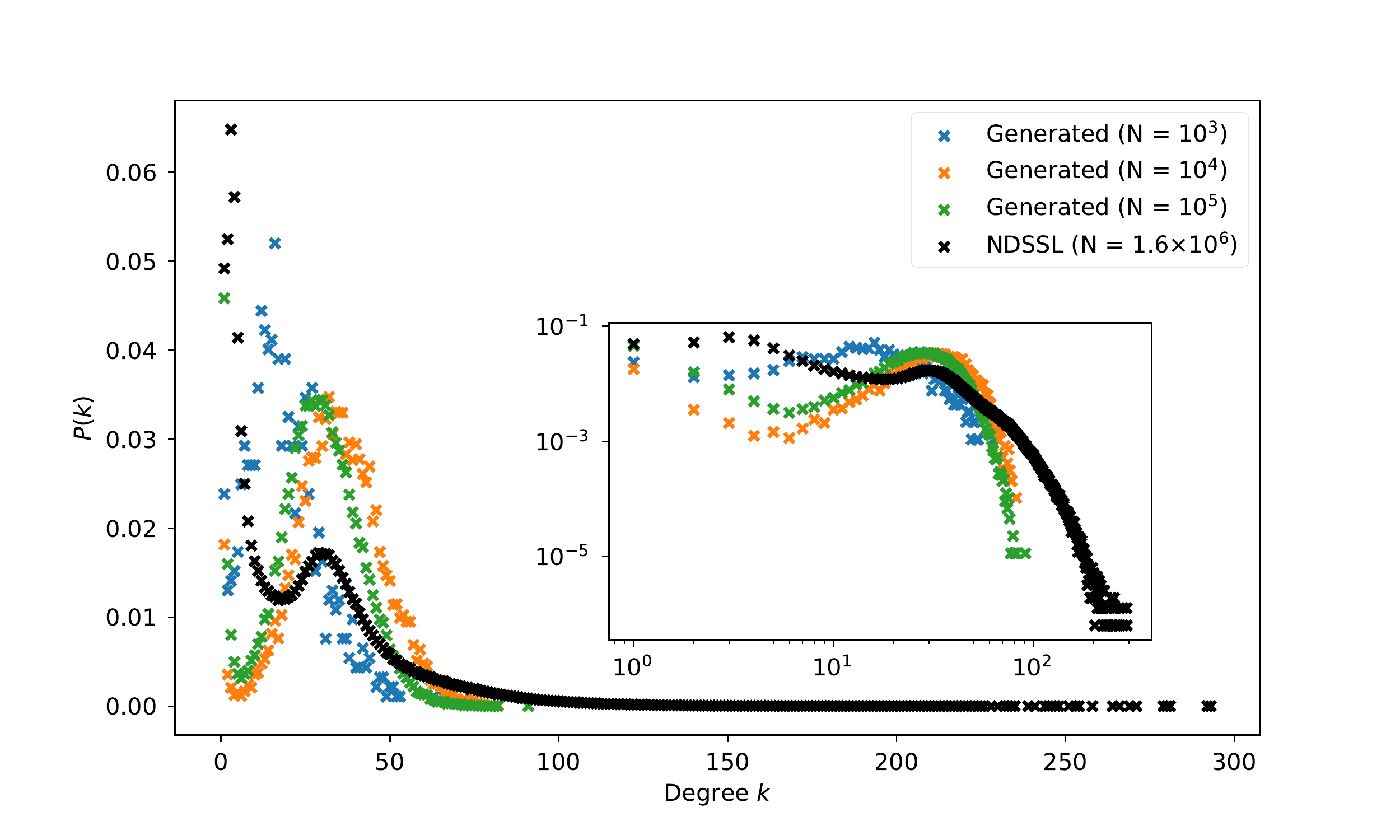}
	\caption{The degree distribution of the three generated graphs and the NDSSL network. The inset shows the data on a log-log scale.}
	\label{fig:graph_generation_degree}
\end{figure}

Next, we analyze the generated networks and compare them to the NDSSL network. First, we print some basic statistics in Table~\ref{table:generated_statistics}. The most significant discrepancy is in the number of triangles. There are about four orders of magnitude fewer triangles in the generated graphs than in the NDSSL data. The number of triangles in a $G(N,M)$ Erdős–Rényi random graph can be computed in the large-$N$ limit to be $N_{\triangle} = \frac{4}{3} (M/N)^3$. Importantly, for constant $M/N$, this does not scale with $N$, and for $M/N \simeq 12.3$, as in the NDSSL graph, this amounts to about 2,500 triangles. Therefore, we see that the algorithm improves upon the initialization and adds more triangles, but it does not add nearly enough to reach the number of triangles observed in the data. In terms of the degree statistics, the $10^5$ network comes closest to matching the NDSSL values. In Figure \ref{fig:graph_generation_degree} we show the entire degree distribution for all four networks. The Generated graphs do a good job of reproducing the peak at degree 30 or so, but they all under-represent the number of low-degree nodes. 

\begin{table}[!htbp]
\caption{\label{table:generated_statistics}Network Statistics} 
\centering
\begin{tabular}{ccccc}
\toprule
    statistic & Gen ($N=10^3$) & Gen ($N=10^4$) & Gen ($N=10^5$) & NDSSL ($N=1.6 \times 10^6)$ \\ \midrule
    largest C.C. & 880 & 9,551 & 86,302 & $1.5 \times 10^6$ \\
    mean degree & 17.3 & 33.2 & 24.8 & 24.3 \\
    median degree & 16 & 34 & 27 & 19 \\
    triangles & 13,442 & 12,924 & 8,589 & $1.3 \times 10^8$ \\
\bottomrule
\end{tabular}
\end{table}

\begin{figure}[!ht]
	\centering
	\includegraphics[width=0.75\textwidth]{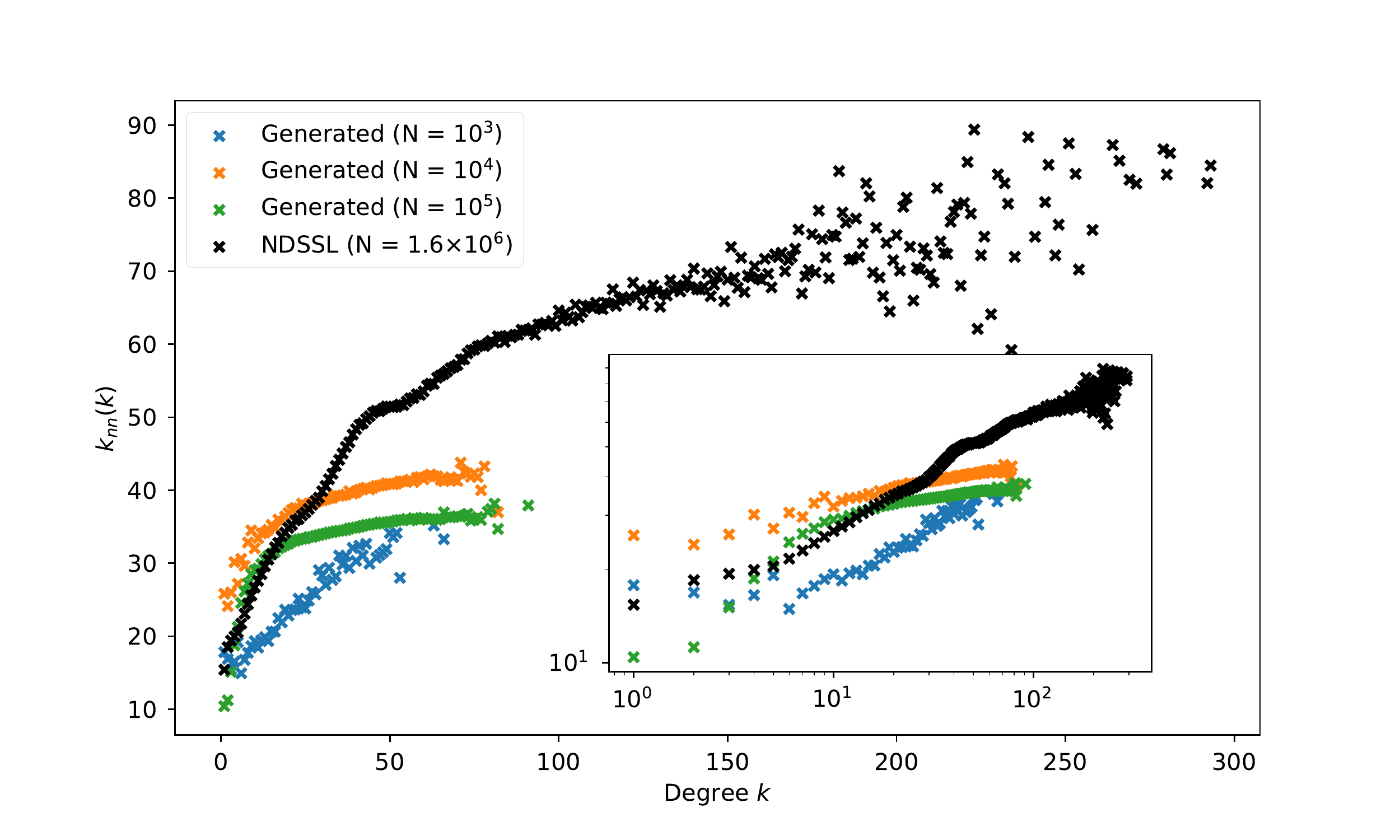}
	\caption{The average degree of nearest neighbors of the three generated graphs and the NDSSL network. The inset shows the data on a log-log scale.}
	\label{fig:graph_generation_assortive_mixing}
\end{figure}

Further insight into the network structure can be gained by examining the average degree of the nearest neighbors, 
\begin{equation}
    k_{nn}(k) = \frac{1}{N_k} \sum_{i/k_i = k} k_{nn,i} \,, \quad \text{where} \quad k_{nn,i} = \frac{1}{k_i} \sum_{j \in \mathcal{N}(i)} k_j \,.
\end{equation}
The general shape of the $k_{nn}(k)$ curve, shown in Figure \ref{fig:graph_generation_assortive_mixing}, indicates whether the network exhibits assortative or disassortative degree mixing. Assortative mixing means that high-degree nodes are likely to have high-degree neighbors, and disassortative mixing means the opposite, i.e. that high-degree nodes are likely to have low-degree neighbors \cite{Barrat2008}. In the present case, both the NDSSL and generated networks are assortative, which indicates that the generative model has successfully captured this property, at least qualitatively.

\begin{figure}[!ht]
	\centering
	\includegraphics[width=1\textwidth]{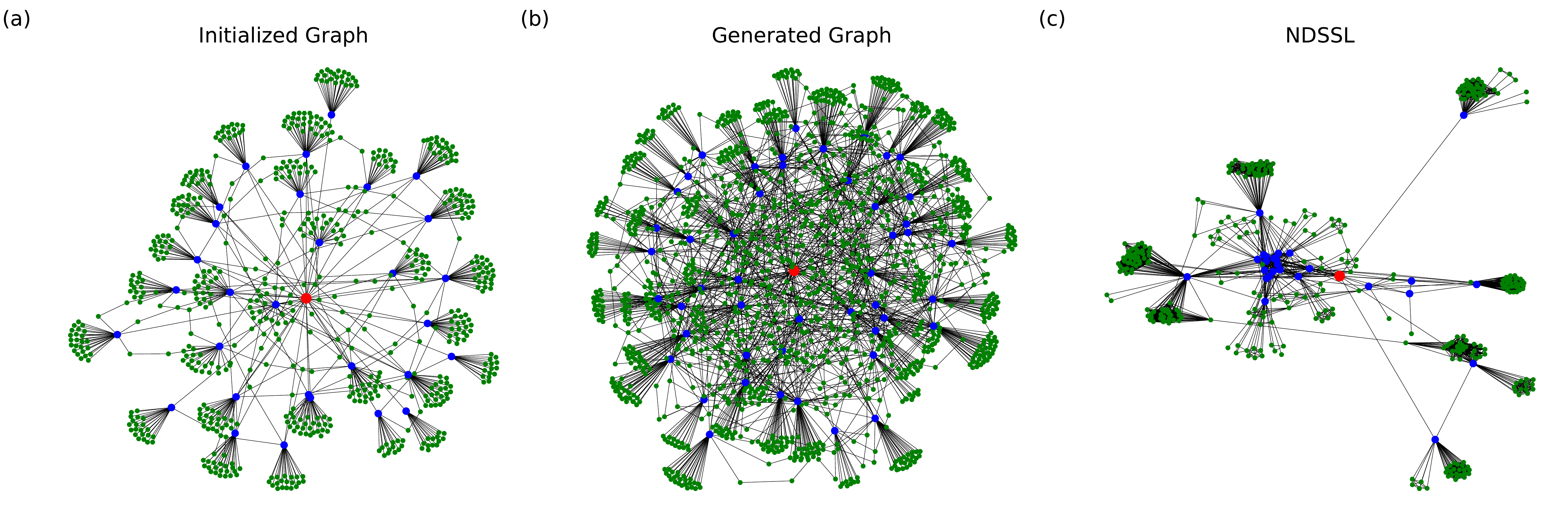}
	\caption{Comparison of ego-centric samples of (a) the randomly initialized ER graph with $N=10^5$, (b) the generated graph with $N=10^5$, and (c) the NDSSL graph.}
	\label{fig:egocentric_comparison}
\end{figure}

There are, of course, many other network statistics that can be evaluated, for example, the clustering spectrum and betweenness or other centrality measures. The massive triangle discrepancy suggests that the generative models will perform poorly with respect to these metrics. Rather than plotting all of these, we simply conclude by showing a single ego-centric sample of the generated graph and the NDSSL graph. Figure \ref{fig:egocentric_comparison} shows a 2-neighborhood ego-centric sample of the initial ER graph, the generated graph, and the NDSSL graph. These plots illustrate that the generated graph does not match well the clustering in the NDSSL data, although it is an improvement over the randomly initialized network.

\begin{figure}[!ht]
	\centering
	\includegraphics[width=1\textwidth]{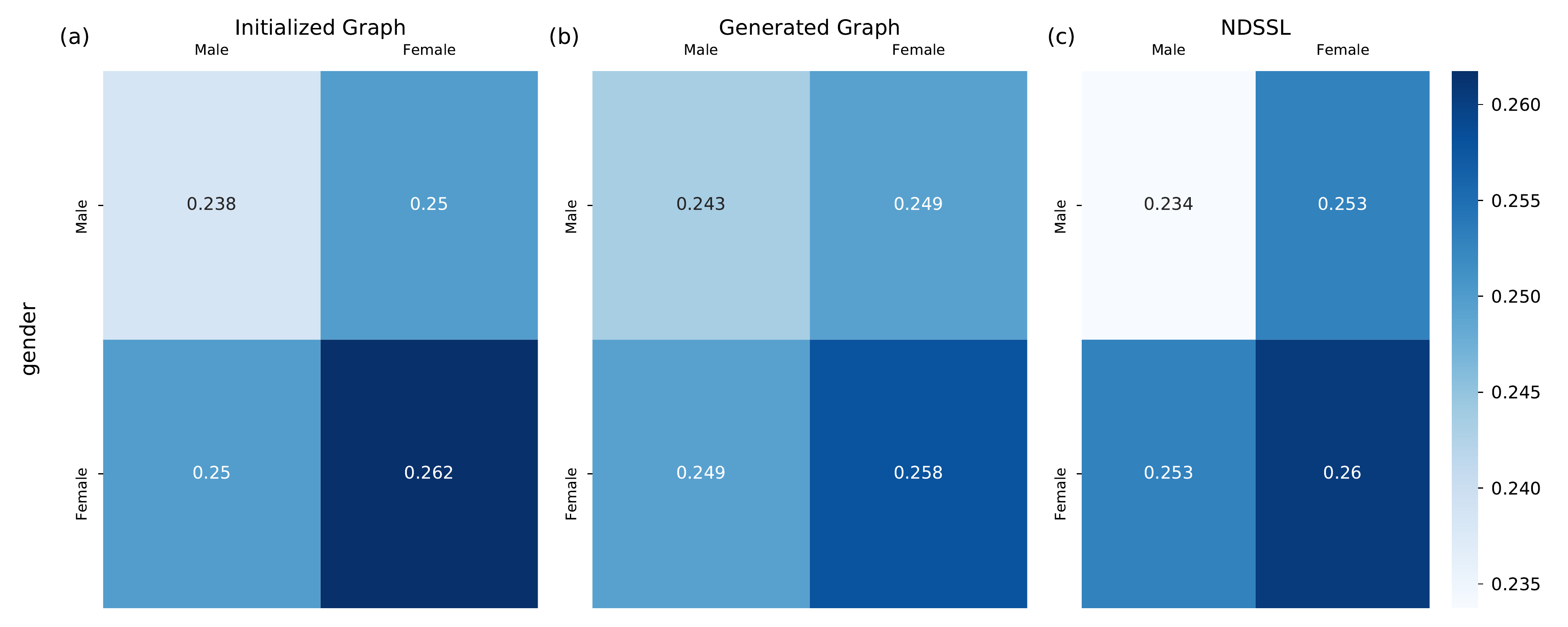}
	\caption{Gender mixing matrix for (a) the $N=10^5$ initialized network, $N=10^5$ generated network, and (c) the NDSSL network.}
	\label{fig:gender_mixing_matrix}
\end{figure}

\begin{figure}[!ht]
	\centering
	\includegraphics[width=1\textwidth]{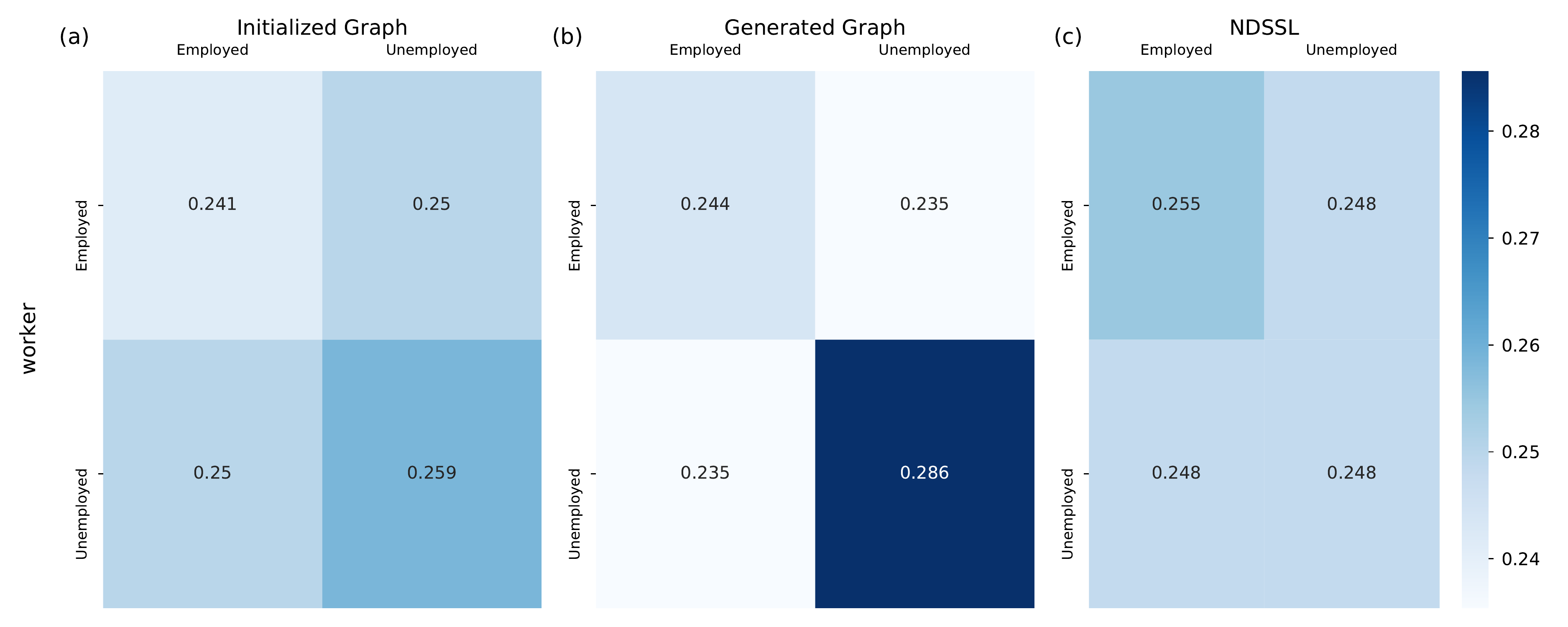}
	\caption{Worker mixing matrix for (a) the $N=10^5$ initialized network, $N=10^5$ generated network, and (c) the NDSSL network.}
	\label{fig:worker_mixing_matrix}
\end{figure}

\begin{figure}[!ht]
	\centering
	\includegraphics[width=1\textwidth]{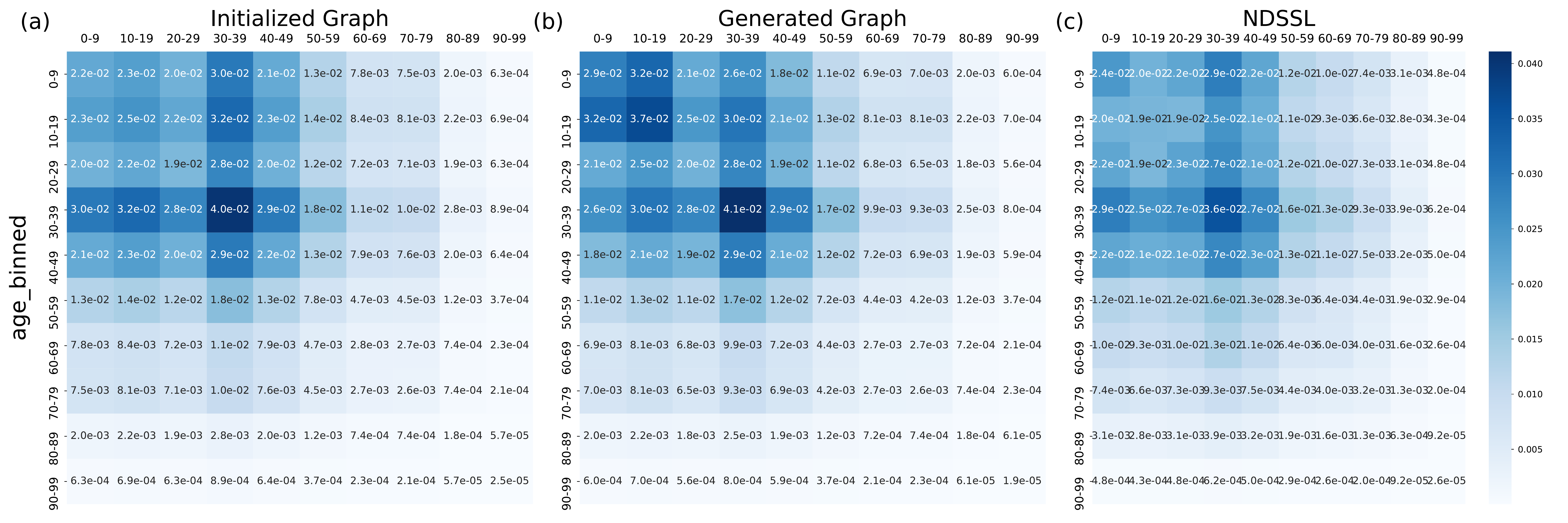}
	\caption{Age mixing matrix for (a) the $N=10^5$ initialized network, $N=10^5$ generated network, and (c) the NDSSL network.}
	\label{fig:age_binned_mixing_matrix}
\end{figure}

Thus far we have investigated the performance of the generated graphs with respect to the node attribute features (i.e. Figure \ref{fig:ctgan_vs_ndssl_histograms_n_100000_ctgan_epochs_300}) and the network structure. Lastly, we investigate the mixing matrices for certain attributes. The mixing matrices are defined as follows for undirected networks. For a categorical variable $y$, let $M_{aa} = |\{(i,j) \in E: y_i = y_j = a\}|$ count the number of edges with $y=a$ at both endpoints, and similarly let $M_{ab} = \frac{1}{2} |\{(i,j) \in E: y_i = a, y_j = b\}|$ count one-half the number of edges with $a$ at one endpoint and $b$ at the other. The entries of the mixing matrix are then defined to be $\mathcal{M}_{aa} = M_{aa}/M$, $\mathcal{M}_{ab} = M_{ab}/M$. Below in Figure \ref{fig:gender_mixing_matrix}, \ref{fig:worker_mixing_matrix}, and \ref{fig:age_binned_mixing_matrix} we show the mixing matrices for the gender, worker, and age attributes for the $N=10^5$ initialized, $N=10^5$ generated, and the NDSSL networks. The agreement between the generated network and the NDSSL network is good, although interestingly the randomly initialized graph also does a good job matching the mixing matrices. Given the quality of the generated node features (shown in Figure \ref{fig:ctgan_vs_ndssl_histograms_n_100000_ctgan_epochs_300}), this suggests that the network structure only weakly affects the mixing ratios.

To summarize, the generative model overall seems to be doing a good job producing graphs with similar statistics to the NDSSL data, although there are many areas for improvement. The most significant discrepancy is in the number of triangles and with the clustering of the network. The under-density of low-degree nodes is also a point of concern. We hope to address these issues with future work.

\section{Exponential Random Graph Models \label{sec:ergm}}
Our main focus in this report is on recent approaches towards these problems that utilize deep neural networks. However, we would be remiss if we failed to mention Exponential Random Graph Models (ERGMs), which are a class of statistical models widely used by social network scientists. ERGMs may be used to create new, artificial graphs of arbitrary size, just like the iterated link prediction algorithm discussed above. Therefore, ERGMs are also a class of models that address the \textbf{Synthetic Network Generation} and \textbf{Rescaling Networks} aims of Section \ref{sec:aims}.

ERGMs are statistical models defined over ensembles of attributed graphs for which the probability distribution may be written as
\begin{equation}
    \label{eq:ERGM}
    P(\mathcal{G};\bm{\theta}) = \frac{\exp\left( \bm{\theta}^T \bm{f}(\mathcal{G}) \right)}{Z(\bm{\theta})} \,.
\end{equation}
Here $\bm{\theta} \in \mathbb{R}^k$ is a vector of parameters, $\bm{f}$ is the feature map, and $Z(\bm{\theta})$ is the normalizing constant or partition function given by
\begin{equation}
    \label{eq:partition}
    Z(\bm{\theta}) = \sum_{\{\mathcal{G}\}} \exp\left( \bm{\theta}^T \bm{f}(\mathcal{G}) \right) \,,
\end{equation} 
where $\{\mathcal{G}\}$ denotes all attributed graphs in the ensemble. The term `Exponential Random Graph Model' arises from the fact that Eq.~\ref{eq:ERGM} describes a probability distribution over random graphs that is in the exponential family, and in this sense, $\bm{f}$ is often referred to as the sufficient statistic. ERGMs are also known as p$^*$ models. There is a long body of work on ERGMs, and they have been studied by mathematicians, network scientists, and social scientists for many decades. For more detailed introductions to these models, see for example \cite{robins2007introduction, hunter2008ergm, van2017introduction}.

Given a dataset of one or multiple observed graphs, the parameters of an ERGM are typically fit by maximum likelihood estimation (MLE),
\begin{equation}
    \label{eq:ERGM_fit_condition}
    \hat{\bm{\theta}} = \text{argmin}_{\bm{\theta}} \, \mathcal{L}(\bm{\theta}) \,,
\end{equation}
where $\mathcal{L}(\bm{\theta}) = \mathbb{E}_{\text{data}} \ln P(\mathcal{G};\bm{\theta})$ is the log-likelihood function. For most problems the log-likelihood is intractable due to the sum over all graphs in the ensemble in the partition function, and thus in practice various approximations and/or Monte Carlo Markov Chain (MCMC) techniques are often employed. As a result of the exponential form of $P(\mathcal{G};\bm{\theta})$, MLE-fitted ERGMs may be given a natural interpretation as the maximum entropy distribution for which the expected value of the feature map agrees with the average value on the dataset, i.e. Eq.~\ref{eq:ERGM_fit_condition} implies
\begin{equation}
    \label{eq:ERGM_statistics_condition}
    \mathbb{E}_{\hat{\bm{\theta}}} \bm{f}(\mathcal{G}) = \mathbb{E}_{\text{data}} \bm{f}(\mathcal{G}) \,.
\end{equation}

As stated above, ERGMs may be used to create new, artificial graphs of arbitrary size. Once the parameters $\bm{\theta}$ have been fit, new graphs may be sampled from the model using Monte Carlo Markov Chain (MCMC) approaches. Moreover, MCMC methods are also used for the fitting procedure itself. This is because the first term in the fitting condition Eq.~\ref{eq:ERGM_fit_condition} involves model expectations. Thus, the ERGM model must be sampled both to train the parameters $\bm{\theta}$ and to generate new graphs after the model has been trained. Consequently, the MCMC approach used is incredibly important for the quality of the ERGM, and much work has been devoted to this issue. Unfortunately, ERGM distributions are multi-modal with the modes separated by large barriers \cite{snijders2002markov}. This can lead to slow mixing times and a potential breakdown of ergodicity for large graphs. Similarly, \cite{bhamidi2008mixing} showed that the MCMC methods and related techniques will only converge to accurate estimates when edges form independently like in an Erdős-Rényi model. So, even sparse networks with correlations and inter-dependencies will be prone to inaccuracies in its estimation. A similar result along these lines was proven in \cite{bannister2014ergms}, who showed that the problem of calculating the partition function $Z(\bm{\theta})$ is \#P-hard, and thus there should be no polynomial-time algorithm capable of accurately sampling an ERGM distribution. 

Thus, although ERGMs are attractive models because of their conceptual simplicity and firm theoretical foundation, they become of limited utility when applied to large-scale networks. In many applications, the goal of maximizing the likelihood is abandoned and instead more computationally efficient approximate methods are used such as the pseudo-likelihood approach \cite{besag1975statistical}. This often leads to good results in practice, but there are many cases where this approach fails in one way or other (see for example the discussion in Section 5.2 of \cite{hunter2008ergm}).

\subsection{NDSSL Example}
While the ERGM framework is generalizable due to its management of complex dependencies in networks, the normalizing factor $Z(\bm{\theta})$ is computationally intractable for large networks. This is due to exponential growth in the number of potential edges as a function of $n$. A major theme of methodological research on ERGMs has been finding workarounds to this computational intractability, such as logistic methods for special cases of the Benouilli graph \cite{holland1981exponential}, capturing dependence using temporal information on network change \cite{almquist2014logistic}, or modeling one or more network samples to later scale using an offset for network size \cite{krivitsky2011adjusting}. There has also been recent work in developing new MCMC algorithms that avoid the slow-mixing time problem \cite{byshkin2016auxiliary, byshkin2018fast}. In this section, we apply ERGMs to the NDSSL data and try three different network sampling techniques as workarounds for modeling networks of this size. The code for this analysis may be found \href{https://code.rand.org/hartnett/dgmnet/-/blob/master/ERGM\%20Analysis/NDSSL_ERGM.Rmd
}{here}. Though a complete treatment of all possible techniques is beyond the scope of this document, we intend to demonstrate techniques in modeling large networks using ERGMs, as well as the considerations that complicate the use of ERGMs in practice. As a target metric, we assess whether ERGMs can reproduce the degree distribution of the NDSSL, since it is a simple and important target for graph prediction.

\begin{figure}[!ht]
	\centering
	\includegraphics[width=0.65\textwidth]{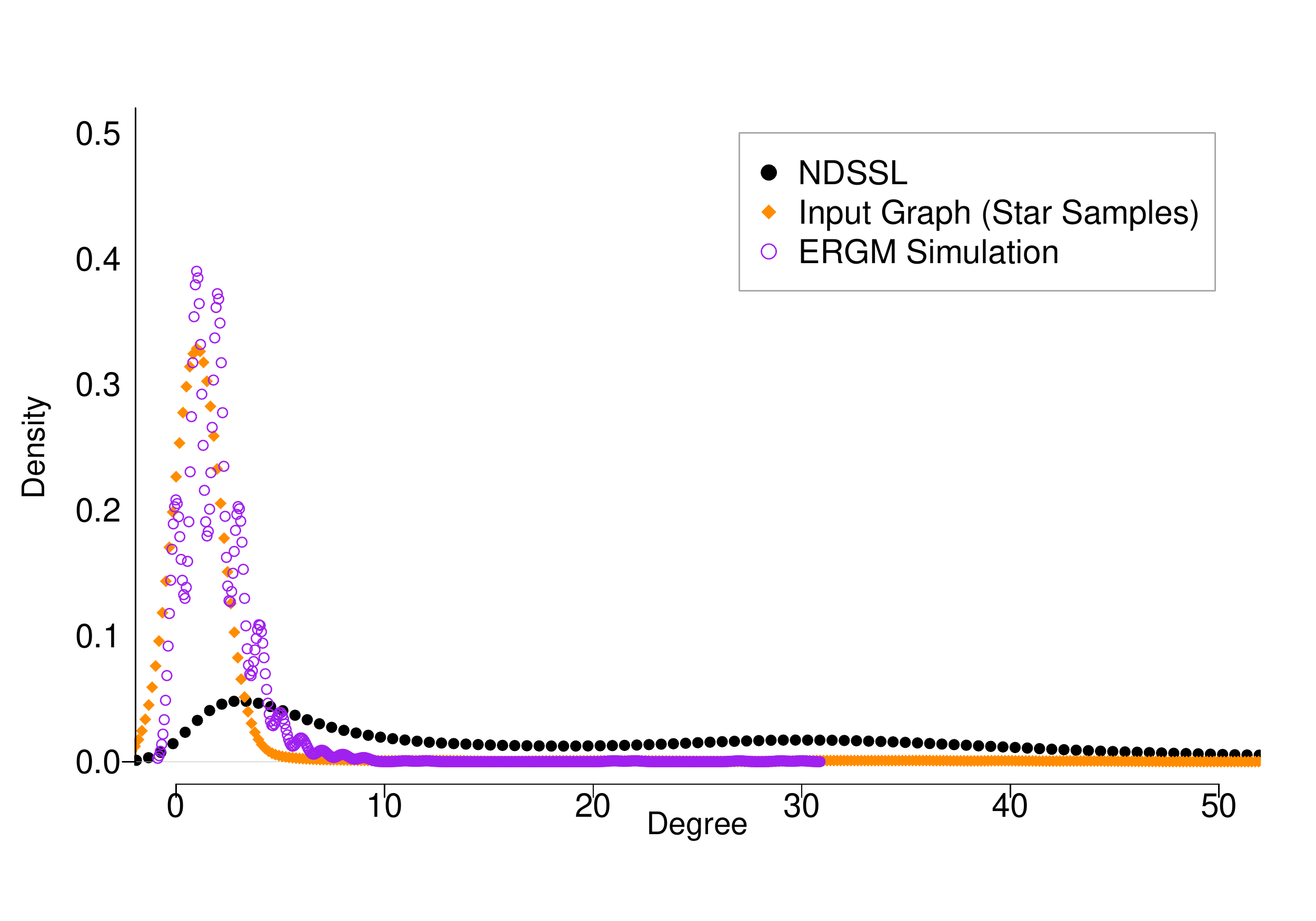}
	\caption{Degree distributions of predicted, input, and target graphs for an ERGM of star samples from the NDSSL.}
	\label{fig:star_deg}
\end{figure}

We first attempt to fit an ERGM to the NDSSL using a random sample of 100 nodes from the network and their local alters with no alter-alter tie information (referred to as a "star sample"). We also know the individual characteristics of each alter, such as household ID and demographic variables. We applied an ERGM using covariates associated with our sampled nodes, such as an employment binary variable, zip code, and household size, along with variables associated with node matching, such as matching on household ID, zipcode, gender, age, income, and employment status. As shown in Figure \ref{fig:star_deg}, simulating from the fitted model, we reproduce the input graph very well, but miss our target on the full graph: The star sampling technique does not carry over enough network structure to reproduce the entire NDSSL network. Indeed, the ERGM can at best reproduce the graph it sees; since much of the network structure is unobserved, the model does not reproduce the population network structure very well.

\begin{figure}[H]
	\centering
	\includegraphics[width=0.65\textwidth]{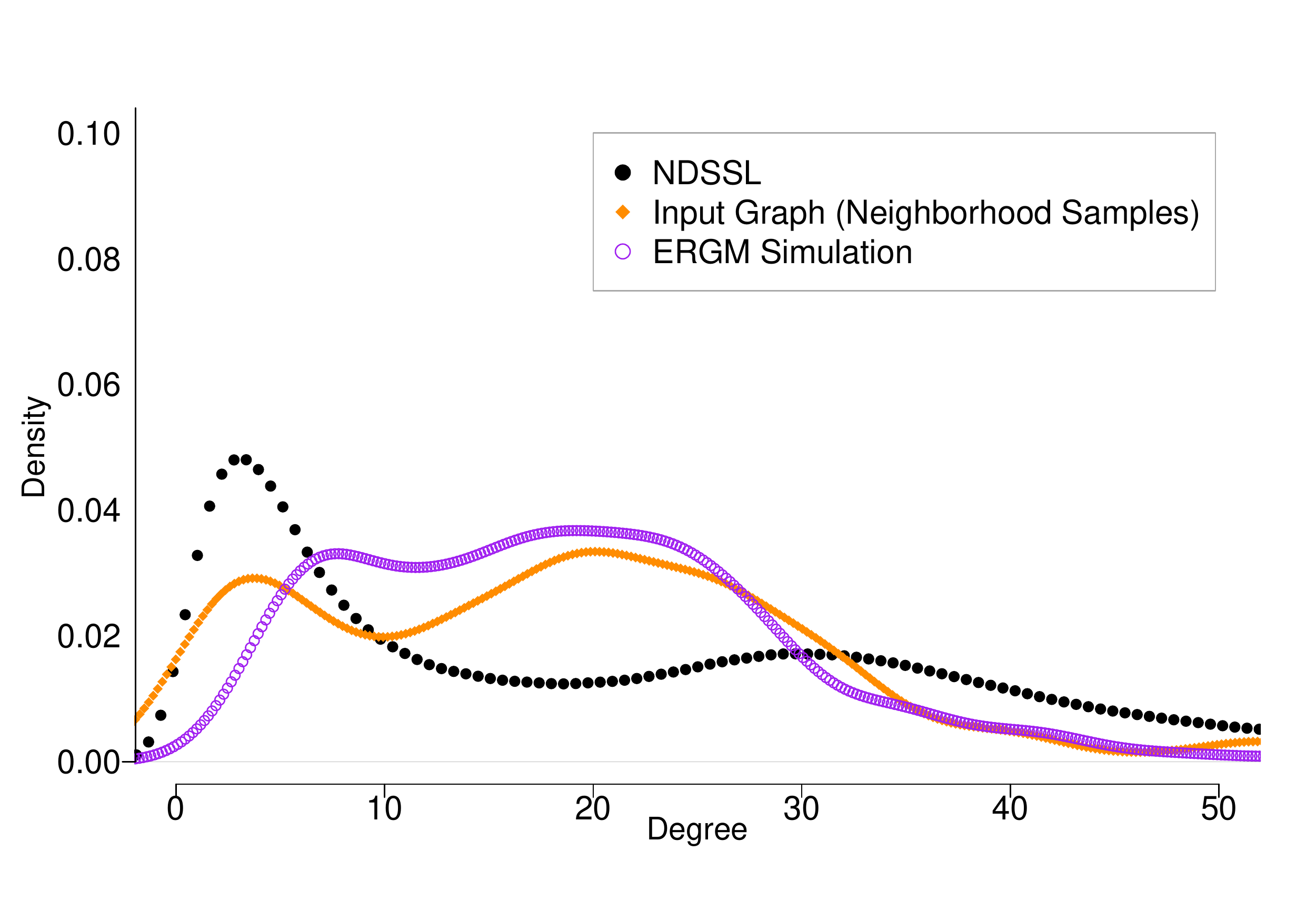}
	\caption{Degree distributions of predicted, input, and target graphs for an ERGM of neighborhood samples from the NDSSL.}
	\label{fig:mod_neigh1}
\end{figure}

We can try to capture more of the network structure by sampling local neighborhoods, i.e., sampling egos but also including the ties between their alters. Since we capture alter-alter ties, we get much richer information on network structure as a whole. Using the same model as before, we get much closer to the target degree distribution, but not quite; see Figure \ref{fig:mod_neigh1}.

\begin{figure}[!ht]
	\centering
	\includegraphics[width=0.65\textwidth]{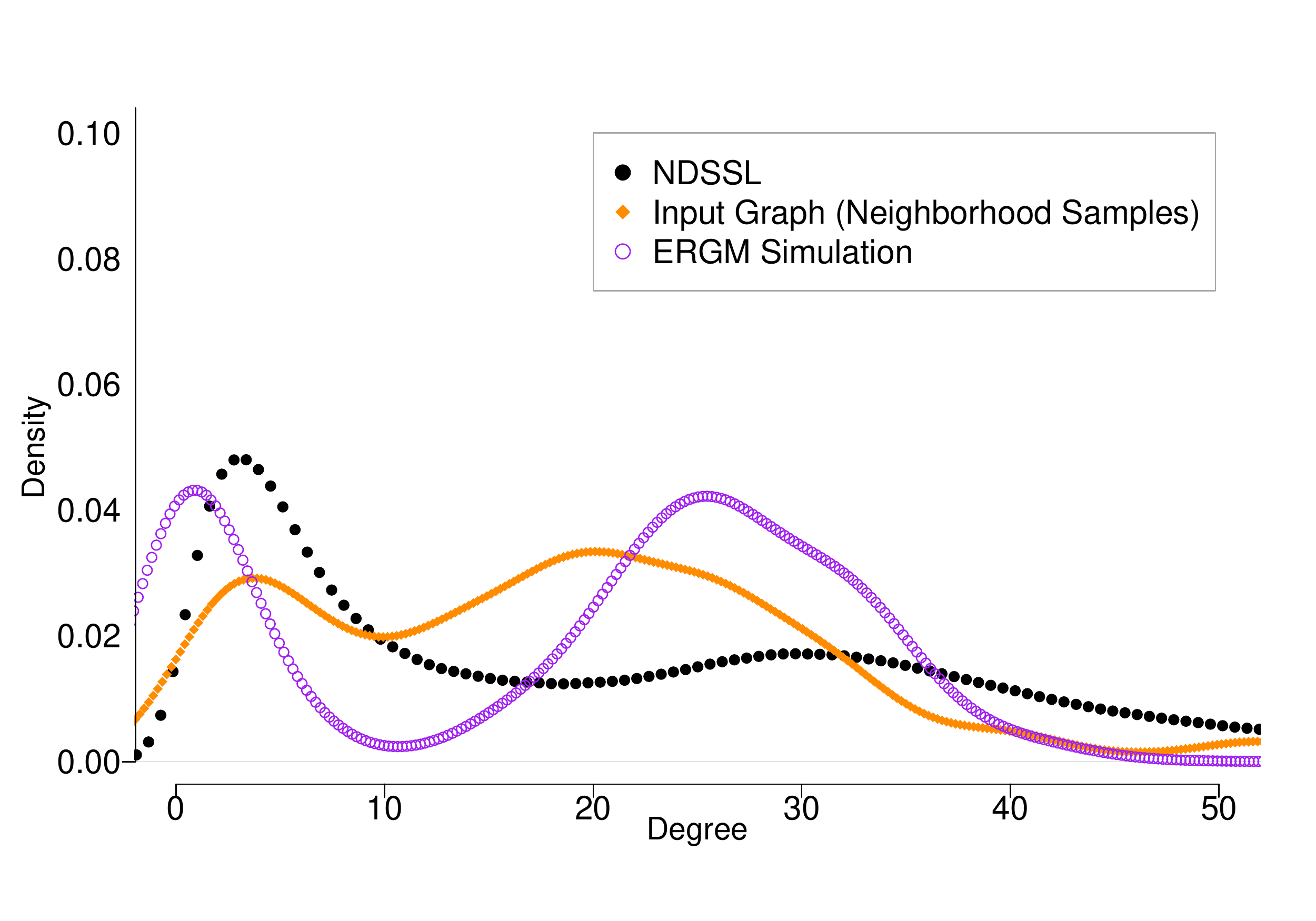}
	\caption{Degree distributions of predicted, input, and target graphs for an ERGM of neighborhood samples from the NDSSL, including shared partner term.}
	\label{fig:mod_neigh2}
\end{figure}

One additional source of information that ERGM could use to reproduce the target graph is in the estimation of shared partner effects. In friendship networks, for example, the number of shared friends between two nodes is related to their probability of being friends -- "a friend of a friend is my friend." Transitivity in social networks is a major reason for the violation of the independence assumption between edge variables. As the number of shared partners increases, there are usually diminishing returns, i.e., going from no shared friends to one shared friend is a bigger gain on the probability of a realized edge than from one shared friend to two shared friends. To account for this, we can use geometrically weighted edgewise shared partners (GWESP), which uses a decay parameter corresponding to the diminishing returns that each additional shared partner adds to the probability that a tie will be present. We try it with a starting decay parameter `2', meaning the added benefit of each additional shared partner decreases the gain in the likelihood of an edge by 1/2. The ERGM will automatically vary this starting parameter to try to find the optimum decay parameter to use to explain the input graph. The result is shown in Figure \ref{fig:mod_neigh2}

Even with ``triangle" effects, we see that we are not quite matching the degree distribution of our target graph. In thinking about the sampling technique used here, we likely oversample triangles in the network due to every alter captured in the sample having at least one shared partner (this stands in interesting contrast to the iterated link prediction model introduced in Section~\ref{sec:generative} that dramatically under sampled the triangles). The only vertices in the graph who do not necessarily have at least one shared partner are the sampled egos. We thus do not observe the true ``denominator'' of the input graph, since we stop our observation of potential edges with the alters of our sampled egos. 

\begin{figure}[!ht]
	\centering
	\includegraphics[width=0.75\textwidth]{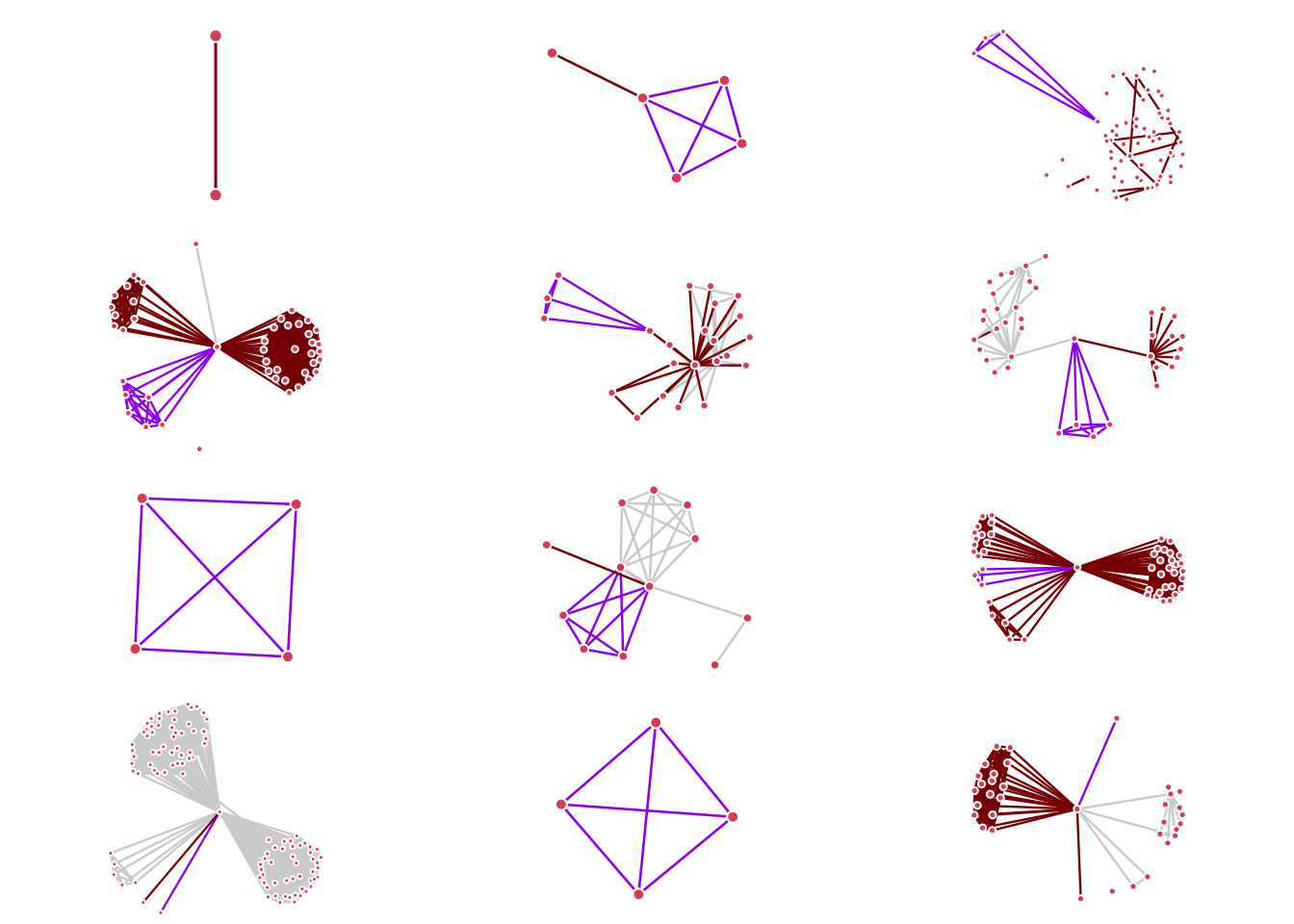}
	\caption{Social networks of twelve randomly sampled egos, with edges colored by purpose (home (red), work (purple), and other (gray). Egos often travel between two to three well-connected subgraphs in a single day. }
	\label{fig:networks_purpose}
\end{figure}

Another feature of the NDSSL - namely, the connection of a single node to many separate, highly-connected networks - is also a complicating factor. To illustrate this, in Figure~\ref{fig:networks_purpose} we show many randomly sampled egos from the NDSSL network. The edges are colored according to edge purpose. Here we can see that many egos (located in the center of each plot) make their way through a series of complete networks throughout the day, i.e., a home, workplace, place of commerce, and/or other setting where every node is connected to every other node. Otherwise, nodes are not that connected. An ERGM model that attempts to find averages for each parameter might have trouble finding a single number for the occurrence of triangles across the entire network. One solution is to include a parameter that represents the tendency for those with only a single shared partner to (not) form a tie. This might help the model to produce networks with heavily connected social settings but keeps between-setting ties at a minimum. We can include in the model a term that separately identifies edge variables with only one shared partner, estimating that parameter apart from the rest of the triangles (estimated by GWESP). However, the model did not converge, and we have issues with degree distribution matching the input graph but not the full network. The result is shown in Figure \ref{fig:mod_neigh3} below.

\begin{figure}[!ht]
	\centering
	\includegraphics[width=0.65\textwidth]{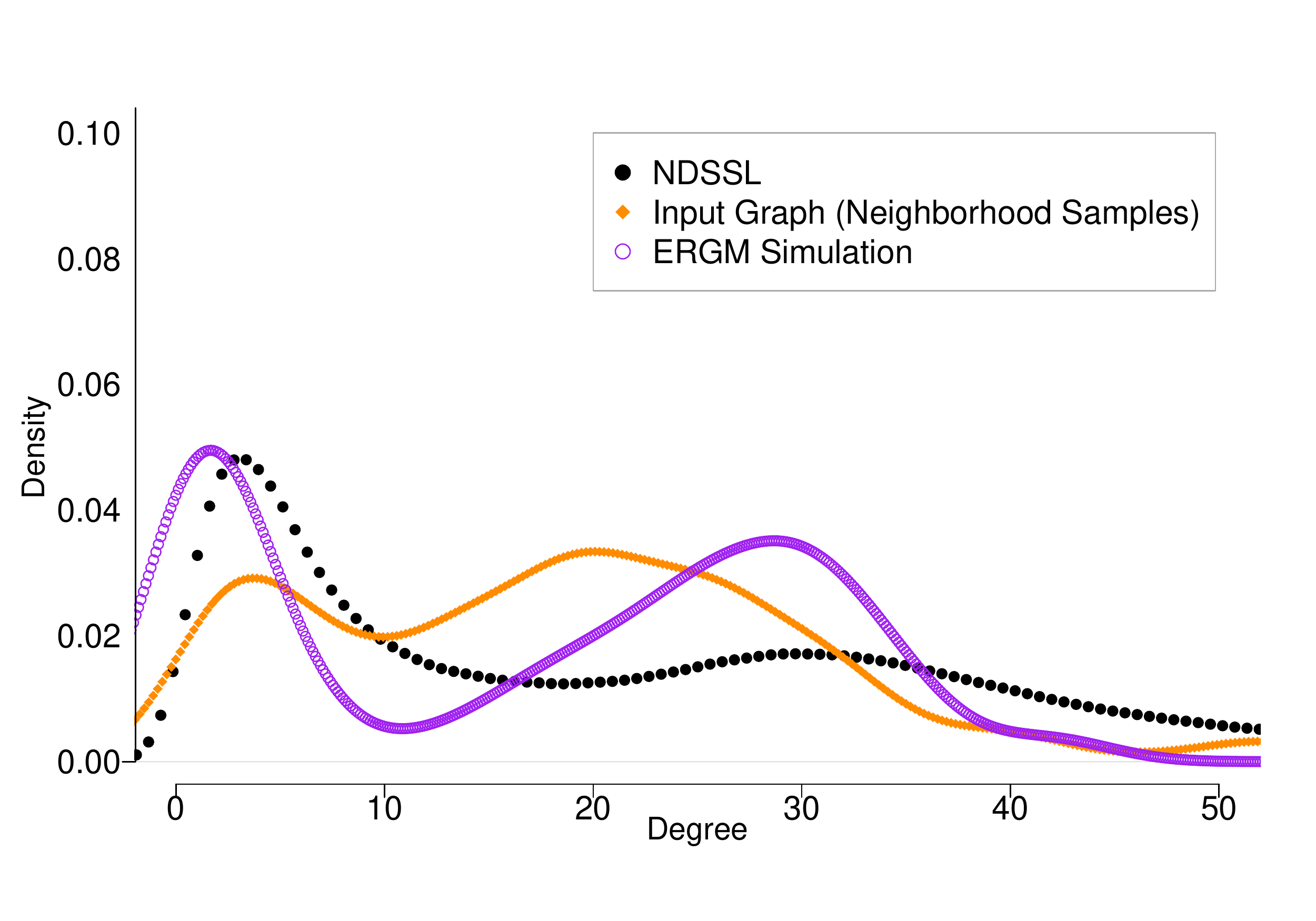}
	\caption{Degree distributions of predicted, input, and target graphs for an ERGM of neighborhood samples from the NDSSL, including shared partner term plus term for a single shared partner.}
	\label{fig:mod_neigh3}
\end{figure}

We are asking the model to produce graphs in a rather unnatural way, namely by ignoring the entire edge structure of our sample's alters. Alters of our egos are often seen to be a part of highly-connected subgraphs. Since here we observe temporally dependent network-ecologies of the seed nodes aggregated by day, we could ask ERGM to consider only the connections between seed nodes. However, with only 100 seed nodes, the ties in-between those 100 nodes cannot represent the structure of the entire network. For example, there is a very, very, small chance that we sample two members of the same household, so the relationship between household membership and tie formation cannot be established in the model. One alternative to this method is to use snowball sampling, which extracts the entire network information of a seed sample, their alters, their alters' alters, and so on. Each new `wave' of the sampling procedure extends the sample to include the neighbors of the nodes in the current sample.

The snowball sampling can be implemented using the ERGM software package in R, which itself uses the statnet suite. This package includes parameters to restrict parameter estimation to only include certain edge variables \cite{handcock2008statnet}. Here, we use a method from \cite{stivala2020exponential}. First, we count as fixed the edge variables in the outer-most wave, since we do not observe any further waves beyond that could account for those edges.  We also hold constant at least one present edge variable in nodes between waves, to account for the fact that they arrived in our sample by nature of being connected to a node in the previous wave. Finally, we hold all ties from Wave 3 to Wave 2 constant, as we do not observe all of Wave 3's edge variables. So, in this case, we only consider those ties that exist within-wave and between Waves 1 and 2 to estimate parameters. The idea is to only consider those edge variables where we have complete information on what could impact the edge being present or absent.  For our sample we used 5 initial seeds, and then expanded the sample in 3 waves, resulting in a final sample size of about 4000 nodes.

The number of nodes in each new wave of snowball sampling scales very quickly, which is natural in networks with high degree.\footnote{See Figure \ref{fig:snowball} in Appendix \ref{app:ndssl} for a plot of this growth rate in the NDSSL network.}  Note that this is not often the case in other types of social networks, such as in friendship networks and sexual contacts. In almost all social networks, density declines with population size, as individuals cannot scale the number of ties linearly with the number of potential ties as the population grows (Friendships and romantic partnerships require much more cognitive and time commitments than the links described in the NDSSL). In the NDSSL, where contacts are measured in seconds of co-location only, average degree is much more free to scale with population, as the social commitment needed to make this contact is low. Most literature on ERGMs do not handle tie growth in a regime like the NDSSL, as the literature has been previously devoted primarily to either friendship or sexual contact networks. Developing strategies to model highly connected networks like the NDSSL with ERGMs is an open problem.

Our effort to train an ERGM on the NDSSL data using this snowball sampling procedure and including both node-level characteristics and node matching characteristics failed. The ERGM estimation procedure became "degenerate," i.e., the maximum pseudo-likelihood procedure failed to find a set of parameters that matched the input graph. Triangles are a primary source of degeneracy for ERGMs. In this case, the model has a hard time distinguishing between ties that occurred at the same time (which are most/all part of a triangle) and ties that occurred apart from time (not triangles). This lopsided distribution likely makes the ERGM parameter estimation procedure oscillate between too many and too few triangles, never finding a set of parameters that approaches MPLE convergence.

This example of ERGMs applied to NDSSL shows both the versatility in the modeling approaches available with ERGMs but also its difficulty in practical application. ERGMs require careful thought into parameterization, MCMC sampling constraints, and network sampling considerations. Modern methodological research on ERGMs is primarily concerned with shortcuts to full network estimations using various techniques. While this is not a complete summary of ERGM techniques for large network estimation, it does demonstrate that successful ERGM techniques for large network modeling is far from settled. 

\section{Network Fusion \label{sec:fusion}}
In this section we address the fourth and final aim described in Section \ref{sec:aims}, network fusion. As mentioned there, the term `network fusion' represents a broad range of fusion approaches applied to network data. Here we describe two such approaches. Unfortunately, we were not able to implement these on the NDSSL and FluPaths data, unlike the other three key aims. However, we nonetheless hope that this discussion will be useful for policy-motivated network fusion problems, and we also hope to return to this subject in future work.

Although increasingly common, network datasets may not be available for every population of interest. For example, the NDSSL data describes the city of Portland, OR, but what if we were interested in another city, say Nashville, TN? Suppose the spread of an epidemic in Nashville was to be simulated using an agent-based model, and to initialize and inform the simulation an attributed graph analogous to the NDSSL dataset was required. A natural solution would be to learn a generative model using the NDSSL data, and to then modify it in some way to generate a network that is broadly consistent with the population of Nashville. More concretely, suppose that only the marginal distributions of the Nashville population are known - for example the distribution of gender, age, income, and other demographic variables. An artificial population can be generated by sampling these marginals, in other words by modeling the node attribute distribution $P(\bm{v})$ as:
\begin{equation}
    P(\bm{v}) = P(v_{\text{gender}}) \times P(v_{\text{age}}) \times P(v_{\text{age}}) \times ...
\end{equation}
Unfortunately, this will not capture the correlations amongst the demographic variables, but this is the best that can be done if only the marginals are known. Then, once an artificial population has been generated, a random network of connections between the individuals in the population can be initialized, and the iterated link prediction algorithm of Section \ref{sec:generative} may be used to re-wire the connections so that the probability that two nodes are connected agrees with the prediction of the GAE link prediction model. This procedure creates a new attributed network that matches Nashville's demographics and where nodes are connected similarly to how they are in Portland. This can therefore be thought of as a simple extension of a generative model into a method for network fusion.

A second common fusion problem we encountered during our expert interviews was the question of how to fuse two network datasets with only partially overlapping variables. For example, the FluPaths data contains many node attributes that are not included in the NDSSL data. Is there some way to impute these values in the NDSSL data in a way that is consistent with the FluPaths data? One way of interpreting this question is that the marginals of the variables added to the NDSSL data should match the marginals of the FluPaths data, and the mixing matrices should match as well. This provides one simple but natural interpretation of what it means for the fused result to be consistent with both the NDSSL and FluPaths data. Within the context of the iterated link prediction algorithm of Section \ref{sec:generative}, we can approximately enforce a target mixing matrix through the addition of a regularization term to the link predictor loss function Eq.~\ref{eq:link_pred_loss}: 
\begin{align}
    \label{eq:link_pred_loss_regulated}
    \mathcal{L}(\bm{\theta}) &= - \frac{1}{|E_{\text{train}}|} \sum_{(i,j) \in E_{\text{train}}} \left( A_{ij} \ln \sigma(\bm{z}_i^T \bm{z}_j) + (1 - A_{ij}) \ln\left(1 -  \sigma(\bm{z}_i^T \bm{z}_j) \right) \right) \\
    &\qquad \qquad - \lambda \sum_{a,b} \mathcal{M}^{\text{target}}_{ab} \ln \mathcal{M}^{\text{GAE}}_{ab}(\bm{\theta}) \nonumber \,.
\end{align}
The second term is the cross-entropy between the target mixing matrix $\mathcal{M}^{\text{target}}_{ab}$ and the mixing matrix predicted by the GAE model $\mathcal{M}^{\text{GAE}}_{ab}(\bm{\theta})$, and $\lambda$ is a hyper-parameter controlling the magnitude of the regularization. For this approach to work, the GAE mixing matrix must be differentiable, and so it should be computed using the raw probabilities output by the model, rather than on the sampled adjacency matrix. Letting $v_i$ denote the categorical node attribute variable of interest, the model mixing matrix is
\begin{equation}
    \mathcal{M}_{ab}^{(\text{GAE})}(\bm{\theta}) := \frac{1}{2} \left(1 + \delta_{ab} \right) \frac{\sum_{ij} \delta_{v_i a} \delta_{v_j b} \sigma(\bm{z}_i^T \bm{z}_j)}{\sum_{ij} \sigma(\bm{z}_i^T \bm{z}_j)} \,.
\end{equation}
In words, the numerator is the sum of probabilities for all potential edges between $a-$ and $b$-valued nodes, and the denominator is the sum of the probabilities for all potential edges in the network. The coefficient accounts for the undirected nature of the graph. 

\section{Outlook \label{sec:outlook}}
Recent efforts in developing simulation-based models of complex adaptive socio-economic systems have shown that there is a growing need for methods that can better inform these models. In particular, these models increasingly require large-scale synthetic populations that can easily be changed and made representative of different geographical areas of the United States (U.S.), and include socio-economic, demographic, and behavioral information. In addition, simulation models increasingly include social influences and how behaviors spread. Therefore, they require network and relationship information about how people interact socially, geographically, and over time. However, one single population or network data set rarely contain all the required information needed to inform these models. Instead, it is often the case that the required information is contained in different datasets and needs to be combined to construct the representative starting conditions of the population to be modeled. 

Recent agent-based simulation models (ABMs) generally include both the effects of networks and influence as well as behavioral rules that are grounded in cognitive-behavioral science. Two examples that fit this description are (i) a model of income tax evasion and risk perception~\cite{vardavas2019rand} and (ii) a model of the coupled dynamics between seasonal influenza transmission and yearly behaviors to vaccinate for the flu~\cite{vardavasModelingInfluenzaVaccination2013}. Both models account for key socio-economic and demographic heterogeneities affecting mixing patterns, social influences, risk perceptions, and behaviors and have been informed by tailored surveys that follow a nationally representative population through RAND’s American Life Panel (ALP). These surveys included social network components, and the datasets contain a rich set of information about each respondent’s social network. Both ABMs required representative synthetic populations of a U.S. city that included (i) merged a large-scale network dataset containing network structure and socio-demographic data with (ii) the ALP survey data containing behavioral ego-centric information on each respondent’s network. In other words, these ABMs required data fusion methods that are capable of merging network datasets from different sources. Moreover, they required methods to scale down the networks while retaining the network structure and key statistical features describing individual features/attributes and how they mix. Although these projects used some simple machine-learning and ad-hoc approaches for this purpose, they suffer from various limitations including being slow and not easily scalable nor generalizable. 

\paragraph{Significance of the Project}
This project was conceived by thinking through the needs and limitations of the methods used by these two ABMs in informing its starting population and the underlying network. The recent success of deep neural networks at generating artificial images after being trained on large datasets of real images as well as the rapidly growing application and field of graph machine learning in generating new graphs lead us to explore these approaches as new cutting-edge methods that can be applied to the types of problems we face in informing our simulation models. 

The significance of the research presented in this report goes beyond the applications to the two ABMs mentioned above. There are many examples of ABMs that consider the diffusion and transmission dynamics of social networks. Other simulation models such as microsimulation models generally do not include the effects of influence and the spread of behaviors, and therefore do not need network data sets. However, there are clear extensions of microsimulation models that can be converted to ABMs to include social influence effects and would benefit from the methods developed in this report.  For example, there is a growing interest in extending microsimulation models to ABMs to include network influences. The first example is RAND’s Comprehensive Assessment of Reform Efforts (COMPARE) microsimulation model that creates one of the most comprehensive models to date for evaluating proposed changes to the United States health care system~\cite{eibner2012small}. Observations of health care choices and health outcomes in one's social network are known to influence health care decisions. Together with network datasets, the methods developed in this project could help inform an ABM version of COMPARE. A second example is the RAND Alcohol Policy Platform (RAPP) where modeling alcohol initiation could be extended to include social network influence effects~\cite{rutter2019rand}. 

\paragraph{Summary of Accomplishments}
In Section \ref{sec:aims}, we presented the aims of this project. To better inform and broaden our aims, we conducted in-person interviews with RAND policy researchers working on applications of social networks and network analysis across different domains. Our interviews allowed us to identify the needs for new methods in social network analysis and how they align with those associated with informing the ABMs. Specifically, they allowed us to gauge the need and interest in using deep neural networks methods. We synthesize the results of the interviews in Table~\ref{tab:RANDresearch} in Appendix~\ref{app:interviews} and identify six challenges: (i) synthetic network generation, (ii) rescaling networks, (iii) data imputation, (iv) network fusion, (v) network dynamics, and (vi) interpretability. Our research presented in this report has focused on developing new methods to meet the first four of these six challenges. In Section \ref{sec:generative} we presented a novel algorithm, Graph Generation by Iterated Link Prediction, that used graph neural networks to simultaneously address synthetic network generation and rescaling aims. We also showed in Section \ref{sec:ergm} that ERGMs, a non-deep learning class of models often used in social network science, could also address these aims in principle, although in practice they do not scale well to large networks. We addressed the problem of imputing missing data at both the node and edge levels in Section \ref{sec:predictive}. Taken together, these methods use different population and network datasets to generatively model large-scale socio-centric synthetic networks that are representative of different U.S. cities, and which can be directly used to inform the starting conditions of current and future simulation models. 
 
Network fusion represents the most ambitious of these challenges. This is partly because this problem can have different interpretations and can easily be ill-defined. For example, the intended outcome of network fusion when the network datasets have very different network structures can often be unclear. A fusion method can produce a network with a structure that is not representative of either of its parent network datasets. For our project, we considered network fusion between a large-scale socio-centric network dataset and an egocentric dataset, and the goal was to keep the network structure of the socio-centric network while merging in network node features containing behavioral information from the egocentric dataset. Section \ref{sec:fusion} describes a methodological approach for this last challenge that builds off of the methods we used and developed for the previous three challenges. %

\paragraph{Broader Impacts}
Generating new synthetic networks using different network datasets is an emerging new problem that has applications in informing the starting population of agent-based simulation models of complex socio-economic systems. Until recently, the single and most apparent approach towards this goal would be to use ERGMs. Unfortunately, ERGMs do not scale well to large-scale networks such as the NDSSL network. One workaround is to apply ERGMs to smaller samples of the full network, and we explored this in Section \ref{sec:ergm}. Although the ERGMs do a good job capturing the statistics of the smaller samples, the samples themselves exhibit significant differences from the full large-scale network. 
Our research has shown that deep generative models applied to social networks provide a powerful new approach that can be used to overcome these challenges and train generative models over large-scale network datasets. Our code and workflows used in this report have been made internally available to RAND researchers and are available at \url{https://code.rand.org/hartnett/dgmnet}. Together with this report, these workflows can be used by interdisciplinary researchers interested in social network sciences as an introduction to the most recent advances in applying neural networks and machine learning algorithms to network science. Consequently, we think that this report has helped bridge the gap between the machine learning research community and the community of quantitative public policy researchers that use social networks.     

\paragraph{Limitations and Future Work}
The research goals of this project were ambitious and interdisciplinary, and as with many ambitious projects that explore new methods connecting disparate research communities, it suffers from a few limitations. The most evident limitation is that, while a method for network fusion is described in general in Section~\ref{sec:fusion}, no results are shown for the fusion of the NDSSL and FluPaths data.
Reasons for this include time-constraints, research interruptions due to other project requirements connected to the COVID-19 pandemic, and the many different approaches and methods we explored to accomplished the first three aims of this project. However, we plan to continue working on this task and update this working report as these tasks and new methods are completed. Another limitation is that our project did not include all the information provided by the NDSSL network structure when generating new synthetic networks. Most notably, we neglected to take into account the edge features, including the edge weight describing the strength and duration of the interaction. Moreover, each edge represents an interaction that occurred under a given activity or mixing mode (e.g., household, work, leisure, etc., as shown in Figure \ref{fig:contact_frequency_duration}) and in a given geographical location. Our methods do not include this information, although in many cases they could be easily extended to do so. Finally, the generative algorithm introduced in Section \ref{sec:generative} produces graphs with far fewer triangles than the NDSSL data it was trained on. We expect that this issue can be rectified with additional work.

In addition to addressing the above shortcomings, there are two obvious areas for future work. Our interviews with policy researchers presented in Appendix~\ref{app:interviews} led to the identification of six methodological aims, and due to the scope of the project, we were only able to consider four of these. It would be interesting and useful to pursue the remaining two aims of network dynamics and interpretability. 
Future projects could apply novel deep learning methods to generate new synthetic networks and examine how they change in time. One clear example of the need for these methods is to describe vital dynamics in social networks. The goal here is to generate large-scale socio-centric networks that include the aging of individuals in the population such that the network structure and key statistical properties of the network do not change over time. As another example, large-scale network data sets such as those created from mobile phone data could be used and analyzed by deep neural networks to learn and find patterns in the way people change how they mix over time. This is especially the case with the latest datasets describing how the COVID-19 pandemic has changed the mixing patterns and social network structures.  

Model interpretability of deep neural networks is a long-standing limiting factor for applications requiring explanations of the features involved in modeling. Although deep learning models can achieve high accuracy, this often comes at the expense of rather inscrutable models that are often described as black boxes. The applications that motivated the research presented in this report do not require interpretability. Even so, in Section \ref{sec:ergm} we have compared our deep-learning methods to the more established and interpretable method of ERGMs. In future work, we plan to expand on the comparisons between ERGMs and deep-learning methods to develop a better understanding and interpretation of the function of each of the deep hidden layers.

We believe that over time the research initiated by this project could have a significant impact on social network research and simulation modeling applied to public policy. Our work presented here represents the first step towards a broader set of ideas and applications. There are many exciting possibilities for machine-learning methods based on deep neural networks when coupled with simulation models and decision-support that go beyond the aims of this research project.   

\clearpage
\subsubsection*{Contributions}
Gavin S. Hartnett (Ph.D.) and Raffaele Vardavas (Ph.D.) were joint Principal-Investigators (PIs) for this project, designed and managed the project, and contributed the most in the preparation of the report. 
Michael Chaykowsky (MSc) and Gavin S. Hartnett (Ph.D.) lead the technical analysis, including the development of the generative algorithm and the engineering of the many workflows referenced in the report. Federico Girosi (Ph.D.) and Osonde Osoba (Ph.D.) acted as advisors providing directions and recommendations in the development of the deep learning algorithms and the analyses.
David Kennedy (Ph.D.) acted as an advisor in relating the algorithms to the social network science field and literature. 
C. Ben Gibson (Ph.D.) led the development and analyses of the Exponential Random Graph Models.
Lawrence Baker (MSc) conducted the interviews with RAND researchers to find how deep-generative models for networks can be applied over a wide range of projects carried out at the RAND Corporation. 

\subsubsection*{Acknowledgements}
We wish to thank the Method-Lab RAND-Initiated Research for funding and supporting this project. In particular, we thank Dr. Lisa Jaycox and Dr. Susan Marquis. We also thank Dr. Jeanne Ringel for reviewing this report. We thank Dr. Andrew Parker, Dr. Sarah Nowak, and Dulani Woods at RAND for discussions about the FluPath ALP egocentric datasets. We thank Ms. Chamsi Hssaine, a Ph.D. student from the School of Operations Research and Information Engineering Cornell University who was a summer associate at RAND in 2018 and did some preliminary work, and we thank Dr. Christopher S. Marcum, a staff scientist and methodologist at the National Institute of Allergy and Infectious Diseases who was a postdoctoral fellow at the RAND Corporation in 2012 and worked on some preliminary ideas. We further thank Professor Xavier Bresson from the School of Computer Science and Engineering Data Science and AI Center, Nanyang Technological University (NTU), Singapore for the interactions and discussions on deep learning and GCNs, and Professor Madhav Marathe of the Network Systems Science and Advanced Computing Division, Biocomplexity Institute at the University of Virginia (previously the director of the NDSSL at Virginia Tech.) for discussions about how these methods can be used to inform simulation models. We gratefully thank the National Science Foundation’s Interdisciplinary Behavioral and Social Science (IBSS) Research program (Award Number 1519116), the National Cancer Institute (R21CA157571), and the National Institute of Allergies and Infectious Diseases (R01AI118705) for providing support in projects that led to preliminary work and ideas that motivated this project. Lastly, we would like to acknowledge the \textit{PyTorch Geometric} Graph Neural Network library \cite{fey2019fast} which was very useful in our software implementations.

\clearpage
\appendix

\section{Methodological Needs in Public Policy Network Science \label{app:interviews}}
The field of public policy research has grown and evolved in response to the rapid growth of large and complex datasets. These new datasets have expanded the questions that policy research can hope to address and opened up many new exciting frontiers of research. Meanwhile, the landscape of policy problems typically studied by researchers has not remained constant. The myriad of secondary and tertiary challenges associated with the COVID-19 pandemic serves as a good example of this general phenomenon. These two effects, more data and a shifting policy landscape, have led to a growing need for new methodologies in social network science. To assess this need, we conducted in-person interviews with 13 RAND policy researchers working on applications of social networks and network analysis across different domains. As a result of these interviews and our understanding of existing methodological gaps, in this section we detail a list of general aims, or goals, that we would like to address. We found it useful to synthesize the results of the interviews into the six discrete aims detailed in Table~\ref{tab:RANDresearch}.

\begin{table}[!htbp]
\footnotesize
\rowcolors{1}{blue!30}{blue!10}
\centering
\caption{\label{tab:RANDresearch} Overview of techniques and application to RAND research. }
    \begin{tabularx}{1.0\textwidth} { 
      | >{\raggedright\arraybackslash}X 
      | >{\raggedright\arraybackslash}X 
      | >{\raggedright\arraybackslash}X
      | >{\raggedright\arraybackslash}X|
      }
     \hline
     \textbf{Challenge} & \textbf{Description} & \textbf{Motivation} & \textbf{Applications} \\
     \hline \hline
    
    \textbf{Synthetic Network Generation} & Given a set of graphs (or a set of subgraphs taken from a graph), learn the underlying probability distribution from which graphs are taken and generate more samples from the distribution. & Creating large sample sizes. Creating new graphs with modified statistics or joint distributions. Inference based on the probability distribution of observed graphs. De-identification of network data. & Creating synthetic versions of kill chain networks for wargaming. Identifying if two networks are statistically significantly different from one another. \\ \hline
    
    \textbf{Rescaling Networks} & A subset of synthetic networks, in which the graphs generated have a different number of nodes than the original graphs. This may require a specification of how statistics are expected to change with population. & Scaling down graphs to reduce computation time for simulations. Scaling up graphs to increase sample size or analyze emergent properties in larger populations. & Reducing the size of an influenza transmission network to allow more runs of an agent-based model in a set timeframe. \\ \hline

    \textbf{Data Imputation} & Given a graph or set of graphs, impute missing edges, nodes, edge properties or node properties. & Using more complete graphs or subgraphs to impute missing data in less complete graphs or subgraphs. Use as part of a data collection strategy in which easily observed data is used to infer difficult to observe data. & Imputing unreported associations in a network of financial relationships. Inferring the reach of information on social media based on observed interactions (e.g. retweets, replies and mentions). \\ \hline
    
    \textbf{Network Fusion} & Given two or more graphs, which can be considered to represent partial views of a more complete graph, estimate the more complete graph. & Joining datasets in the case where there is no shared ID between nodes, or where one dataset only contains information on a subset of the nodes. & Combing ego networks (i.e. surveys) with a larger graph. Relevant to both analyzing American Life Panel data and the spread of influenza in Portland. \\ \hline
    
    \textbf{Network Dynamics} & Given timesteps in a dynamic graph, predict the graph at other timesteps. & Analyzing the process through which the graph evolves. Predicting future behavior based on initial dynamics. & Using graph dynamics from one part of the graph to predict changes in another section. Reconstructing and predicting the spread of fake news stories across social media networks. Predicting changes in alter-alter ties from changes in ego-alter ties \\ \hline
    
    \textbf{Interpretability} & Use learned parameters to better specify or understand the graph. & Understand which features of a graph are most important. Specifying the set of statistics sufficient to specify a graph. & Identify what factors influence the spread (i.e. probability of an edge) of public awareness campaigns. \\ \hline
    \end{tabularx}
\end{table}

\clearpage

\section{Mathematical Glossary \label{app:glossary}}
To keep this report as self-contained as possible, here we provide a glossary of mathematical concepts used throughout the report.

\paragraph{Graph} To start, a \textit{graph} $G$ is a pair $(V,E)$ where $V$ is the set of vertices (also referred to as nodes), and $E$ is the set of edges. A graph may be either \textit{undirected} or \textit{directed}. In the first case, an edge consists of an unordered pair of vertices, whereas in the second case an edge consists of an ordered pair. Throughout this report, we denote the number of vertices by $N = |V|$ and the number of edges by $M=|E|$. Also, we use the terms graph and network interchangeably, as well as vertices/nodes and edges/links. Many mathematical concepts are useful for studying graphs. Given the current focus on using machine learning for network science problems, here we focus on those concepts that are related to the numerical representation of graphs. 

\paragraph{Edge-List} An \textit{edge-list} $E$ is a data structure used to represent a graph as a list of its edges. Each row entry represents an edge on the graph. It has a minimum of two columns, giving the id of the start and end vertex. A third column provides information on the edge weight, and additional columns can provide attributes of the edge. 

\paragraph{Self-referring edges} A \textit{self-referring edge} (also known as a \textit{self-loop}) connects a vertex with itself. Generally, a graph does not have self-referring edges with loops connecting vertices to themselves, and vertices can be connected by at most one edge. However, including a self-referring edge in the mathematical description of a graph can be important when describing graph properties. 

\paragraph{Adjacency matrix} 
Network ties are often recorded as a square actor-by-actor matrix known as the \textit{adjacency matrix} and denoted by $\bm{A}$.
It provides the same information contained in the first three columns of the edge list. However, it provides a poor data structure for representing sparse graphs, in which case the edge list $E$ is typically used.

A value in a given cell of the adjacency matrix indicates that there is a tie from the row actor to the column actor
The adjacency matrix of an unweighted graph is a $N \times N$ matrix, with $A_{ij} = 1$ if the edge connecting $i$ and $j$ is present, otherwise $A_{ij} = 0$. For a weighted graph, the non-zero elements contain the edge weights. The adjacency matrix for a non-directed graph is symmetric, and diagonal entries indicate the presence of self-loops. Generally, the matrix $\bm{A}$ denotes the adjacency matrix with no self-referring edges. The adjacency matrix for an unweighted graph with self-referring edges is then defined as $\tilde{\bm{A}} = \bm{A} + \bm{I}$, where $\bm{I}$ represents the identity matrix.

The sum of the rows of the adjacency matrix gives a centrality measure known as the \textit{strength centrality} of each vertex id.  For an unweighted graph the sum gives the \textit{degree centrality} measure of each vertex id, which is simply the number of vertices it is connected to. The degree centrality can be expressed as a diagonal matrix $\bm{D}$ where $D_{ii}= \sum_j A_{ij}$ or by the vector $\bm{d}= \bm{D}\cdot \bm{h}$ where $\bm{h}=(1,1...1)^T$ is known as the \textit{homogeneity vector}. The \textit{symmetric normalized adjacency matrix} of an unweighted graph can be found by dividing each row of the adjacency matrix by the degree centrality of the corresponding node. This can be shown to be equal to $\bm{D^{-{\frac {1}{2}}}AD^{-{\frac {1}{2}}}}$. 

The adjacency matrix $\bm{A}$ for an unweighted and undirected graph can be used to describe the diffusion of a non-conserved quantity such as a rumor spreading over a network in discrete-time. Consider a vector $\bm{v}_0$ of length equal to the number of nodes on the network, taking a dichotomous variable representing which nodes provide the initial source of a rumor. We can find the nodes that have been exposed to the rumor $\bm{v}_n$  after $n$ iterations by computing $\tilde{\bm{A}}^n\cdot\bm{v}_0$. The use of $\tilde{\bm{A}}$ instead of $\bm{A}$ is because self-connections are needed to account for the fact that nodes that act as the sources of the rumor remember the information of the rumor in the next iteration. 

\paragraph{Attributed graphs}
In the applications we consider, the actual network of connections only represents part of the story. There may be additional variables associated with the vertices and edges of the graph. An example that arises in public policy research is the problem of understanding how an individual's decision to vaccinate is affected by the decisions of their social circle. In this case, it would be useful to employ a richer dataset than just the underlying network of social connections and to also consider individual-level demographic data such as age, gender, ethnicity, etc., as well as data on the nature of the social connection (friend, family, coworker, etc.) \cite{bruine2019reports}. To this end, suppose that in addition to the graph $G$ we are also provided with $n$ variables for each vertex and $m$ variables for each edge. These may be real-valued, discrete, categorical, or a mixture of each type. The node variables may be organized to form a $N \times n$ node-attribute matrix $\bm{V}$, and the edge variables may be organized to form a $M \times m$ edge-attribute matrix $\bm{E}$. Thus, in this work, we shall consider a \textit{node and edge-attributed graph} $\mathcal{G}$ to be a tuple consisting of an ordinary graph together with the vertex and edge variables:
\begin{equation}
    \mathcal{G} = \left(G, \bm{V}, \bm{E} \right) \,.
\end{equation}
Similarly, we may also consider just node-attributed graphs or edge-attributed graphs, in which case there will be only one feature matrix. Lastly, we use the notation $\mathcal{A}, \mathcal{V}, \mathcal{E}$ to denote the space of all possible adjacency matrices, node attribute matrices, and edge attribute matrices.

\paragraph{Random graphs}
In network science one is often interested in \textit{random graphs} for which the graph structure is determined according to some random process. In this case, the graph may be thought of as a random variable, distributed according to some probability distribution. Of course, there is an infinite number of graphs, and so often this distribution is taken to have support over some finite \textit{ensemble} of graphs. For example, we might wish to consider a distribution defined over graphs of $N$ nodes. The simplest example of distribution over this ensemble is the Erdős–Rényi (ER) $G(N,p)$ model \cite{erdHos1960evolution}, where each possible edge in a given graph is present with probability $p$. More complex models that are often better approximations to real-world networks include small-world networks and scale-free networks \cite{newman2018networks}.

In this work, one of our main goals will be to \textit{learn} a random graph model that well represents real-world networks. This may be accomplished by performing some sort of statistical fit of the parameters of a random graph model. At first glance, the combinatorial explosion of the random graph ensembles suggests that this may be an intractable problem, even for graphs of modest size. For example, the ensemble of all graphs with $N$ nodes consists of $2^{N(N-1)/2}$ distinct graphs. For $N=10$, this amounts to roughly $10^{13}$, and for $N=20$, it grows to be $10^{57}$. However, as we will see, there has been much success in modeling these complex distributions using deep learning. %

\paragraph{Graph isomorphisms}
A very important mathematical consideration for the generative modeling of networks is the role of graph isomorphisms. Informally, two graphs are said to be isomorphic if they possess the same network structure. For example, Figure \ref{fig:isomorphic} below shows two isomorphic graphs with five vertices. The graphs appear to be quite different at first glance - each graph has a distinct vertex set, and the diagrammatic depictions of the graphs are also clearly different. The second difference is not intrinsic to the graphs and is simply a consequence of how the graphs have been visually represented. In contrast, the different vertex sets do reflect a real distinction, although upon inspection the two graphs can be seen to have the same network structure. More formally, this observation corresponds to the fact that the two graphs are isomorphic, as evidenced by the existence of a bijection between the vertex sets of the two graphs that maps the edge set of the first graph to the edge set of the second graph: $(\text{Alice, Bob, Charlie, David, Eleanor}) \rightarrow (\text{Alfred, Beatrice, Connor, Denice, Eric})$.
\begin{figure}[!ht]
	\centering
	\includegraphics[width=0.75\textwidth]{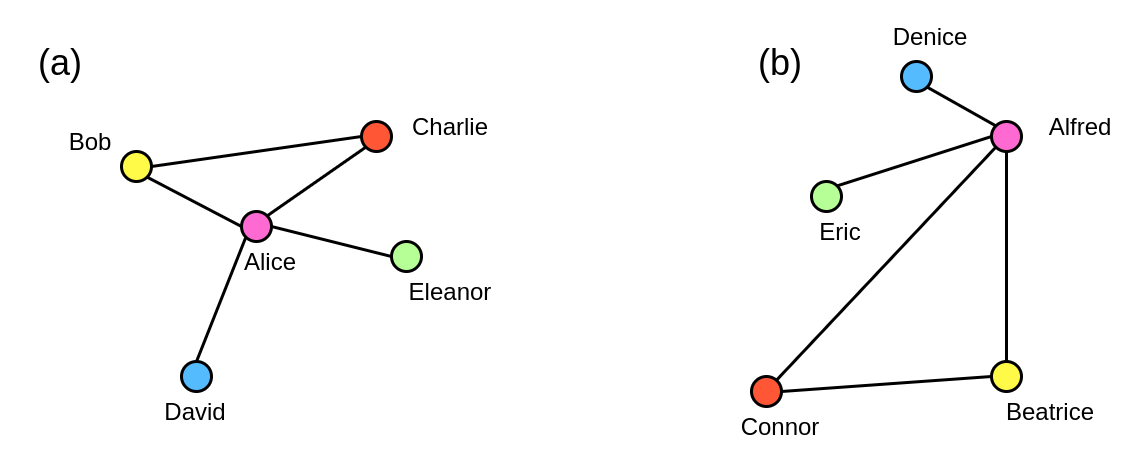}
	\caption{Two isomorphic graphs.}
	\label{fig:isomorphic}
\end{figure}

A closely related issue is that of the node ordering. Many of the methods we discuss rely in some way or other on the adjacency matrix. As discussed above, the $i,j$ entry of the adjacency matrix, $A_{ij}$, is a binary variable indicating the presence or absence of an edge between nodes $i$ and $j$. The adjacency matrix for a given graph depends on the arbitrary ordering of the nodes. For example, in the left graph of Figure \ref{fig:isomorphic} it might seem natural to order the nodes as: (Alice, Bob, Charlie, David, Eleanor), but any other ordering will be equally valid, and in general, a different ordering will lead to a different adjacency matrix. Under a permutation $\pi : \{1, ..., N\} \rightarrow \{1, ..., N\}$ of the node indices, the adjacency matrix will transform as
\begin{equation}
    \bm{A} \rightarrow \bm{P}_{\pi}^T \bm{A} \bm{P}_{\pi}  \,,
\end{equation}
where $\bm{P}_{\pi}$ is the associated $N \times N$ permutation matrix with indices $(P_{\pi})_{ij} = \delta_{i \pi(j)}$, where here $\delta_{ij}$ is the Kronecker delta. In terms of components, the above relation may be also written as $A_{ij} \rightarrow A_{\pi(i) \pi(j)}$. This transformation property may be used to provide an equivalent definition of graph isomorphism: two graphs $G, G'$ are isomorphic if and only if their adjacency matrices satisfy $\bm{A}' = \bm{P}_{\pi}^T \bm{A} \bm{P}_{\pi}$. 

These concepts may be simply extended to the node- and edge-attributed graphs. Under a permutation $\pi$, the node attribute matrix transforms as $\bm{V} \rightarrow (\bm{P}_{\pi})^T \bm{V}$, and the first index of the $M \times m$ edge attribute matrix $\bm{E}$ transforms the same way as the edge list, which itself transforms as $E \rightarrow \{ (v_{\pi(i)}, v_{\pi(j)}) | (v_i, v_j) \in E\}$. Two attributed graphs are said to be isomorphic if and only if their adjacency and attribute matrices differ by the action of a permutation. For generative modeling, the precise node labels of a graph and their ordering are irrelevant. Therefore, all the methods we consider assign equal probability weight to isomorphic graphs. This criterion provides an important constraint on the functional form used to construct the generative models. 

\paragraph{Laplacian matrix} The Laplacian is a symmetric  square matrix found from the adjacency matrix in which diagonal components are all non-negative (representing node degrees) while other components are all non-positive. It is defined as $\bm{L} = \bm{D} - \bm{A}$. The \textit{symmetric  normalized Laplacian matrix} of an unweighted graph is defined in a similar way to the adjacency matrix counterpart and is given by $D^{-{\frac {1}{2}}}L D^{-{\frac {1}{2}}}=I-D^{-{\frac {1}{2}}}A D^{-{\frac {1}{2}}}$. 

The Laplacian matrix $\bm{L}$ has useful dynamic properties that can reveal the structure of the network.  Two important properties of  $\bm{L}$ are that at least one of its eigenvalues is zero and all the other eigenvalues are either zero or positive. The ﬁrst property arises, because $\bm{L}\cdot \bm{h}=\bm{(D-A)}\cdot\bm{h}=\bm{d}-\bm{d}=0$, where $\bm{d}$ is the vector made of node degrees. Hence, $\bm{h}=(1,1...1)^T$, and known as the \textit{homogeneity vector}, is the eigenvector that corresponds to eigenvalue 0. The second property comes from the fact that the diagonal components are all non-negative while other components are all non-positive, and the sum over each row is zero. This makes $\bm{L}$ a symmetric, positive-semidefinite matrix which is important for where both the row and column sum of $\bm{L}$ are zero: $\sum_i L_{ij} = \sum_j L_{ij} = 0$. 

The properties of the Laplacian matrix $\bm{L}$ can be used to describe the diffusion of conserved quantities over a network such as the spread of a limited resource. This is in contrast to the adjacency matrix that is instead used to describe the diffusion of non-conserved quantities over the network. To understand this we first briefly revise diffusion over Euclidean space in continuous-time. This is described by the diffusion equation $\dot{C}=-\alpha \nabla^2 C$, where $C$ is our conserved quantity,  $\alpha$ is a diffusion constant, and $\nabla^2$ is the Laplacian operator.  Using the continuous-time diffusion equation as a model, the Laplacian matrix can be used to describe the network diffusion process in discrete time.  The first iteration of the diffusion process can be expressed as $-\bm{L}\bm{v}_0$. This operation removes the conserved quantity from the source nodes and distributes it to the alters. We can find how the conserved quantity spreads over the network after $n$ iterations $\bm{v}_n= [-\bm{{L}}]^n\cdot\bm{v}_0$.  Using the equivalence with the continuous-time diffusion equation the actual coefﬁcient matrix of the diffusion is $-\bm{{L}}$. This means that in contrast to the adjacency matrix,  the zero eigenvalue is dominant. The importance of this is that the corresponding dominant eigenvector tells us the asymptotic state of the network (i.e., the steady-state distribution of the conserved quantity). If the network is fully connected, then the conserved quantity will be homogeneously distributed asymptotic state given by the homogeneity vector. If the network is made of multiple connected components, diffusion can be modeled separately and independently on each component, converging to a different asymptotic value. Therefore, the asymptotic state of the whole network has as many degrees of freedom as the connected components.  Hence, an additional property of $\bm{L}$ is that the number of its zero eigenvalues corresponds to the number of connected components in the network. The smallest non-zero eigenvalue of $\bm{L}$ is called the {\it spectral gap} of the network and determines how quickly the diffusion takes place on the network. The spectral gap of the Laplacian matrix describes topologically how well the nodes are connected from a dynamic viewpoint.  It corresponds to the largest non-zero eigenvalue of $-\bm{L}$ used in the iterative process describing diffusion. When its value is large diffusion occurs faster.

\clearpage

\section{The NDSSL Sociocentric Dataset \label{app:ndssl}}
In this Appendix, we provide more details on the NDSSL data, including its network structure. The NDSSL data corresponds to a large, attributed graph describing the simulated activities of the population of Portland, OR. The nodes of the graph correspond to individuals, and the edges correspond to interactions between individuals. The attributed graph $\mathcal{G} = (G, \bm{V}, \bm{E})$ is comprised of three objects, the graph or network of connections $G$, the node feature matrix $\bm{V}$, and the edge feature matrix $\bm{E}$. We first discuss the node and edge features and then turn to a discussion of the network structure.

\begin{figure}[!ht]
	\centering
	\includegraphics[width=\linewidth]{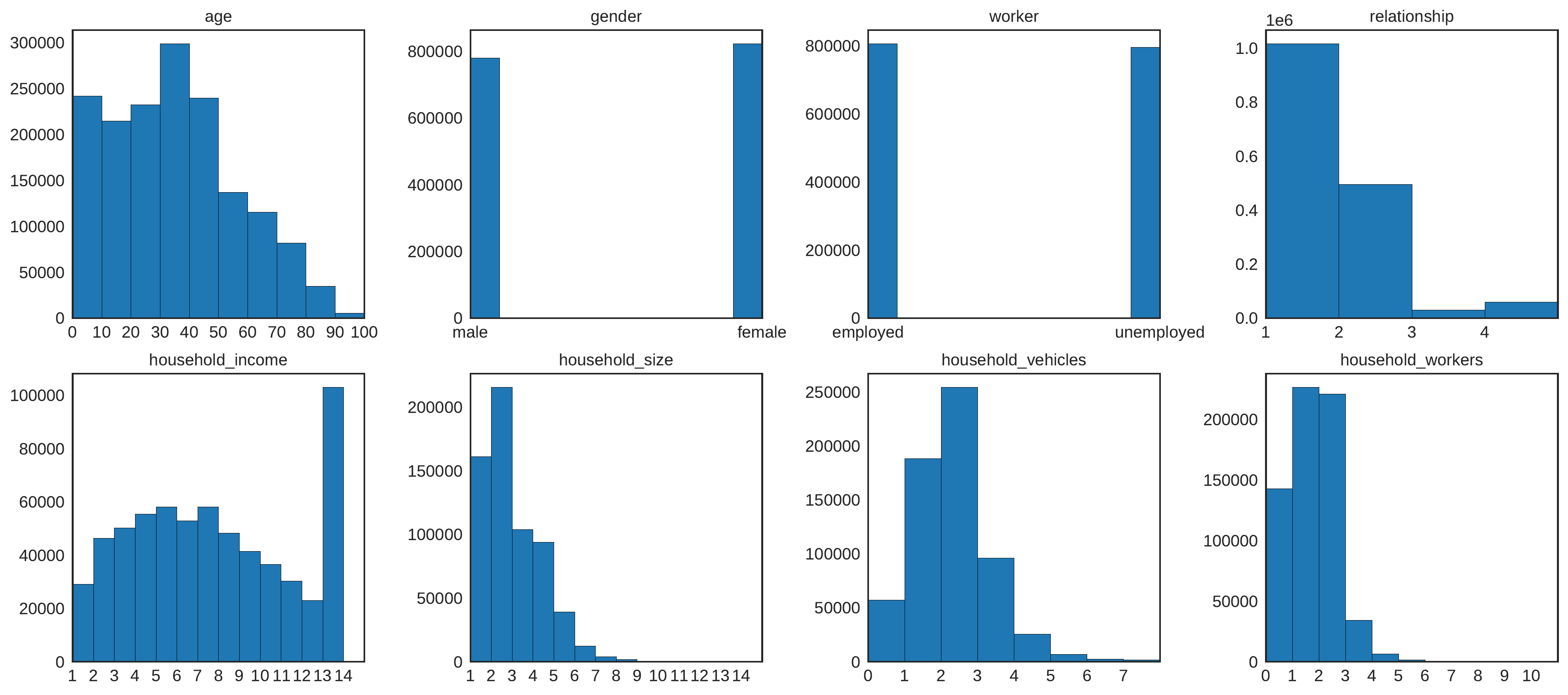}
	\caption{Node feature histograms.}
	\label{fig:node_histograms}
\end{figure}

\begin{figure}[!ht]
	\centering
	\includegraphics[width=0.8\linewidth]{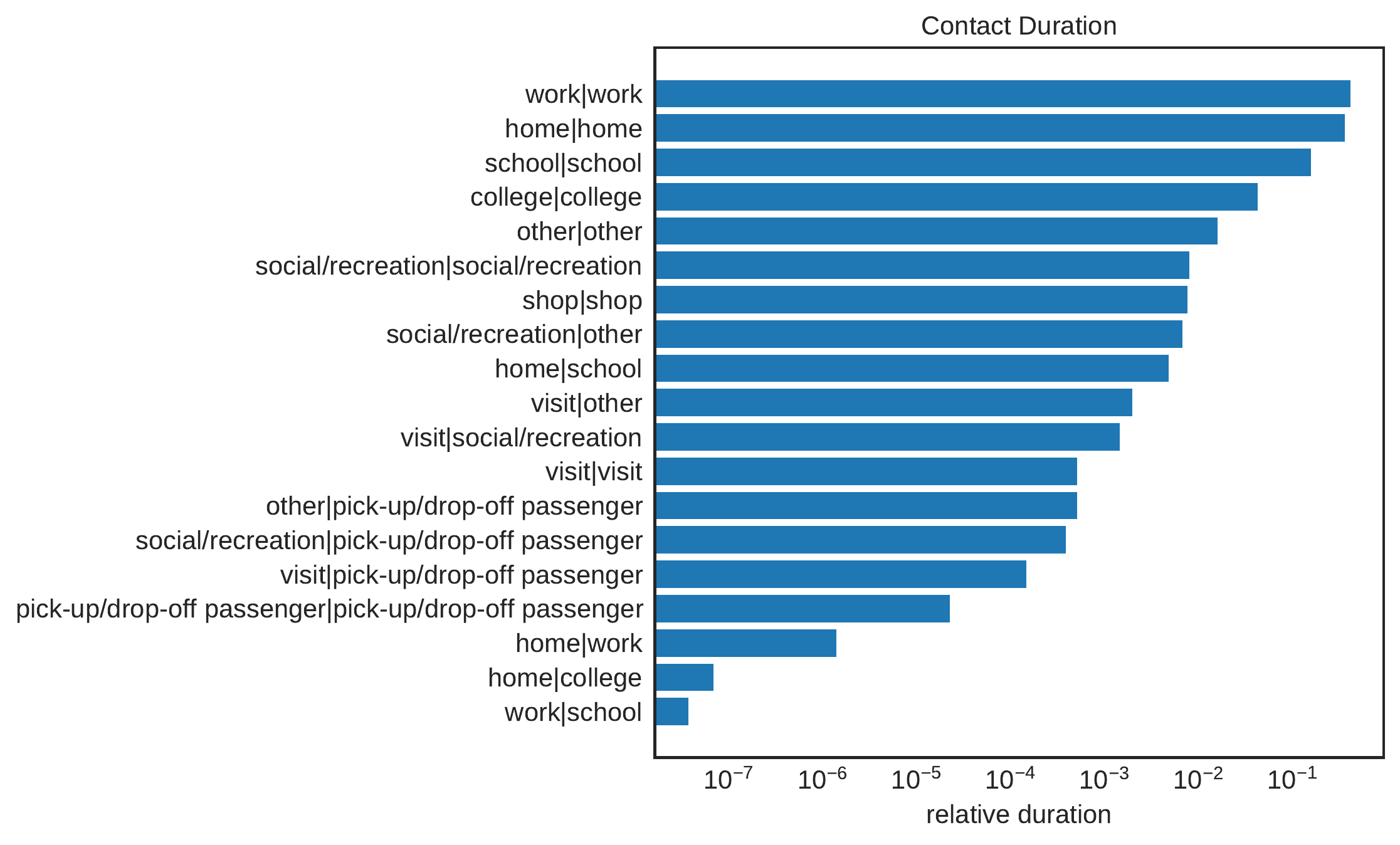}
	\caption{Duration of contact by activity type.}
	\label{fig:contact_frequency_duration}
\end{figure}

Table \ref{table:ndssl_features} lists the node and edge features. In Figure \ref{fig:node_histograms} we show histogram plots for the node features (excluding the id's and the zip codes). The top row shows the distribution of node features that are unique to the individual, whereas node features shared by all members of a household are displayed in the bottom row. The edge features are the duration of a person-person contact, and the activities of each person during that contact. The activity pairs, weighted by their relative cumulative duration throughout the data, are shown in Figure \ref{fig:contact_frequency_duration}.

\begin{figure}[!ht]
	\centering
	\includegraphics[width=1.0\linewidth]{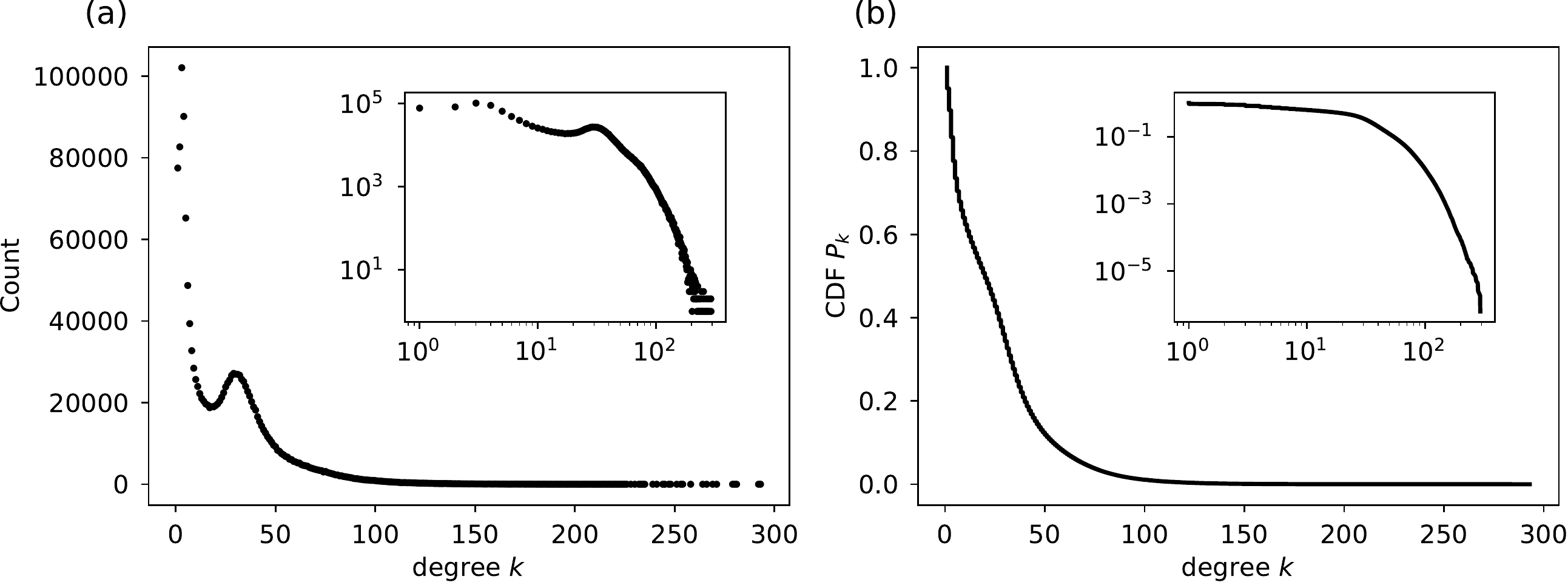}
	\caption{(a): The degree histogram for the NDSSL contact network. (b): The CDF function $P_k$.}
	\label{fig:degree_dist}
\end{figure}

The final component of the attributed graph is of course the graph itself, $G$. This is an undirected network capturing the person-to-person contacts. In Table ~\ref{table:ndssl_graph} we include some basic properties of this graph, such as the number of nodes, edges, and density. Importantly, the graph is not connected, and there are many degree-0 nodes (or isolates). However, the majority of nodes (97.7 percent) are contained in the giant component. Further insight into the network can be gained by examining the degree distribution. In Figure \ref{fig:degree_dist} (a) we plot the number of nodes with degree $k$. The most common degree is $k=3$ and the average degree is $24.73$. There are a large number of low-degree nodes, and as the degree increases, the count decreases slowly, with a slight bump around degree $~30$, and then the count decreases more rapidly after that. Thus, the likelihood of finding a high degree node decreases as the degree is taken to be large. 

\begin{table}[!htbp]
\rowcolors{1}{blue!30}{blue!10}
\caption{\label{table:ndssl_graph}NDSSL Social Contact Graph Properties}
\centering
\begin{tabular}{|c|c|}
 \hline
 nodes & 1,575,861 \\ \hline
 edges & 19,481,626 \\ \hline
 density & 1.57e-05 \\ \hline
 isolates & 25,469 \\ \hline
 giant component size & 1,539,920 \\ \hline 
 number of triangles & 388,850,385 \\ \hline
\end{tabular}
\end{table}

Many real-world networks exhibit this general behavior, with the degree distribution decaying rapidly as the degree increases. There has been much interest in networks where the decay obeys a power-law, i.e. 
\begin{equation}
    \label{eq:scalefree}
    p_k = C k^{-\alpha} \,,
\end{equation}
where here $p_k$ is the degree probability mass function, $\alpha$ is called the exponent of the power-law, and $C$ is a normalization constant. Networks obeying Eq.~\ref{eq:scalefree} are known as scale-free networks \cite{newman2018networks}, and many real-world networks across a large variety of disciplines have been claimed to be scale-free, although the actual prevalence of truly scale-free networks has been recently challenged \cite{broido2019scale}. In the context of epidemiological modeling, scale-free networks have been observed in the location-location contact network (as opposed to the person-person contact network) \cite{eubank2004modelling}. We can test to what extent the NDSSL contact-contact network is scale-free by performing a statistical fit to the power-law behavior. Of course, at most the data should exhibit a power-law tail, rather than obeying true power-law behavior over the entire range of degrees $k$. If $p_k$ obeys a power-law, then so too does the cumulative distribution function (CDF) $P_k$, the fraction of nodes with degree k or greater: $P_k = \sum_{k' = k}^{\infty} p_{k'}$. In Figure \ref{fig:degree_dist} (b) we depict the CDF on a log-log scale. Plotting a power-law on a log-log scale yields a linear function, and the plot is not very linear, except perhaps in the far right tail. The power-law behavior can be rigorously tested for using the statistical methodology described in \cite{clauset2009power}, which is nicely implemented by the \textit{powerlaw} Python package \cite{alstott2014powerlaw}. The result is that the tail is not well-described by a power-law distribution - in particular, a log-normal distribution yields a better fit of the tail. %

\begin{figure}[!ht]
	\centering
	\includegraphics[width=0.65\linewidth]{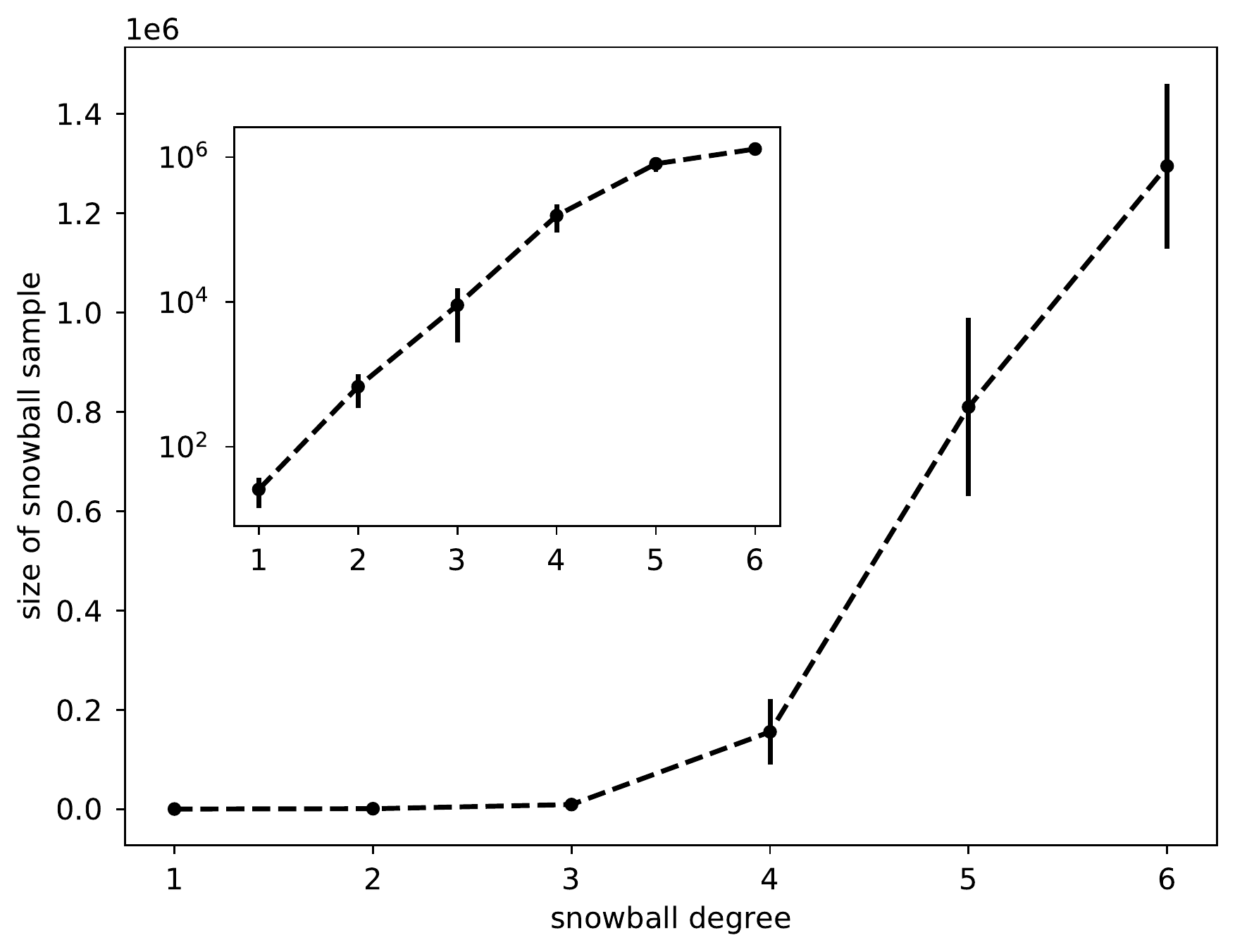}
	\caption{Exponential growth of snowball samples for a single randomly chosen seed node. The snowball degree is the number of waves, and the vertical lines represent standard errors computed with 100 samples.}
	\label{fig:snowball}
\end{figure}

Lastly, we considered the growth of snowball samples of the NDSSL graph. A snowball sample is obtained by starting with one or more initial seed nodes and then iteratively adding the neighborhood nodes to the sample. So, for example, a degree-3 (or 3-wave) snowball sample corresponds to all the nodes a distance of three or less from the initial seed nodes. Compared to other sampling approaches, snowball sampling does a good job capturing the local structure of a graph, and this approach was explored as a method for training ERGMs in Section \ref{sec:ergm}. However, in many graphs, it suffers from exponential growth. Indeed, Figure \ref{fig:snowball} depicts this growth as a function of the `snowball degree'. The inset shows that the curve is roughly linear on a log-linear scale, indicating exponential growth. Of course, as the sample grows to encompass the majority of nodes in the network, this growth tapers off.

\clearpage
\section{The FluPaths Egocentric Dataset \label{app:EgoFluPaths}}
\begin{figure}[!ht]
\centering
\begin{subfigure}
  \centering
  \includegraphics[width=0.5\linewidth]{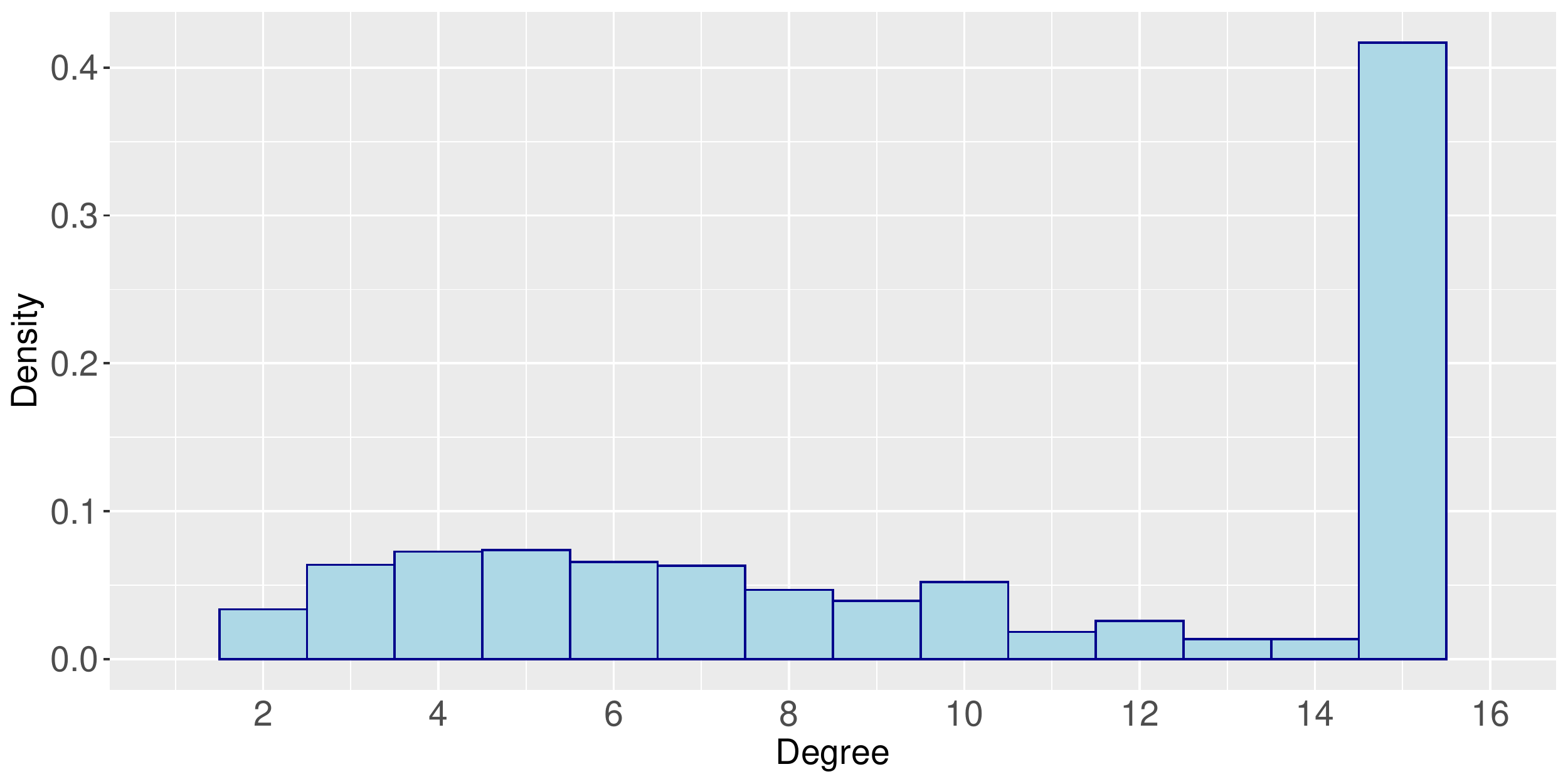}
\put(-230,110){(a)}
\end{subfigure}%
\begin{subfigure}
  \centering
  \includegraphics[width=0.5\linewidth]{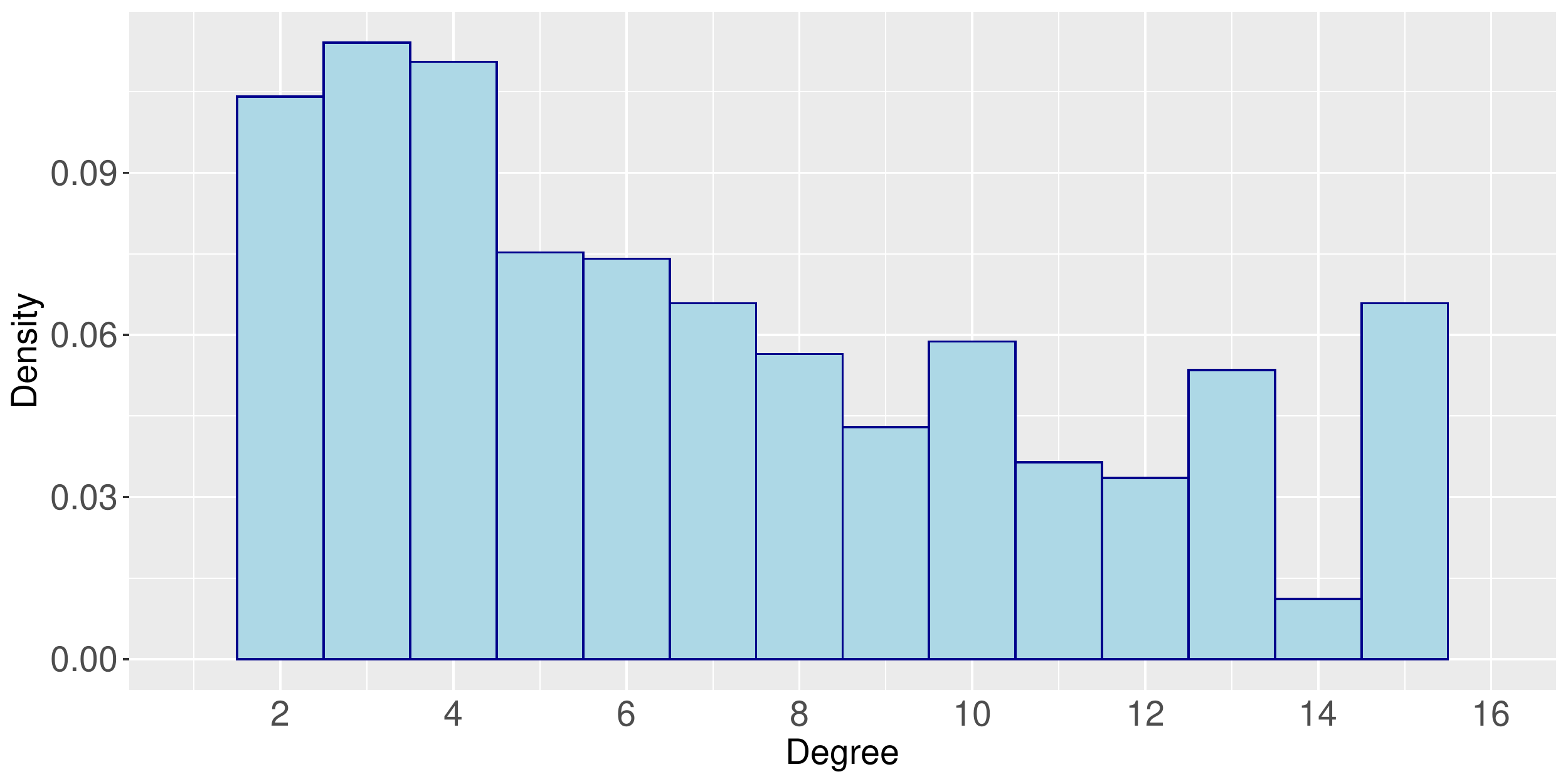}
\put(-230,110){(b)}
\end{subfigure}%
\caption{Degree histograms that consider (a) ego-alter ties and (b) frequent ego-alter ties.   \label{Fig:EgoDegree}}
\end{figure}

In this Appendix, we provide more details on the ALP FluPaths Egocentric data introduced in Section \ref{sec:EgoFluPaths}. As described in the sections that follow Section \ref{sec:EgoFluPaths}, one of our goals is to fuse the NDSSL sociocentric data with the FluPaths egocentric data. In particular, we want to generate large-scale representations of the NDSSL network for Portland, OR that includes node behavioral features contained in the FluPaths egocentric data such as attitudes towards vaccination and whether or not each individual in the network vaccinates for influenza. As shown previously in Figure \ref{fig:NDSSL_Projection}, networks have both a structure and features associated with the nodes and edges. When considering merging a sociocentric network with an egocentric network, we need to realize that they may have very different network structures. Therefore, when describing the fusing of the two types of networks, we need to specify that data from the sociocentric network takes priority in determining the resulting network structure. One measure of network structure is given by the distribution of the degree centrality. In Table~\ref{table:FluPaths_degree_summary} shown in Section \ref{sec:EgoFluPaths} we presented the summary statistics of the degree distribution of our egocentric networks. The table helps illustrate the difference in network structures between the NDSSL sociocentric network and the FluPaths egocentric network.  More details of the FluPaths network structure are revealed from the histograms of the degree distribution. Figure \ref{Fig:EgoDegree} shows the histograms of the degree distributions that consider (a) ego-alter ties and (b) frequent ego-alter ties. The degree distribution of the frequent ego-alter ties excludes alters that interact with the ego less frequently than once every six months, irrespective of face-to-face or non-face-to-face interaction. We note that many respondents in the survey provided fifteen alters, which represented the maximum number of alters they could list. However, when we look at the number of alters with which respondents have frequent interactions, defined as interactions that occur at least every other month by at least a non-face-to-face interaction, the number of alters decreases considerably. This is an important observation since behavioral influences and influenza transmission are not likely to occur for infrequent interactions.

\begin{wrapfigure}{r}{0.4\textwidth} %
\vspace{-12pt}
\begin{center}
	\includegraphics[width=0.90\linewidth]{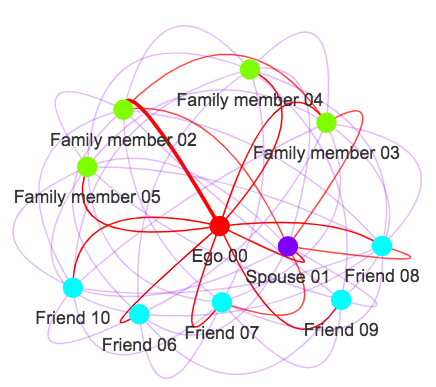}
	\caption{A visualization of the egocentric network of one of the respondents in the wave 1 FluPaths data}
	\label{Fig:FluPathEgoExample19}
\end{center}
\vspace{-10pt}
\end{wrapfigure}

\begin{figure}[!ht]
\centering
\begin{subfigure}
  \centering
  \includegraphics[width=0.5\linewidth]{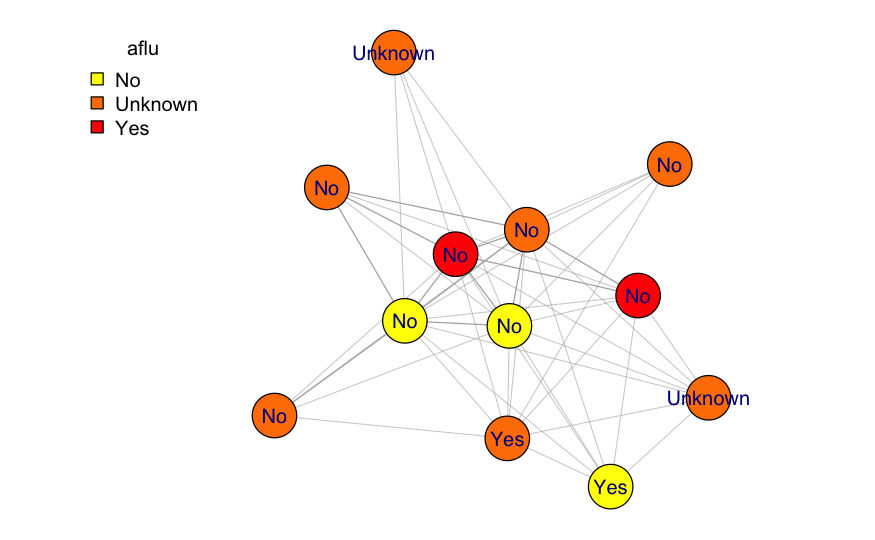}
\put(-20,140){(a)}
\end{subfigure}%
\begin{subfigure}
  \centering
  \includegraphics[width=0.5\linewidth]{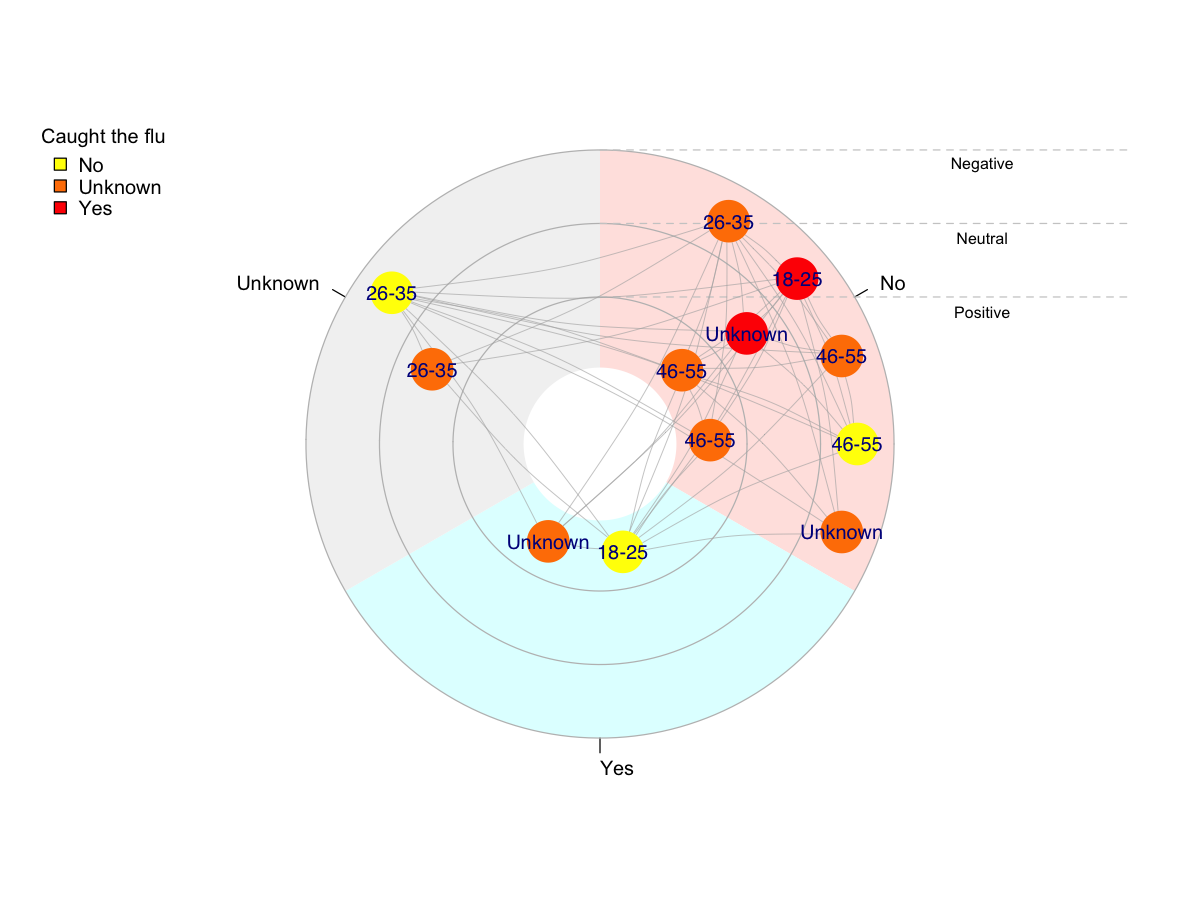}
\put(-20,140){(b)}
\end{subfigure}%
\caption{Alternative egocentric visualizations that do not display the ego. The first visualization (a) shows whether or not an alter chose to get the influenza vaccine (text overlaying each node) and whether or not they contracted the flu (node color). The second visualization (b) is a sociogram that positions the nodes around the center (representing the ego) and shows some of node features of the alters as described in the text.
\label{Fig:EgoVenn}}
\end{figure}

Having described the FluPaths network structure in a little more detail, we now turn to the FluPaths node features. Analyses of the network node features are important for our purpose as they reveal the way that the behavioral attributes contained in the  FluPaths egocentric networks can be merged into the large-scale sociocentric network.  Here we provide a brief description of some of the types of analyses and visualizations of egocentric data \cite{perry_pescosolido_borgatti_2018}. The main reason to visualize networks is to detect patterns to help generate hypotheses. Egocentric network diagrams can be used to compare the immediate surroundings of different kinds of egos. Powerful egocentric network visualizations show both node and edge features and arrange the nodes spatially in some meaningful way that relates to either the node or edge attributes. Figure~\ref{Fig:FluPathEgoExample19} shows an example of an egocentric visualization
from the FluPaths data that includes the ego. The ego is present in the center of the diagram and is a visual anchor that facilitates comparison across different egocentric networks. However, including all ego-alter ties makes the visualization very busy.  Figure \ref{Fig:EgoVenn} provides two examples of egocentric data that remove the ego from the visualization. In particular, Figure \ref{Fig:EgoVenn} (b) is an egocentric sociogram visualization generated using the {\bf egor} package in R, where the position of the alters depends on their attitude towards vaccination and on whether or not they vaccinated for influenza as reported by the respondent. Although the ego is not displayed, the center of the circular plot represents its virtual position and alters closer to the ego have positive attitudes towards vaccination and those further ways have neural and negative attitudes. The alter's influenza vaccination feature is displayed by their position in one of the three-prongs. This shows whether an alter did or did not vaccinate for the flu, or whether the respondent did not know whether or not the alter chose to get the flu vaccine. The sociogram also shows the age group feature of the nodes as an overlaying text and whether or not the alter contracted influenza by using a node coloring scheme.  

In Section \ref{sec:EgoFluPaths} we introduced two types of mixing-matrices and describe how they can be used to understand the level of associative mixing present in the egocentric network across different node features or combinations of these. Figure~\ref{Fig:EgoMixMatAgeVacc} in Section~\ref{sec:EgoFluPaths} showed examples of the two types of mixing matrices using the combination of age-group and influenza vaccination status. For influenza vaccination status the mixing matrix considers just two states, namely that they either did vaccinate or did not vaccinate for influenza. However, it is important to note that many respondents in the FluPaths data did not know the vaccination status of their alters. Therefore, a more accurate but more busy mixing matrix can be constructed using all the influenza vaccination statuses. Similar simplifications can be made when considering other node attributes. For example, we can choose to remove nodes with an unknown attitude towards vaccination as these nodes are unlikely to be influential on vaccination behavior. Moreover, the mixing matrix can depend on other simplifying choices that can be made. For example, we can include all edges connecting the nodes in the data or, as was done in Figure \ref{Fig:EgoMixMatAgeVacc} exclude edges representing infrequent interactions. The rationale for the latter choice is that interactions that are less frequent than once every six months are unlikely to result in the social influence of influenza vaccination behavior. One other analysis is to create the mixing matrix that includes both ego-alter and alter-alter ties and compare it to a mixing matrix that includes only ego-alter ties. The comparison can help reveal the uncertainty range in the mixing matrix and can be used to specify a tolerance measure in how accurately it needs to be reproduced by a generative algorithm. The reason for including and excluding alter-alter ties is that in contrast to the ego where we have all the ties, the ties connecting the alters are only a sample based on what the ego reports. In reality, each elicited alter is likely to have many other ties to other people which is not apparent to the ego (i.e., the respondent). 
\begin{figure}[!ht]
\centering
\begin{subfigure}
  \centering
  \includegraphics[width=0.5\linewidth]{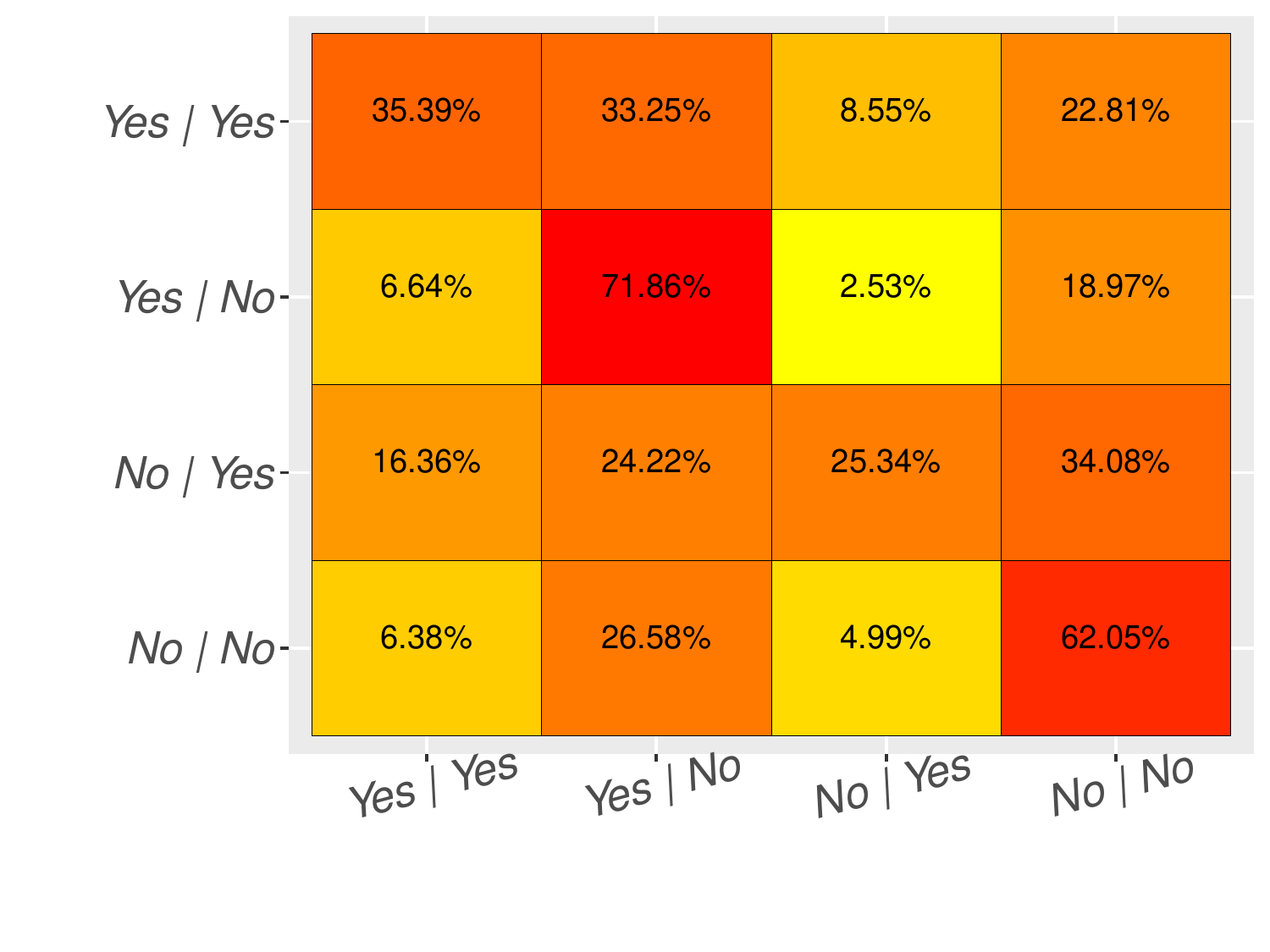}
\put(-220,190){(a)}
\end{subfigure}%
\begin{subfigure}
  \centering
  \includegraphics[width=0.5\linewidth]{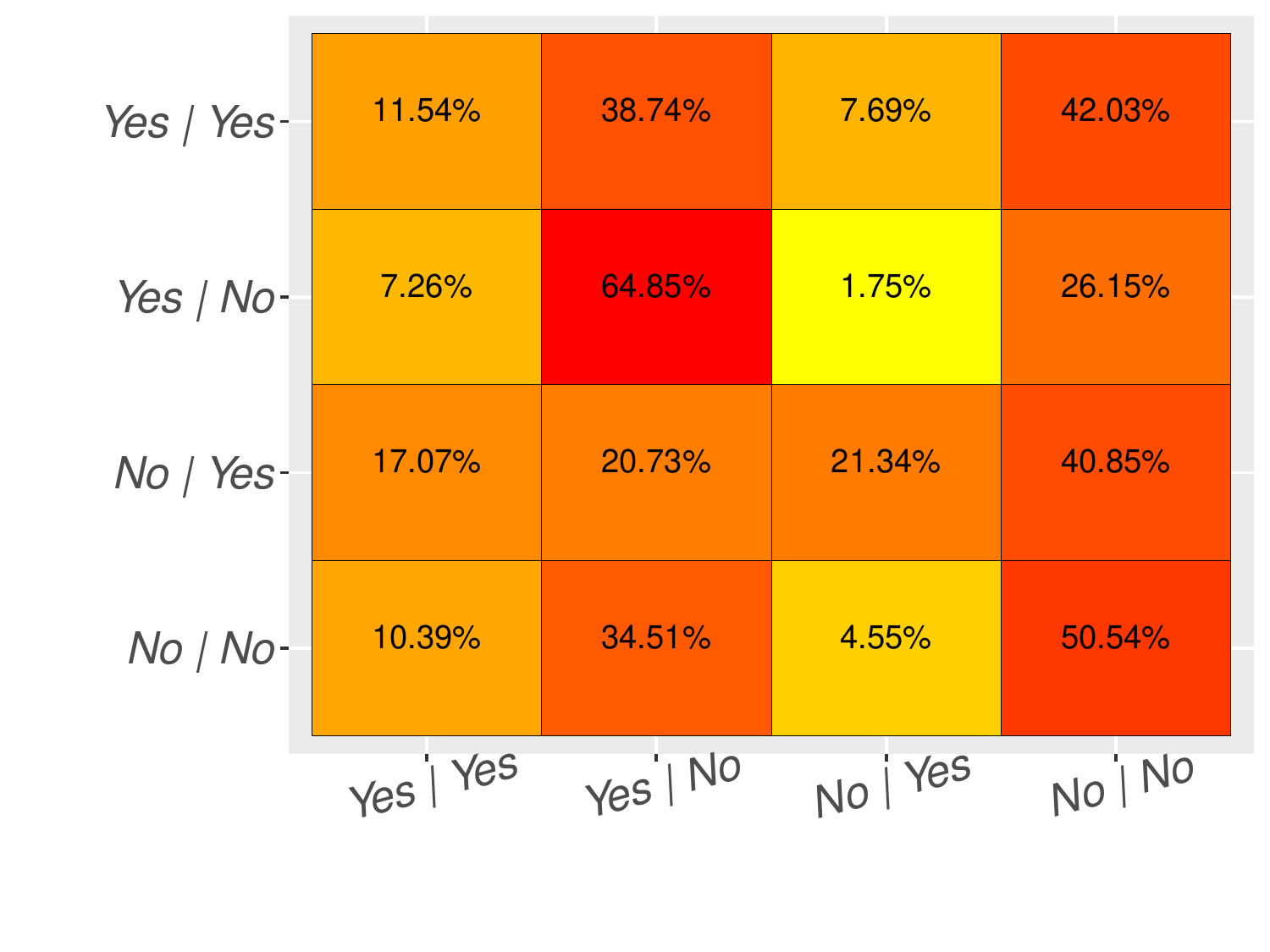}
\put(-220,190){(b)}
\end{subfigure}%
\caption{Conditional mixing-matrices using the Vaccinated for the Flu and the Contracted Flu attributes respectively describing whether or not each node vaccinated for influenza and whether or not they contracted influenza as observed by the ego (i.e., the respondent). The two conditional mixing-matrices consider the case with both ego-alter and alter-alter ties (a), and the case where the alter-alter ties have been removed (b).   
\label{Fig:EgoMixMatVaccFlu}}
\end{figure}

\clearpage
\bibliography{refs, COVID-19, Influenza_ABM}
\bibliographystyle{JHEP}

\end{document}